\documentclass[sn-mathphys]{sn-jnl}

\usepackage{csquotes}
\usepackage{booktabs}
\usepackage{longtable}
\usepackage{makecell}
\usepackage{hyperref}
\hypersetup{colorlinks,allcolors=black}
\usepackage{footmisc}
\usepackage{natbib}

\jyear{2021}%

\theoremstyle{thmstyleone}%
\theoremstyle{thmstyletwo}%

\theoremstyle{thmstylethree}%

\begin{document}
\title[Article Title]{Explainable AI for clinical and remote health applications: a survey on tabular and time series data}


\author*[1]{\fnm{Flavio} \sur{Di Martino}}\email{flavio.dimartino@iit.cnr.it}

\author*[1]{\fnm{Franca} \sur{Delmastro}}\email{franca.delmastro@iit.cnr.it}

\affil*[1]{\orgdiv{Institute for Informatics and Telematics (IIT)}, \orgname{National Research Council of Italy (CNR)}, \orgaddress{\street{Via Moruzzi 1}, \city{Pisa}, \postcode{56100}, \state{Italy}}}


\abstract{Nowadays Artificial Intelligence (AI) has become a fundamental component of healthcare applications, both clinical and remote, but the best performing AI systems are often too complex to be self-explaining.
Explainable AI (XAI) techniques are defined to unveil the reasoning behind the system's predictions and decisions, and they become even more critical when dealing with sensitive and personal health data.
It is worth noting that XAI has not gathered the same attention across different research areas and data types, especially in healthcare.
In particular, many clinical and remote health applications are based on tabular and time series data, respectively, and XAI is not commonly analysed on these data types, while computer vision and Natural Language Processing (NLP) are the reference applications.
To provide an overview of XAI methods that are most suitable for tabular and time series data in the healthcare domain, this paper provides a review of the literature in the last $5$ years, illustrating the type of generated explanations and the efforts provided to evaluate their relevance and quality. Specifically, we identify clinical validation, consistency assessment, objective and standardised quality evaluation, and human-centered quality assessment as key features to ensure effective explanations for the end users.
Finally, we highlight the main research challenges in the field as well as the limitations of existing XAI methods.
}




\maketitle

\section{Introduction}\label{intro}
Artificial Intelligence (AI) has become in the last few years a building block of modern health services, improving efficiency and providing a concrete support to the decision making process.
However, the lack of transparency and interpretability of AI still remains one of the main barriers to its real adoption in the clinical practice \cite{topol2019high}, and even more in those systems that require a direct interaction with a non-expert user (e.g., remote patient monitoring and personalised support).
In fact, in the healthcare domain we have to distinguish among different types of users dealing with AI systems: (i) the clinical/medical personnel (i.e., the domain expert), who needs explanations to increase their trust in the system and, at the same time, can provide a clinical validation; (ii) the technical developer, who is in charge of the reliability of the model; (iii) the patient or monitored subject, who needs interpretable and personalised explanations.
Explainable AI (XAI) techniques have the potentiality to support all these types of users by making AI models more expressive and improving human understanding and confidence in AI-empowered Decision Support Systems (DSSs) \cite{das2020opportunities}, but they generally do not offer a one-fits-all solution \cite{cina2022we}.
\\
In addition, the healthcare domain includes a variety of AI applications related to different research areas, each of them requiring appropriate explanations.
To date, biomedical imaging (e.g., X-rays, CT-scans, ultrasounds) is one of the most active XAI application fields, trying to explain model classification by generating saliency maps to highlight the relevance of different image regions (aka, super-pixels) to a given prediction \cite{tjoa2020survey}. It is generally applied in Computer-Aided Diagnosis (CAD) with different targets, such as cancer \cite{gulum2021review} and COVID-$19$ \cite{mondal2021xvitcos,faruk2021residualcovid}.
However, many health applications are also based on several other data types, such as tabular and time series data that can derive from clinical information, such as Electronic Health Records (EHRs), as well as from real-world data collected by IoT and personal mobile devices. Currently the explainability of models applied to those data has not gathered the same attention by the research community yet.
This trend is actually unexpected and in contrast with their widespread in the real life.
EHRs are the major source of tabular data for clinical settings \cite{payrovnaziri2020explainable}, which contain rich, longitudinal, heterogeneous, and patient-specific information including demographics, clinical information, questionnaire outcomes, lab tests, and vital sign measurements.
On the other hand, the recent diffusion of e-health and m-health systems offers increasing opportunities for remote health monitoring and decision making that heavily rely on the analysis of Multivariate Time Series (MTS) \cite{kok2022explainable}.
The integration between these data sources and data-driven AI models may provide a fundamental contribution to the delivery of early, personalised, and high-quality care both in clinical and remote settings, and explainability becomes fundamental to provide effective explanations both to expert and non-expert users.
\\
However, several existing XAI methods are currently not suitable for tabular data due to the unique characteristics that distinguish them from images (and also text records), such as the potential interactions between features and the coexistence of continuous and categorical predictors \cite{sahakyan2021explainable}.
In addition, the majority of methods applied to time series data are generally adapted from computer vision and Natural Language Processing (NLP) fields in order to highlight which specific signal components get the most attention from the model while the classification is performed.
As a result, these methods might not account for specific features of time series data, such as recurrent spatio-temporal patterns and correlations between multiple channels and/or sensing modalities. The unintuitive nature of some time series also poses additional challenges as even domain experts may struggle in understanding the information hidden in the most relevant signal components \cite{rojat2021explainable}.
\\
Selecting existing XAI methods suitable for this data is not sufficient to effectively bring explainability in healthcare applications without an extensive assessment of the generated explanations \cite{markus2021role}.
Clinical validation is currently one of the most widely discussed requirements to build trust in AI-empowered decision making in healthcare \cite{amann2020explainability}, as it is a critical step for a model to gain clinical credibility by matching data-driven knowledge with evidence-based assessment.
In addition, evaluating the level of consistency of explanations generated by multiple methods may provide some preliminary insights into AI systems' reliability, despite sensitivity analysis may be needed to better expose model vulnerabilities and draw more accurate conclusions about stability and robustness properties \cite{linardatos2020explainable}.
However, evaluating clinical soundness and consistency of explanations does not enable a formal quality assessment as well as a systematic comparison of XAI methods, but standard metrics and practices are still missing in the research community \cite{guidotti2018local}. 
According to \cite{doshi2017towards}, quality evaluation approaches can be divided into {\it functionality-grounded}, {\it application-grounded}, and {\it human-grounded}.
The former represents an initial and objective assessment of explanations based on the definition of quantitative metrics, which enable to select the best method regardless end users’ needs and preferences.
The other two approaches are complementary: human-in-the-loop evaluations are necessary to tune the explanations with respect to the target audience, by considering both domain experts and non-expert users (generally the patient).
However, evaluating visual and textual explanations supplied by an algorithm is necessary but not sufficient to enable informed and confident decision-making if the interactivity with the end users is neglected \cite{arrieta2020explainable}.
In fact, multidisciplinary collaboration is the premise to detect relevant interactions between end users and AI systems leading to a better interpretation of model predictions, which should be integrated by design or iteratively added to meet end users' needs.

\subsection{Contribution}
This survey focuses on the application of XAI techniques to models learned from clinical data stored in EHRs, and real-world data collected by IoT and personal mobile devices. 
However, it is not sufficient to focus on the application of existing XAI methodologies in this field but it is essential to understand how explanations can be validated and evaluated to close the loop. For this reason, we investigated those works who include also clinical validation, consistency assessment and quality evaluation of the proposed XAI techniques for healthcare. 
During our research we found several works experimenting XAI methodologies without providing any kind of assessment of the generated explanations. We summarise them in a table to highlight the impact of this research field on the scientific literature, but then we focus on more structured studies including the explanations evaluation, from different perspectives. Specifically, we consider clinical validation to satisfy the strict constraints imposed by the medical domain, consistency assessment of explanations across multiple models and/or XAI methods, and finally formal quality evaluations including novel, objective, and quantitative metrics as well as user-centered studies.
Therefore, the contribution of this survey can be summarised as follows:
\begin{itemize}
    \item an overview of the most prominent XAI methods applicable to tabular and time series data (Section \ref{background});
    \item a literature survey (methodology is reported in Section \ref{methods}) related to the usage of XAI in healthcare applications targeting these data types and explanations' assessment (Section \ref{results});
    \item a discussion section to highlight the main limitations of the presented methods, and open research challenges to improve explainability from both a methodological and user-based perspective (Section \ref{discussion}).
\end{itemize}

\section{Background}\label{background}
Several technical features come into play when analysing the emerging landscape of XAI, which makes the taxonomy of existing methods not unique.
Prior surveys addressing XAI from a more general and application-independent perspective classifying methods based on different aspects, which can be summarised as follows:
\begin{itemize}
    \item \textbf{Scope}: \textit{local} or \textit{global}.
    Local methods aims to explain predictions only for single data instances, whereas global methods enable understanding the reasoning of a learning algorithm as a whole.
    \item \textbf{Stage of applicability}: explainability may be applied throughout the main stages of AI development pipeline, namely pre-modelling, model-development, and post-modelling. However, this classification generally distinguishes between \textit{ante-hoc} and \textit{post-hoc} methods.
    In the first case, explainability is embedded in the structure of the model and is available directly at the end of the learning phase, whereas in the second case explanatory techniques are used to unveil the \textit{\enquote{black-box}} of complex models after their training.
    \item \textbf{Target model}: \textit{model-agnostic} methods can be theoretically applied to any kind of AI model, whereas \textit{model-specific} ones are tailored to certain model classes, such as Convolutional Neural Networks (CNN).
    \item \textbf{Explanation form}: attribution methods generate importance scores for each input, also providing input ranking. Similarly, \textit{heatmaps} such as saliency and attention maps, compute and visualise adaptive weights related to the relevance of each data point. \textit{Decision rules} (i.e., IF-THEN rules), as well as decision trees, represent other common explanation formats.
    Finally, \textit{dependency plots} show the expected target response as a function of the input features of interest, thus potentially revealing both relationships between target and inputs (e.g., linear, non-linear, monotonic) and interactions among input variables.
    \item \textbf{Algorithmic nuances}: the underlying algorithm used to extract explanations.
    \textit{Perturbation-based} methods manipulate parts of the input by replacing, removing, or masking them in order to generate attributions for individual features, data points, or signal regions.
    \textit{Gradient-based} methods are tailored to Deep Neural Networks (DNN), as they obtain attributions by using gradient (i.e., partial derivatives) to compute the impact of each input on model outcomes via one or more forward/backward pass through the network.
    On the other hand, \textit{instance-based} methods extract a subset of relevant features that is needed to retain/change a given prediction without applying any perturbation to original data.
\end{itemize}
In addition, taxonomy of XAI methods also depends on the data type that is fed as input to the model to be explained, which can be images, text, graph, tabular, or time series data.
As already outlined in Section \ref{intro}, most of existing techniques have been originally conceived for images or text data, therefore they could not be suitable or readily applicable to tabular and time series data.
For this reason, in the next subsections we first provide the reader with a summary of current XAI methods that are best suited for these data types.

\subsection{XAI for tabular data}\label{background_tabular}
From the literature analysis, it may be noticed that the majority of the existing XAI techniques applicable to tabular data are model-agnostic.
Feature ablation and permutation methods are straightforward options to estimate feature importance for any black-box estimator, by measuring how the prediction error changes when removing a given feature or randomly shuffling its values, respectively.
\textit{Mean Decrease in Accuracy} (MDA) is a popular choice in permutation studies, but other scoring metrics can be used as well.
Tree-based models also provide an alternative measure of feature importance based on the \textit{Mean Decrease in Impurity} (MDI), in which impurity is quantified by the splitting criterion of the decision tree (normally, Gini's index).
Therefore, MDI computes feature importance as the total decrease in node impurity (i.e., homogeneity of labels within the node) for every splits across all trees that include a given feature, weighted by the proportion of samples reached at each node.
\\
\textit{Shapley Additive Explanations} (SHAP) \cite{lundberg2017unified} is probably the state-of-the-art method for XAI, and it is built on the concept of Shapley values coming from the coalitional game theory. This concept has been transferred to the Machine Learning (ML) domain by considering a prediction task as a game, features as players, and coalitions as all possible feature subsets, thus making it very suitable for tabular data.
SHAP computes feature importance scores as the \textit{average marginal contribution} that each feature brings to an individual prediction, where {\it \enquote{marginal}} stands for the difference between the actual predicted value and a \textit{base value} used as a reference.
According to \cite{lundberg2017unified}, this value is defined as \textit{\enquote{the value that would be predicted if we did not know any feature for the current output}}; in other words, it represents the average prediction over training/test set.
On the other hand, the {\it \enquote{average}} terms implies computing the mean value across all permutations, i.e., all the possible subsets that include a specific feature.
To apply SHAP provides several advantages.
First, local explanations can also be aggregated to get global explanations. In addition, due to the axiomatic assumptions included in SHAP theoretical foundations, global explanations are more reliable than those obtained by most feature attribution methods.
Finally, SHAP offers different algorithmic implementations to explain any kind of model.
\\
\textit{Local Interpretable Model-agnostic Explanations} (LIME) \cite{ribeiro2016should} technique is another popular model-agnostic methods to obtain local interpretability.
Although a model may be very complex globally, LIME produces an explanation by approximating it by an interpretable surrogate model (generally, a sparse linear model) only in the neighborhood of the instance to be explained.
This is achieved by first creating a new dataset consisting of data points randomly drawn in the proximity of the instance of interest, along with the corresponding predictions of the original model.
Then, a linear classifier is trained using the perturbed data set, in which each sample is also weighted by its proximity to the target instance through an appropriate weighting kernel.
Finally, a very similar method to Least Absolute Shrinkage and Selection Operator (LASSO) regularisation is applied to keep only the most important features. As a result, regression coefficients are used as feature importance scores.
\\
LIME works better for local interpretability, but global explanations may also be derived.
A first option is to simply average importance scores across data instances, but this approach may suffer from a high variance due to multiple local approximations.
On the other hand, \textit{LIME SubModular Pick} (LIME-SP) optimisation algorithm allows the selection of a representative, non-redundant set of explanations as exemplary cases of how the model behaves for each class. However, this method just provides some global understanding, and not a comprehensive picture of the overall model reasoning.
\\
Despite the key intuition of using local surrogate models cuts down LIME computational complexity (and time), it reduces outcome stability as well.
The choice of simple sparse linear model implies that if the underlying model is highly non-linear even in the locality of the prediction, the explanations may not be faithful. In addition, explanations are originated from random perturbations of the original input space, which may not be representative of the instance to explain.
Therefore, several techniques have been proposed trying to improve LIME stability.
ALIME \cite{shankaranarayana2019alime} exploits an auto-encoder as weighting function, whereas hierarchical clustering is adopted in DLIME \cite{zafar2019dlime} instead of random perturbations to group training data, and then it selects the cluster closer to the target instance.
In addition, an alternative weighting approach has been proposed in ILIME \cite{elshawi2019ilime}, in which each perturbed instance is weighted based on its influence on the target instance to be explained, and the distance from it.
\\
\textit{Anchors} algorithm \cite{ribeiro2018anchors} represents an evolution of LIME that exploits reinforcement learning and graph search to detect a region in the neighbourhood, defined by a range of values for some features, representing a sufficient and high-precision condition (i.e., an \textit{\enquote{anchor}}) to guarantee local prediction, such that any changes to other features do not essentially alter model outcomes.
These range values are then converted into IF-THEN rules, which can be used to explain not only the target instance but also every other instance meeting the anchor.
\\
All the above mentioned methods are aimed at explaining how model outcomes are generated. However, there also exists other techniques, falling within the umbrella of \textit{counterfactual explanations}, aiming at detecting the minimal feature changes that are necessary to drive a prediction towards a desired different output.
Counterfactual explanations are generally formulated as an optimisation problem, so the main difference between existing techniques lies in the optimisation method and/or in the loss function to be minimised.
A first method, called \textit{unconditional counterfactual explanations}, has been proposed in \cite{wachter2017counterfactual} for differentiable models, such as neural networks, in which the gradients of the loss function can be computed.
The loss function to be minimised in this case is the distance between the counterfactual and the original data point, subject to the constraint that the model classifies the counterfactual with the desired (and different) label.
\cite{guidotti2018local} proposed \textit{Local Rule-Based Explanations of black-box decision systems} (LORE), a model-agnostic method to extend counterfactual explanations beyond differentiable models.
This approach exploits a genetic algorithm to create a synthetic neighborhood for a target instance, then it retrieves both a decision rule (similar to an anchor) and a set of counterfactual rules to identify changes leading to different predictions.
More recently, \cite{looveren2021interpretable} proposed a technique to obtain counterfactual explanations for differentiable classifiers based on prototypes, in which each class-specific prototype is computed as the average encoding over the $K$ nearest instances with the same class label in the latent space generated by a CNN encoder. Once found, prototypes are embedded into the model objective function to guide the perturbations towards an interpretable counterfactual.
\\
Sensitivity analysis is another category of XAI methods aimed at computing feature relevance that works by measuring how much model predictions are sensitive with respect to changes in one or more input parameters.
In addition, it may also be used for model inspection, to detect how altering some internal components/properties affects the model outcomes.
Traditional sensitivity analysis methods estimate the importance of each input variable as its contribution to the output model variance. Morris' method \cite{morris1991factorial} is one of the most popular approaches for sensitivity analysis.
It works by dividing the range of each variable and iteratively making one change at time within the range of each input variable, in order to cluster inputs in three categories: $1$) features with no effect, $2$) features with linear effect and no interactions, $3$) features with non-linear effects and/or interaction effects.
Despite this method is very complete, it is also very computational costly, in particular as the number of predictors increases.
Therefore, other lightweight solutions have been proposed, such as those based on Analysis of Variance (ANOVA) decomposition \cite{saltelli2010variance}.
Moreover, adversarial examples represent a more recent and innovative approach to achieve sensitivity analysis, by exploiting the vulnerability of AI models against adversarial attacks as proxy of input relevance.
Specifically, they apply intentional changes to input variables in order to generate new samples that can mislead model predictions, then quantify variable relevance depending on how the changed inputs are able to fool the model. However, adversarial example-based sensitivity analysis methods are currently used for computer vision and NLP tasks, while their effectiveness for other data types, such as tabular data, still need to be deeply investigated.
\\
Visual explanation techniques are also available to highlight the relationship among target and input variables.
\textit{Partial Dependence Plots} (PDP) \cite{friedman2001greedy} show the average marginal effect of one or two features on model outcomes, assuming that the features are uncorrelated (which may not always be true).
Their equivalent for local predictions, called \textit{Individual Conditional Expectation} (ICE) plots, has been proposed by \cite{goldstein2015peeking} to visualise the dependence of the prediction on a feature for each instance separately.
Finally, \textit{Accumulated Local Effects} (ALE) \cite{apley2020visualizing} plots represent an unbiased alternative to PDP, as they account for feature correlation when showing feature influence on model outcomes.
\\ 
To conclude this overview of XAI techniques suitable for tabular data, it is worth to briefly mention some methods available to globally explain complex models by approximating them with simpler ones, such as decision rules/trees.
\textit{InTrees} \cite{deng2019interpreting} has been proposed as a framework to extract a compact set of decision rules from tree ensembles, by selecting and pruning rules according to a trade-off among their frequency within tree nodes, their error rate, as well as their length. Additional methods have been also developed for approximating DNN \cite{wu2018beyond} and Support Vector Machine (SVM) models \cite{barakat2007rule}.

\subsection{XAI for time series data}
To date, Recurrent Neural Networks (RNN) generally represent the best strategy to deal with time series data, thanks to their memory state and their ability to learn relations through time.
CNN with temporal convolutional layers are able to build temporal relationships as well, also extracting high level features from raw data. The introduction of these models to solve MTS classification and forecasting tasks significantly boosted predictive performance without requiring heavy data pre-processing.
As a result, the majority of existing XAI methods applicable to this data are specific for these models.
\\
As far as CNN is concerned, almost all methods are inherited from computer vision field to obtain post-hoc explainability.
According to the underlying algorithm concepts, they can be divided into gradient-based and perturbation-based methods.
Gradient-based methods measure how much a change around a local neighborhood of the input corresponds to a change in the model output by running a single forward or backward pass of Gradient Descent (GD) algorithm (or similar) in the network.
They have been originally conceived as pixel attribution methods, also referred to as saliency maps, in order to highlight the pixels/super-pixels that are relevant for a certain image classification. However, they may also be adapted for time series data in order to highlight the most relevant data points within a $1$-D sequence.
\\
A first gradient-based explanation method has been proposed in \cite{simonyan2013deep} to create saliency maps corresponding to the gradient of an output neuron with respect to changes in a small neighborhood around the input, thus highlighting image regions that are relevant for a target class. 
Afterwards, \cite{sundararajan2017axiomatic} proposed \textit{Integrated Gradients}, which essentially represent a variation of the gradient computing technique implemented in the previous method, by directly attributing the network predictions to its input features.
\\
\textit{Deep Learning Important FeaTures} (DeepLIFT) \cite{shrikumar2017learning} is one of the most popular algorithms for post-hoc explanation of deep networks. It also works by attributing importance scores to input features, in which each score represents the impact on network outcomes of changing the original feature value to a reference baseline, which can be empirically chosen by end users. This approach is also known as \textit{\enquote{Gradient*Input}} methods, since it also multiplies the gradient by the input signal.
This operation essentially represents a $1^st$ order Taylor approximation of the output changes when inputs are set to zero, and it has proven to enhance the visualisation of saliency maps with respect to previous gradient-based methods.
Moreover, DeepLIFT has been later combined with Shapley values to build \textit{DeepSHAP}, a specific framework to approximate SHAP feature attributions for any Deep Learning (DL) model.
This method differs from the original DeepLIFT by using a distribution of background samples instead of a single reference value to change each selected feature, and using Shapley equations to linearise network components such as softmax operators.
\\
\textit{Deconvolution} \cite{zeiler2014visualizing}, is a technique to visualise CNN-based saliency maps by using \textit{DeconvNets} networks \cite{zeiler2011adaptive}, which leverage the same CNN layers and operators in an exactly reversed order for mapping encoded features to input (i.e.,pixels), as opposite to the standard CNN data processing pipeline.
Moreover, the \textit{guided backpropagation} technique \cite{springenberg2014striving}, also known as guided saliency, has been proposed as a variant of the deconvolution approach to extend its applicability to all possible CNN architectures to visualise saliency of the learned features.
\\
\textit{Class Activation Mapping} (CAM) \cite{zhou2016learning} is another CNN-based XAI methods, originally developed to detect class-specific image regions used by the network to make predictions. CAM computes a vector by concatenating the average activations of convolutional feature maps that are placed immediately before the last prediction layer, then it feeds a weighted sum of this vector to the final layer. In this way, the relevance of class-specific image regions, and of features learned in the latent space in general, can be retrieved by projecting back the weights of the output layer onto the convolutional feature map.
However, CAM implementation requires CNN to have a specific architecture in their final layers, thus limiting its applicability. In addition, it is suitable only to highlight high-level representations learned at the last stages, whereas it cannot provide any explanation of low-level representations that are learned at earlier stages.
To overcome these limitations, a more general CAM implementation, \textit{Gradient-weighted CAM} (Grad-CAM) \cite{selvaraju2017grad}, has been developed to extend its applicability to any CNN, which relies on gradient information flowing to the last convolutional layer to locate the most important image regions within a saliency map in an architecture-independent fashion.
\\
Finally, \textit{Layer-wise Relevant Propagation} (LRP) \cite{bach2015pixel} is an interpretability method to decompose DNN by propagating their predictions backward without altering the output magnitude.
By starting from neurons in the final prediction layer, and moving back to neurons of the input layer, the prediction value is backpropagated in such a way that each neuron redistributes to the preceding layer the same amount of information received from the higher layer.
\\
Unlike gradient-based methods, perturbation-based methods compute the contribution of individual parts of the input by removing or randomly replacing them, then they exploit a distance metric to measure the difference in the model decision function.
As a result, a higher difference in the prediction outcome indicate a higher contribution of the input component that has been altered.
In this context, the \textit{Occlusion} method by \cite{zeiler2014visualizing} is one of the most used techniques coming from computer vision applications. It acts as sensitivity analysis method by systematically replacing different contiguous parts of an input with a given baseline, then monitoring the decrease in the prediction function. The implementation of this method is not computationally expensive and can be applied to any network architecture as it does not require specific internal components.
\\
For what concerns RNN, \textit{Attention} mechanism \cite{bahdanau2014neural}, sometimes also referred as \textit{Self-Attention}, is currently the state-of-the-art method for explainability.
Attention originates from NLP domain, where it has been initially proposed as a solution to overcome the bottleneck of the original Encoder-Decoder RNN model employed for machine translation. It encodes the input sequence to one fixed length vector from which decoding the output at each time step. The main criticism of this approach is related to the difficulties for DL models to cope with very long sentences, especially those that are longer than the ones contained in the training corpus.
Differently, the attention model does not encode input sequences as a whole, but rather it develops context vectors that are filtered specifically for each output time step. Then, it searches for a set of positions within a source sentence where the most relevant information is concentrated according to the context vectors associated with these positions and all the previous generated target words, in order to predict the next word.
Besides providing a gain in predictive performance, attention also represents a powerful explanatory technique, as it highlights which words in a text corpus are the most relevant for a certain prediction.
\\
Attention mechanism is at the foundation of \textit{Attentive transformers} \cite{vaswani2017attention}, which consist of specific RNN modules that can be embedded into the learning process of neural networks to obtain ante-hoc explainability. 
The inner functioning of a transformer relies on a sparse-max function to obtain a mask, which is subsequently scaled and multiplicatively applied to the input, in order to learn adaptive weights that reflect the impact of each input data on the final prediction. 
Transformers can be used to enable interpretability at different levels, such as input features and time points. For instance, they can detect globally important variables, persistent temporal patterns, as well as significant events within a data trajectory leading to a target outcome.
Transformers have found their major applications in NLP, as demonstrated by very popular language models such as \textit{BERT} \cite{devlin2018bert} and \textit{GPT-$3$} \cite{radford2018improving}.
However, with the advent of vision transformers for image-based applications \cite{mondal2021xvitcos}, they have been also embedded into CNN architectures.
As a result, attention is the primary XAI method to explain RNN, while attention-enhanced CNN may also be found in some cases.
\\
Eventually, there are also additional data mining methods applicable to time series data, although less used in practice.
Fuzzy inference systems are a viable solution to simulate logical thinking, whereas \textit{Symbolic Aggregate ApproXimation} (SAX) \cite{lin2003symbolic} works by converting time series into strings.
Specifically, it first divides each time series into equally-length segments. Then, by assuming a Gaussian distribution of input data, it assigns a symbol to each segment by mapping the average segment value with the corresponding probability, in order to discover recurrent patterns. Prototype Learning (PL) is also a compatible approach for time series data, which generates samples to be used as reference to explain the typical pattern of all data instance belonging to the same class. In this context, \textit{Shapelets} \cite{ye2009time} is a time series-specific, prototype-based method to explain AI models by extracting input sub-sequences that are representative to each class. This method also provides more interpretable, accurate, and faster results with respect to the standard PL approach that selects class prototypes from the nearest samples in latent embedding space to a given data point.

\section{Survey methodology}\label{methods}
We searched \textit{IEEExplore}, \textit{Springer}, \textit{ACM}, and \textit{Elsevier} digital libraries, using different search strings obtained as a combination of the following subsets of keywords:
\begin{enumerate}
    \item \textit{\enquote{Explainable}}, \textit{\enquote{Interpretable}};
    \item \textit{\enquote{AI}}, \textit{\enquote{Artificial Intelligence}}, \textit{\enquote{Machine Learning}};
    \item \textit{\enquote{Healthcare}}, \textit{\enquote{Health}}, \textit{\enquote{Wellbeing}}.
\end{enumerate}
In addition, we also analysed \textit{Scopus} to double check the screening process conducted for the previous databases. This search also highlighted an additional cluster of relevant works coming from other sources, such as Nature Research journals, which have been further investigated.
\\
Motivated by the very recent development and application of XAI methods, and especially in the healthcare domain, we filtered out the search by selecting papers published in the last $5$ years (i.e., between $2017$ and $2021$).
Table \ref{tab_paper_by_year} reports the percentage of selected papers for each year within the date range, and it further confirms the latest exponential increase of XAI applications in the health domain, with the majority of research studies published in the last $2$ years.
\\
In order to focus only on XAI methods suitable for tabular and time series/sequence data, we first excluded applications targeting biomedical images, such as computerised tomography scans, magnetic resonance imaging, and ultrasound images, which are typically $3$-D data with additional time dimension and/or multiple channels ($4$-$5$D) represented by tensors to be fed to DNN models.
Then, we also excluded NLP tasks to extract meaning from unstructured medical text records (e.g., patient prescription notes), which generally make use of high-dimensional word embeddings as input data.
Finally, we did not consider genomics and molecular biology applications based on graph data structures.
\\
In turn, tabular data include independent (i.e., \enquote{static}) EHR, as well as feature datasets derived from physiological signals.
In the first case, each patient's record is treated as an independent observation, or else multiple observations are aggregated over a specific time window, still resulting in static data.
In the second case, feature datasets are a high-level and multidimensional input representation resulting from the application of signal processing pipelines, such as signal framing/windowing and handcrafted feature extraction.
On the other hand, time series/sequence data may be divided into longitudinal EHR, which consist of a sequence of visits/admissions to model patient trajectories using clinical variables or medical codes (e.g., ICD-$10$-CM codes), and raw signal time series.
\\
For what concerns the stage of XAI applicability while building a model, we considered as eligible both post-hoc methods to add interpretability to already developed black-box models (e.g., tree ensembles, CNN), and ante-hoc methods that embed interpretability in the structure of the model, thus making it available directly at the end of the learning phase.
However, we did not include \textit{\enquote{glass-box}} approaches in which interpretability is simply addressed in terms of development of intrinsically Interpretable ML (IML) models.
This typically involves three model classes, namely sparse linear classifiers (e.g., linear/logistic regression, generalised additive models), discretisation methods (e.g., rule-based learners, decision trees), and instance-based models (e.g., k-Nearest Neighbors, (k-NN)) \cite{du2019techniques}.
A summary of the inclusion and exclusion criteria used for the literature survey is listed in Table \ref{tab_criteria}.
\\
As a result, we reached a total of $71$ publications at the end of the search, including $46$ journal articles ($64.8\%$) and $25$ conference papers ($35.2\%$) reporting original studies.
The number of selected papers for each digital library is shown in Table \ref{tab_data_sources}, whereas Table \ref{tab_paper_by_journal} illustrates the distribution of surveyed articles across the journals.

\begin{table}[!htbp]
\centering
\begin{minipage}{\textwidth}
\caption{Inclusion and exclusion criteria.}\label{tab_criteria}%
\begin{tabular}{ll}
\toprule
\textbf{Inclusion criteria} & \textbf{Exclusion criteria}\\
\midrule
Application to tabular data & Application to biomedical images\\
Application to time series/sequence data & Application to medical text data\\
Post-hoc XAI & Application to graph data\\
Ante-hoc XAI & IML models only\\
\bottomrule
\end{tabular}
\end{minipage}
\end{table}

\begin{table}[!htbp]
\begin{minipage}[t]{0.5\linewidth}
\centering
\caption{Percentage of research works by date.}\label{tab_paper_by_year}
\begin{tabular}{c c c}
\toprule
\textbf{Year} &\textbf{Papers (\%)} \\
\midrule
$2021$ &$60.6$\\
$2020$ & $31.0$\\
$2019$ &$5.6$ \\
$2018$ &$1.4$\\
$2017$ & $1.4$\\
\bottomrule
\end{tabular}
\hfill%
\end{minipage}%
\begin{minipage}[t]{0.5\linewidth}
\centering
\caption{Number of research works by data source.}\label{tab_data_sources}
\begin{tabular} {c c c}
\toprule
& \multicolumn{2}{c}{\textbf{Papers (\#)}}\\
\cmidrule(r){2-3}
\textbf{Digital library}& \textbf{Journal} & \textbf{Conference}\\
\midrule
IEEExplore&$15$&$15$\\
Springer&$10$&$3$\\
ACM&$7$&$7$\\
Elsevier&$6$&$0$\\
Scopus&$8$&$0$\\
\bottomrule
\end{tabular}
\hfill%
\end{minipage}
\end{table}

\begin{table}[!htbp]
\centering
\noindent\begin{minipage}[t]{\textwidth}
\footnotesize
\caption{Number of surveyed articles by journal.}
\label{tab_paper_by_journal}
\begin{tabular}{l c}
\toprule
\textbf{Journal} & \textbf{Publications (\#)} \\
\midrule
IEEE Journal of Biomedical Health Informatics& $7$\\
BMC Medical Informatics and Decision Making & $7$\\
Nature Scientific Reports& $6$\\
IEEE Access & $4$\\
ACM Transactions on Computing for Healthcare&$3$\\
\shortstack[l]{Proceedings of the ACM on Interactive, Mobile,\\Wearable and Ubiquitous Technologies (IMWUT)}& $3$\\
IEEE Transactions on Visualization and Computer Graphics&$2$\\
Computers and Biology in Medicine&$2$\\
Artificial Intelligence in Medicine& $2$\\
Nature Communications & $2$\\
IEEE Internet of Things Journal&$1$\\
BMC Bioinformatics&$1$\\
IEEE Journal of Translational Engineering in Health and Medicine&$1$\\
Journal of Translational Medicine&$1$\\
Journal of Management Information Systems&$1$\\
Computer Methods and Programs in Biomedicine&$1$\\
ACM Transactions on Management Information Systems&$1$\\
The Lancet Digital Health & $1$\\
\bottomrule
\end{tabular}
\end{minipage}
\end{table}

\begin{figure}[!htbp]
\centering
\includegraphics[width=.8\textwidth]{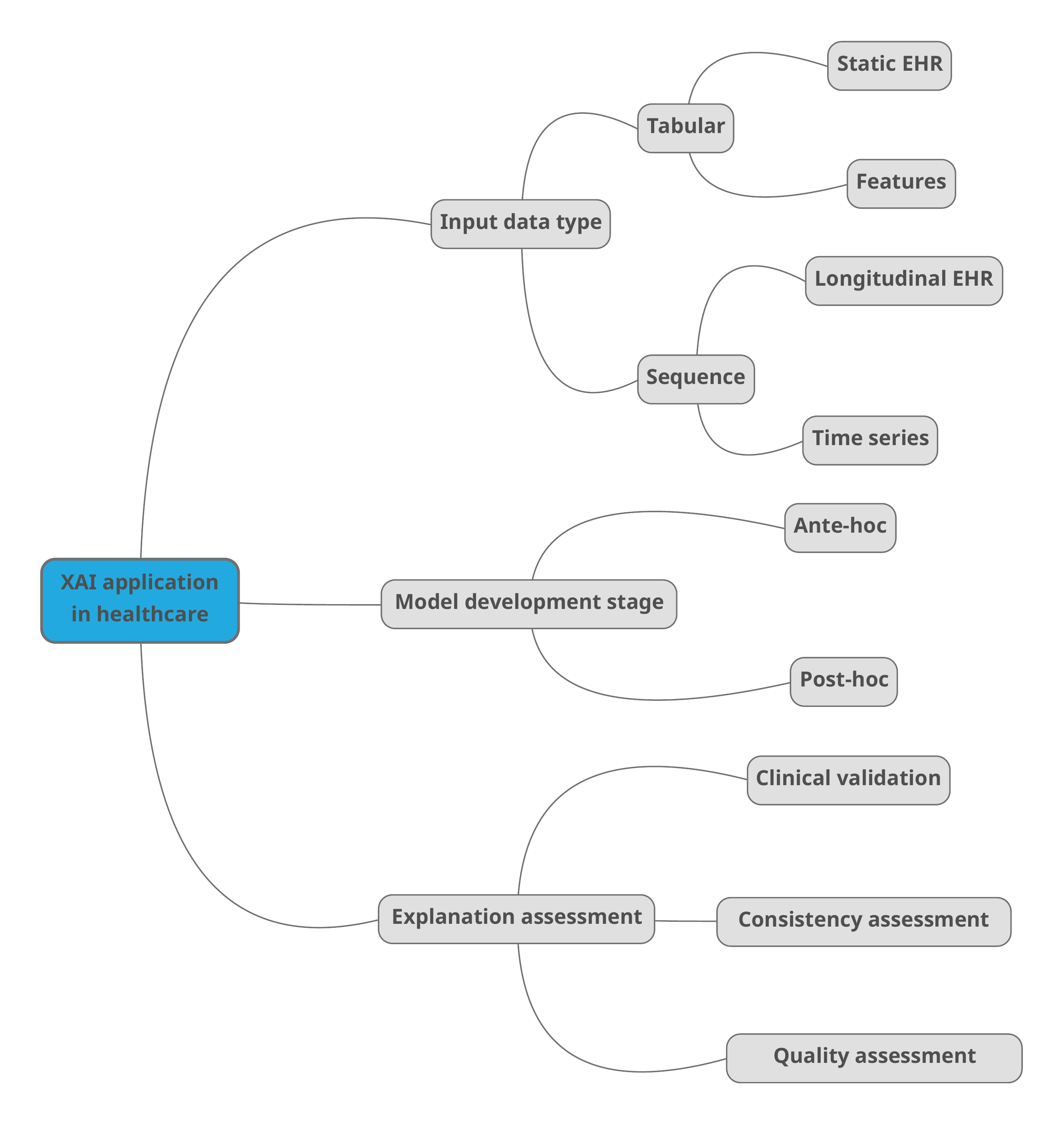}
\caption{Taxonomy mind map of XAI methods used for this survey.}\label{fig_tax_map}
\end{figure}

\section{Results}\label{results}
In the previous section, we outlined the search strategy adopted in this survey, with particular reference to the input data type and the stage of application of XAI methods within the AI development pipeline.
In this section, we present the revised works based on their main contribution in evaluating the explainability applied to the target health application.
For a complete list of acronyms used in tables and throughout the text please refer to Table \ref{tab_acronyms} in Section \ref{secA1} of the Appendix.
\\
A first research branch consists of exploratory studies aimed at experimenting XAI methodologies in order to demonstrate their possible integration with complex models to enable global/local interpretability in specific health applications, while maintaining good predictive performances.
The most relevant preliminary works using tabular and time series data are summarised in Table \ref{res_tab_1} and Table \ref{res_tab_2} in Section \ref{secB} of the Appendix, respectively.
However, these studies do not perform any evaluation of the proposed explanations, leaving room to further investigation as fundamental step to enhance confidence and trust of the medical community in decision making based on predictive AI.
After this analysis, we move a step forward by focusing on studies that also included an explanation assessment from one or more of the following perspectives: 
\begin{enumerate}
    \item {\it clinical validation}: alignment with existing medical knowledge/practice;
    \item {\it consistency assessment}: level of agreement of explanations provided by multiple models and/or XAI methods;
    \item {\it quality assessment}: it includes both quantitative evaluations based on novel metrics, and qualitative evaluations through clinician ratings and feedback.
\end{enumerate}
Such evaluations are complementary, so they may be conducted concurrently to strengthen the trust in a model. In the case more than one of the above explanation assessment procedures is performed, we present the most relevant findings.
A taxonomy mind map is shown in Figure \ref{fig_tax_map} to support the reader in understanding the main aspects covered in the overall process of selection, grouping, and analysis of the collected research works.
\\
For each category, the revised works are also summarised in tables specifying the target application, input data type and datasets, AI models, and XAI methods.
These tables also include a citation analysis derived from Google Scholar and updated to early September $2022$, to highlight the impact of the research works.
However, it is worth noting that this literature survey is limited to the past $5$ years due to the very recent application of XAI in healthcare, and that $>90\%$ of the selected papers has been published in the last $2$ years. Therefore, the number of citations provides only a preliminary analysis of the impact of the research in this field.

\subsection{Clinical validation}
Evaluating explanations from a clinical standpoint is crucial to guarantee that the inner reasoning of a model follows the domain knowledge, at least with respect to the most important and well-established clinical guidelines.
In other words, it demonstrates that model behaviour appears to be \enquote{human-like}), which also adds clinical credibility to the model itself.
The research works presented in this section propose a preliminary clinical validation of the generated explanations, which is conducted either through domain expert surveys or through the comparison with the related medical literature.
For tabular data this generally includes to investigate the global relationships between a target health condition and the input predictors evidenced by feature attribution or rule-based methods. Related research studies are summarised in Table \ref{res_tab_3}. 
On the other hand, clinical validation of explanations generated by AI systems based on time series data is performed for two main reasons:
\begin{itemize}
    \item for longitudinal EHR, to assess if the evolution of patient trajectories resulting in a target health condition is aligned with the general clinical knowledge.
    \item for physiological signals, to investigate whether the most important components highlighted by the model are clinically relevant, in order to assert decisions are made upon meaningful features.
\end{itemize}
Research studies in this area are shown in Table \ref{res_tab_4}.
Eventually, clinical comparison may also support knowledge discovery to learn novel relationships and patterns with emergent risk factors, which might be further investigated for a future integration in the current clinical practice.
\\ 
XAI methods have been applied to a wide variety of CAD applications based on tabular data, and especially on EHR.
Several solutions have been identified in this survey explaining the detection of neurological disorders.
\cite{beebe2021efficient} proposed an explainable risk assessment for imminent (i.e., within $3$ years) dementia diagnosis, by training XGBoost, Multi-Layer Perceptron (MLP), and Long Short-Term Memory (LSTM) models with multi-year, extensive cognitive testing data coming from the Religious Orders Study and Rush Memory and Aging Project (ROSMAP) \cite{a2012overview}. \\
Then, they applied SHAP to the best XGBoost classifier for both feature selection and model explanation.
Results indicate that the most relevant features come from cognitive tests collected in the most recent year, which is further confirmed by the absence of improvements in model performance when considering older visits or cumulative data to make predictions.
This finding suggests that longitudinal testing may not be necessary for future dementia diagnosis, which is consistent with other studies reporting the modest value of gradual cognitive changes in predicting future dementia onset.
In addition, SHAP feature ranking shows that the final model focuses on few cognitive tests that can be collected in a single visit in less than $20$ minutes. Specifically, combining episodic memory tests with executive functioning or language tests led to a predictive accuracy comparable with that of a full cognitive test battery ($98$ minutes), in line with the current neuropsychology.
\\
\cite{sha2021smile} proposed a novel computational framework, \textit{Systems Metabolomics using Interpretable Learning and Evolution} (SMILE), for supervised
metabolomics data analysis aimed at Alzheimer disease (AD) diagnosis.
SMILE exploits Linear Genetic Programming (LGP) as evolutionary algorithm to generate a compact predictive model, and it uses metabolite concentrations as input features stored in separated registers in a LGP program.
The algorithm has been implemented using a metabolomic dataset reported in \cite{wang2014plasma}, which includes plasma concentration of $242$ metabolites from $57$ AD subjects, $58$ Mild Cognitive Impairment (MCI) subjects, and $57$ healthy controls.
For explainability purpose, features have been ranked according to their occurrence in the best evolved models, thus providing a way to assess their importance in predicting the disease. In addition, the co-occurrence frequency between each pair of features has been considered as a correlation measure.
SMILE analysis highlighted many key metabolites that have been previously linked to AD, but also others less clinically investigated that can be potentially correlated with AD onset.
In addition, SMILE performance degraded when detecting AD from MCI subjects, suggesting for possible similarities in biomarkers between the two conditions.
\\
\cite{kim2021interpretable} proposed an interpretable model to predict Early Neurological Deterioration (END) in stroke patients from heterogeneous EHR data.
They trained several models by using data from $2363$ subjects included in the Korean Atrial Fibrillation Evaluation Registry in Ischemic Stroke Patients (K-ATTENTION) \cite{jung2019long}, a real-world dataset composed of multi-center prospective registries that mainly focus on characteristics, oral anticoagulant use, and clinical outcomes of stroke patients.
Then, SHAP has been applied to the best performing Light Gradient Boosting Machine (LightGBM) model to identify the most relevant risk factors for END.
Specifically, results obtained from the analysis of SHAP dependence plots reveal clear cut-off values associated with positive and negative probabilities of END occurrence for the $4$ main representative features, whose clinical implication may be applicable to real-world clinical practice.
For instance, they suggest that patients with severe stroke tend to develop END, thus imposing a special attention for them, while patients with mild to moderate stroke have a lower probability to develop END.
In addition, the cut-off value of fasting glucose concentration for predicting END is very similar to the current diagnostic criteria used for diabetes diagnosis ($126$ mg/dL).
\\ 
XAI methods can also be used to assess the impact of several risk factors to develop chronic health disorders.
\cite{rashed2021clinically} proposed a ML framework to detect Chronic Kidney Disease (CKD) from lab test data. To this aim, they evaluated different tree ensemble algorithms on the UCI CKD dataset \cite{rubini2015uci}, which consists of clinical tests collected from $400$ patients with a total of $24$ variables.
SHAP values have been computed for each model, then $13$ features with both the highest ranking and overlap across the models have been chosen as optimised subset to train a smaller RF model.
Moreover, these predictors have been further categorised according to their source of acquisition (.e.g,, blood test, urine test, others) to develop additional RF models based on all possible combinations.
The optimised and pathologically-selected subsets obtained by analysing SHAP explanations reach similar performance to the full input set, demonstrating that an accurate and early CKD diagnosis may be achieved by using few, low-cost, and clinically-relevant screening tools.
\\
\cite{pang2019understanding} proposed an interpretable ML approach to analyse the main risk factors associated to early childhood obesity, including demographic characteristics, lab parameters, as well as anthropometric markers and vital sign measurements.
The authors trained several ML models on the Pediatric Big Data repository, a clinical database derived from the EHR system at the Children’s Hospital of Philadelphia with more than $860$ children, then they applied SHAP to explain the best performing XGBoost model.
Results show that Well-known obesity risk factors such as weight, height, weight-for-height, geographic location, race and ethnicity appear among the most important features to the model.
On the other hand, SHAP analysis also highlights novel factors that are related to human metabolism, such as Body Temperature (BT) and respiratory rate, which may deserve further investigation to unveil possible physiological mechanisms causing these associations.
\\
Clinical validation is also fundamental when evaluating explanations related to risk prediction models developed for clinical settings, such as those related to life expectancy, post-operative outcomes, and hospital attendance.
\cite{zeng2021explainable} proposed an explainable ML model for post-operative complication risk prediction of congenital heart surgery patients from patient- and surgery-specific features and intra-operative physiological time series.
To this aim, they trained a XGBoost model on a private dataset containing data from $1964$ pediatric patient, reaching $83.1\%$ accuracy and $0.85$ Area Under the ROC Curve (AUC) values for multi-label classification with five complication types (i.e., lung, cardiac, rhythm, infectious, others).
Then, SHAP has been used to detect the most relevant factors and to perform an extensive clinical comparison of the main risk profiles learned by the model, all of which resulted to be clinically relevant. In particular, high blood pressure and prolonged cardiopulmonary bypass time patterns confirmed a high correlation with worse post-operative outcomes.
\\
\cite{zhang2021explainable} developed an explainable model for the early prediction of Acute Kidney Injury (AKI) after Liver Transplantation (LT) using more than $100$ variables mainly covering patient/donor demographic and clinical characteristics, such as comorbidities, laboratory values, and medications.
They performed a retrospective data collection of adult LT cases to build two separate datasets, which have been used for internal and external validation of several ML models, respectively.
SHAP explanation analysis for the best performing RF classifier indicated that higher pre-operative indirect bilirubin concentration, lower urine output, lower platelet count, longer anesthesia time, and high graft steatosis percentage are associated with a higher probability of AKI onset.
The distribution and relation of these risk factors with AKI diagnosis match with the current physio-pathology, thus adding clinical credibility to the final model.
\\
Explaining health predictions based on static EHR may be useful to detect the most important features, and to evaluate their relevance and correlation with respect to the model outcome from a clinical standpoint.
On the other hand, modeling longitudinal EHR data collected during multiple visits/examinations or hospital admissions may enable learning how the evolution of patients' clinical trajectories impact on a target health condition.
\cite{shashikumar2021deepaise} presented \textit{Deep
Artificial Intelligence Sepsis Expert} (DeepAISE), a novel interpretable recurrent survival model for periodical sepsis prediction after ICU admission from longitudinal lab tests and physiological measurements, such as HR, mean arterial pressure, pulse oximetry (SpO$_2$), respiration, and BT.
DeepAISE combines a Gated Recurrent Unit (GRU)-based RNN with a Weibull Cox Proportional Hazards (WCPH) semi-parametric survival model \cite{cox1992regression} to learn predictive features related to higher-order interactions and temporal patterns among clinical risk factors that maximise the data likelihood of observed time to septic events.
Specifically, it starts by producing risk scores $4$ hours after ICU admission, then it predicts the probability of onset of sepsis with $2$-hour resolution for the next $12$ hours.
The model has been subjected to a internal validation using the Emory cohort dataset (ICU patients admitted to the hospitals within the Emory Healthcare system in Atlanta, Georgia from $2014$ to $2018$), and to an external validation using a patient cohort taken by the Medical Information Mart for Intensive Care (MIMIC)-III database \cite{johnson2016mimic}, in order to ensure robustness against potential changes due to different internal procedures and patients' characteristics.
To get model explanations, feature relevance scores have been computed as the gradient of the sepsis risk score with respect to all input features, in a similar way to CNN-based saliency maps. 
Analysis of the top-$10$ features confirmed that the system exploits predictors that have been already identified as risk factors for sepsis, such as recent surgery, length of ICU stay, HR, Glasgow coma score, white blood cell count, and temperature, as well as some less appreciated but known factors such as low blood phosphorous levels.
A feature permutation study has also been performed by replacing the top-$10$ features at both global and local levels with random and/or missing values in order to assess the impact on the model performance. Obtained results indicate that locally important features may provide a better overview of the actual top contributing factors to individual risk scores, since local perturbations yield a significant drop in the model performance with respect to the global replacement strategy.
\\
\cite{sun2021attention} proposed \textit{AttenSurv}, an attention-based RNN for Heart Failure (HF) survival prediction of seriously ill patients from longitudinal and heterogeneous EHR. The network consists of three modules: $1$) a Bidirectional LSTM (Bi-LSTM) network to learn the latent representation of a patient trajectory; $2$) a MLP network for survival prediction; $3$) an attentive transformer to detect global critical risk factors.
In addition, the authors also proposed an enhanced variant, named \textit{GNNAttenSurv}, which also incorporates a Graph Neural Network (GNN) module to extract the latent correlation between risk factors.
Both networks have been tested on three public follow-up datasets, namely WHAS \cite{lemeshow2011applied}, SUPPORT \cite{knaus1995support}, and METABRIC \cite{curtis2012genomic}, and on two EHR datasets, the MIMIC-III DB and the Chinese PLAGH dataset, using different sets of dataset-specific features including lab tests, vital signs, demographics, and treatment information.
The top-$10$ features identified by the model for different survival time horizons (ranging from $3$ days up to $2$ years) have been reviewed by different medical experts, and the resulting assessment demonstrated that they represent truly informative risk factors, with some of them currently adopted for HF survival prediction in the clinical practice.
\\ 
\cite{zheng2020tracer} proposed a general DL framework, \textit{TRACER}, to facilitate accurate and interpretable decision making in healthcare applications, using in-hospital acquired AKI and mortality prediction as cases studies.
The framework relies on a RNN model as core component, named Time-Invariant and Time-Variant (TITV) network, which is designed to learn both time-variant and time-invariant feature importance scores for each patient into two separate sub-modules, by using a self-attention network and a Feature-wise Linear Modulation (FiLM)-based network \cite{perez2018film}, respectively.
The proposed network has been evaluated on the NUH-AKI dataset ($>100$k patients from the National University Hospital in Singapore) for AKI diagnosis, and on the MIMIC-III dataset for mortality prediction.
Then, extensive feature-level clinical interpretation has been performed in both domains. This analysis highlighted similar temporal patterns for features that share a similar physiological functionality, whereas diverging patterns have been found for features that have contrasting functionalities or reflect different patient clusters. Overall, feature rankings generally agree with the clinical relevance of the corresponding risk factors. 
\\
\cite{kwon2018retainvis} developed a novel visual analytics tool, named \textit{RetainVis}, to enhance interpretability and interactivity of RNN outcomes for disease diagnosis from longitudinal EHR, by integrating model explanations with additional functionalities, such as visualisation of historical patient trends, patient grouping according to desired criteria, and comparison with reference values in selected patient cohorts.
For what concerns the DL model, RetainVis relies on a variant of the original REverse Time AttentIoN (RETAIN) network \cite{choi2016retain}, named RETAINEx, which exploits Bi-LSTM modules with non-uniform time interval embedding to model irregular time spacing across consecutive visits, and a double attention mechanism at both time and feature levels.
The network has been tested over the HIRA-NPS dataset \cite{kim2014guide}, containing medical information of approximately $1.4$ million of Korean patients, using HF diagnosis as main case study.
In addition, the authors conducted an extensive analysis to review if the medical diagnoses and treatments that received the highest attention within the trajectory of patients that develop heart failure are supported by clinical evidence. Obtained results confirmed the premise that hypertensive, metabolic, and ischaemic heart diseases, and obesity are the main leading factors for heart failure, as well as one of the main related comorbidities.
\\
The recent COVID-$19$ pandemic has also fostered the development of several solutions aimed at explaining disease diagnosis based on different data sources and analysis approaches, involving both tabular data and signal processing methods.
\cite{lu2020explainable} proposed an explainable system to diagnose COVID-$19$ in suspected patients and then to predict mortality of confirmed cases using lab test data (e.g., nucleic acid test, blood test), also including medical text reports of basic diseases and symptoms.
Specifically, they used a Gradient Boosting Decision Tree (GBDT) model for disease diagnosis, whereas RF was the best choice to predict mortality.
In the first case, the GBDT model has been evaluated on a private COVID-$19$ dataset coming from Wuhan hospital (EHR data from $350$ patients), whereas a public dataset available in \cite{yan2020interpretable} has been used for the mortality prediction task ($485$ patients) by considering only lactic dehydrogenase, lymphocyte, and C-Reactive Protein (CRP) as features.
SHAP analysis shows that procalcitonin and white blood cell are the most relevant features for COVID-$19$ diagnosis, in line with the current clinical findings.
Unfortunately, the analysis of the explanatory power of textual features extracted through Term Frequency Inverse Document Frequency (TF-IDF) \cite{weiss2010fundamentals} method is not reported, which might add further value to explanation assessment.
As far as mortality prediction is concerned, results demonstrate that when the level of LDH and CRP rises and the level of lymphocyte decreases, the death probability is higher, which agrees with clinical features of death cases.
In addition, SHAP dependence plots also highlight clear boundaries associated with rising and decreasing patterns of death probability for each feature, which may act as starting point to further investigate the impact of these risk factors.
\\
\cite{gupta2021interpretable} detected COVID-$19$ recovered subjects from healthy controls using ECG-based Heart Rate (HR) and Heart Rate Variability (HRV) features.
They trained seven ML models by using $1$-minute ECG recordings from a total of $447$ subjects collected at two hospitals in Delhi, India, then they applied SHAP to the best performing XGBoost model.
From this study, it may be inferred that high-frequency power, normalised high-frequency power, HRV standard deviation, low-frequency power, and low-to-high frequency power ratio are the most influenced features after COVID-$19$ infection, and that changes exhibited by these features are related to an increased vagal activity. These findings match with earlier studies suggesting that heart vagal stimulation increases in the post-COVID recovery phase \cite{bonaz2020targeting}.
\\
\cite{pal2021pay} proposed a mixed approach for early COVID-$19$ diagnosis by integrating symptoms metadata and cough sounds.
The model architecture consists of two sub-components that are concatenated to obtain a final prediction: a TabNet \cite{arik2021tabnet} for generating embeddings from patient characteristics, diagnosis, and symptoms, and a DNN to generate cough embeddings from temporal and spectral acoustic features extracted through signal processing.
Both networks integrate an attentive transformer to learn feature relevance from each data modality.
The evaluation conducted on a medical dataset containing $30$k cough audio segments and associated symptoms from $150$ patients with four cough classes (COVID-$19$, asthma, bronchitis, and healthy) pointed out that more accurate predictions can be achieved using symptoms metadata than cough features, while the overall performance increase by combining both data sources.
As it may be expected, attention distribution highlight fever, cough, dizziness, and chest pain as the most recurrent symptoms for infected subjects.
In addition, the authors perform an in-depth clinical analysis of the main significant differences in the energy distribution of the cough spectrum between COVID-$19$ and other cough types. Overall, results confirm that the model is able to learn the main relationships between the frequency distribution of the most discriminating features and the underlying cough sound characteristics for each class. These findings are also supported by t-distributed Stochastic Neighbor Embedding (t-SNE) \cite{van2008visualizing} visualisation, showing a clear separation between the four clusters of cough features learned by the model.
\\
Several solutions based on signal processing methods have also been proposed to explain the detection of heart disorders, which are mainly based on the analysis of ECG recordings.
\cite{ivaturi2021comprehensive} presented a XAI framework for AF prediction from single-lead ECG signals.
To this aim, they first trained a \textit{MobileNet} \cite{howard2017mobilenets} CNN architecture on the PhysioNet/Computing in Cardiology Challenge $2017$ dataset \cite{clifford2017af}.
Then, global explanation analysis has been performed by dividing each ECG cycle into $8$ equally size segments, each one corresponding approximately to a region of interest (e.g., P wave, T wave, isoelectric baseline), and applying feature ablation, feature permutation, and LIME methods to highlight the most relevant segments. 
Moreover, saliency maps with guided back-propagation technique have also been used to compare the direct contribution of raw input data to local predictions with global, segment-based analysis.
Clinical analysis of explanations shows that the network effectively focuses on physiological features that match with those used by cardiologists for the clinical AF diagnosis, such as the absence of P-wave, variability of R-R intervals, and electrical activity in the isoelectric region of the ECG.
\\
\cite{mousavi2020han} proposed \textit{HAN-ECG}, an alternative solution for explaining AF predictions from single-lead ECG.
This system differs from \cite{ivaturi2021comprehensive} as it relies on a stacked Bi-LSTM ensemble with a hierarchical attention mechanism to learn relevant components of the input signal at different levels, namely beat, wave, and time windows, respectively.
The network has been evaluated over two datasets, including the PhysioNet $2017$ and the MIT-BIH AFIB\footnote{https://physionet.org/content/afdb/1.0.0/\label{afib}}, then the visualisation of attention layers has been exploited to demonstrate that the model focuses on clinically relevant heart beats and waves for detecting AF arrhythmia.
As in \cite{ivaturi2021comprehensive}, the absence of P-waves, which may be occasionally replaced with a series of small waves called fibrillation waves, and the irregular R-R intervals in which the heartbeat intervals are not rhythmic played an essential role in AF detection.
\\
\cite{dissanayake2020robust} developed an interpretable DL framework for heart anomaly detection from Mel-Frequency Cepstral
Coefficient (MFCC) spectral features \cite{clifford2017recent} extracted from phonocardiogram (PCG) signals (i.e., heart sounds).
The framework combines a pre-trained LSTM network for automatic segmentation of the input MFCC maps, a CNN encoder to perform spatial feature learning on the supplied feature map, and a MLP network to get the final prediction.
Different network architectures have been tested through the combination of the above modules in order to explicitly examine the importance of signal segmentation as a prior step to classification.
Then, both SHAP and occlusion maps have been used to explain the hidden representations learned by the model.
Experimental results obtained on the benchmark PhysioNet database \cite{goldberger2000physiobank} indicate that the network architecture without segmentation module reaches the highest accuracy, outperforming the state-of-the-art methods.
In addition, both XAI methods show that a correct classification of PCG signals occur when the model mainly focuses on learned features that are placed within (or between) two fundamental heart sound locations, namely S$1$ and S$2$ segments. This model behaviour accords with the clinical assessment followed in digital phonocardiography.
For what concerns the role of signal segmentation, these findings suggest that if the model is robust enough to learn the segmentation function while extracting associated features from S$1$ and S$2$ locations, then signal segmentation may not be necessary as preliminary data processing.
\\
\\
Eventually, XAI is also gaining increasing attention in the field of Human Activity Recognition (HAR), based both on wearable sensing and smart home environments \cite{arrotta2022dexar}. Most HAR applications are related to human well-being and fitness through physical activity monitoring, as well as to active and healthy aging by supporting older and impaired subjects in the correct execution of daily activities in their home environment, and/or by detecting abnormal behavioural and locomotion patterns \cite{khodabandehloo2021healthxai}.
However, using XAI methods in these applications is currently limited to explain why (and how) simple/complex activities are detected, so the impact of explainability methods on decision making may be limited.
Differently, the contribution of XAI to HAR applications in the healthcare domain is much higher, as it provides evidences for the final diagnosis and justifications for successive interventions, with particular reference to gait analysis to detect orthopedic or neurodegenerative disorders, such as Parkinson Disease (PD), and/or fall detection. 
Such applications are often based on a vast stream of inertial and/or kinematic data, which are in current need of interpretability in order to detect which signal characteristics are effectively used by AI models to take meaningful decisions.
To this aim, \cite{filtjens2021modelling} proposed an interpretable DL framework to detect movements preceding the occurrence of Freeze of Gait (FoG) episodes in PD patients.
The framework is based on a CNN to model the reduction of movement prior to a FOG episode from $3$-D kinematic joint trajectories of hip, knee, and ankle, respectively, and on LRP explanatory technique to identify the most influential features.
The model has been built by using an existing dataset \cite{spildooren2010freezing} containing $3$-D motion data from $28$ PD patients with and without FoG, and $14$ healthy subjects.
LRP interpretability analysis indicated that the movements perceived as the most relevant by the model are fixed knee extension during the stance phase, reduced peak knee flexion during wing phase, and the fixed ankle dorsiflexion during the wing phase.
On the other hand, very little relevance has been observed for these movements in PD patients without FOG and in healthy controls.
Therefore, this behaviour suggests that model decisions are made upon meaningful kinematic features, which are actually related to movement reductions during gait.

\begin{table}[!htbp]
\begin{minipage}{\textwidth}
\caption{List of XAI studies with tabular data performing clinical validation of explanations.}\label{res_tab_3}
\begin{tabular*}{\textwidth}[l]{p{.05\textwidth} p{.05\textwidth}p{.15\textwidth}p{.1\textwidth}p{.15\textwidth}p{.15\textwidth}p{.15\textwidth}}
\toprule
\textbf{Ref.} & \textbf{\# Cit.}&  \textbf{Application} & \textbf{Input Data} & \textbf{AI model(s)} & 
\textbf{XAI method(s)} &\textbf{Dataset(s)} \\
\midrule
\cite{beebe2021efficient} ($2021$)& $7$ &Imminent dementia diagnosis& EHR& XGBoost&SHAP& ROSMAP dataset \cite{a2012overview}\\
\midrule
\cite{sha2021smile} ($2021$)&$4$&AD diagnosis&Plasma metabolite concentrations& Evolutionary algorithms& LGP& AD metabolomic dataset \cite{wang2014plasma}\\
\midrule
\cite{kim2021interpretable} ($2021$)&$3$& END detection in stroke patients& EHR& LightGBM& SHAP&K-Attention dataset \cite{jung2019long}\\
\midrule
\cite{rashed2021clinically} ($2021$)&$6$ &CKD diagnosis& EHR& RF, GBDT, XGBoost& SHAP&UCI CKD dataset \cite{rubini2015uci}\\
\midrule
\cite{pang2019understanding} ($2019$)&$10$&Early childhood obesity prediction&EHR&XGBoost&SHAP& Retrospective study\\
\midrule
\cite{zeng2021explainable} ($2021$)&$5$&Post-operative complication risk prediction&EHR&XGBoost&SHAP&Retrospective study\\
\midrule
\cite{zhang2021explainable} ($2021$)&$10$&post-LT AKI prediction&EHR&RF&SHAP&Retrospective study\\
\midrule
\cite{lu2020explainable} ($2020$)&$5$&COVID-$19$ diagnosis and prognosis&EHR&GBDT, RF&SHAP&Retrospective study\\
\midrule
\cite{gupta2021interpretable} ($2021$)&$1$&COVID-$19$ recovered subject detection& HR/HRV features& XGBoost&SHAP& Retrospective study\\
\midrule
\cite{pal2021pay} ($2021$)&$50$&COVID-$19$ diagnosis&Cough sounds, symptoms metadata& TabNet, DNN& Attention&Pilot study\\
\midrule
\cite{dissanayake2020robust} ($2020$)&$18$&Heart anomaly detection& PCG features&stacked LSTM-CNN-MLP network& SHAP, occlusion maps&PhysioNet DB \cite{goldberger2000physiobank}\\
\bottomrule
\end{tabular*}
\end{minipage}
\end{table}

\begin{table}[!htbp]
\begin{minipage}{\textwidth}
\caption{List of XAI studies with time series data performing clinical validation of explanations}\label{res_tab_4}
\begin{tabular*}{\textwidth}[l]{p{.05\textwidth} p{.05\textwidth}p{.125\textwidth}p{.15\textwidth}p{.15\textwidth}p{.125\textwidth}p{.15\textwidth}}
\toprule
\textbf{Ref.} &\textbf{\# Cit.}& \textbf{Application} & \textbf{Input Data} & \textbf{AI model(s)} & 
\textbf{XAI method(s)} &\textbf{Dataset(s)} \\
\midrule
\cite{shashikumar2021deepaise} ($2021$)&$15$& Sepsis prediction& Longitudinal EHR& WCPH-RNN& Saliency & Retrospective study\\
\midrule
\cite{sun2021attention} ($2021$)&$1$&Survival prediction&Longitudinal EHR& RNN, GNN & Attention& WHAS \cite{lemeshow2011applied}, SUPPORT \cite{knaus1995support}, METABRIC \cite{curtis2012genomic}, MIMIC-III, PLAGH\\
\midrule
\cite{zheng2020tracer} ($2020$)&$9$& Mortality prediction, AKI prediction& Longitudinal EHR& TITV network& FiLM, attention& NUH-AKI, MIMIC-III\\
\midrule
\cite{kwon2018retainvis} ($2018$)&$196$&HF diagnosis&Longitudinal EHR& RETAIN variant&Attention& HIRA-NPS \cite{kim2014guide}\\
\midrule
\cite{ivaturi2021comprehensive} ($2021$)&$4$&AF detection& ECG& CNN& Feature ablation/permutation, LIME, guided saliency& PhysioNet $2017$ \cite{clifford2017af}\\
\midrule
\cite{mousavi2020han} ($2020$)&$42$&AF detection& ECG &Bi-LSTM ensemble& Attention& PhysioNet $2017$, MIT-BIH AFIB\footref{afib}\\
\midrule
\cite{filtjens2021modelling} ($2021$)&$3$&pre-FoG movement detection in PD patients& $3$-D kinematic joint trajectories&CNN&LRP& Gait dataset \cite{spildooren2010freezing}\\
\bottomrule
\end{tabular*}
\end{minipage}
\end{table}

\subsection{Explanation consistency assessment}
The evaluation of the level of agreement between explanations generated by different methods is often used by researchers to get some preliminary insights into stability and robustness of AI models.
These properties are often used interchangeably as they both refer to the model ability to withstand perturbations introduced in input data, even if a slight difference among the two concepts exists.
Indeed, stability is evaluated with respect to unintentional perturbations that may occur in the real world, such as data noise, while robustness refers to subtle yet intentional changes in input data, namely adversarial attacks.
\\
Given a target high-performing model, if similar explanations are generated by different methods, then the model should provide correct outcomes for the same reasons for equal or similar data instances over the time. On the other hand, similar explanations obtained by multiple models with the same explanatory technique(s) might indicate that common patterns are discovered within data and used to make decisions.
As a result, the model should be able to deal with both random changes and adversarial examples without leading to systematic misclassification.
However, more specific XAI approaches, such as sensitivity analysis, should be required to draw more accurate conclusions about these model desiderata.
\\
From a practical standpoint, consistency assessment applies to methods providing the same explanation format, generally by matching rankings coming from feature attribution methods or by evaluating the degree of overlap between different decision rule sets.
These methods are mainly applied to tabular data, while saliency and attention remain the benchmark methods to explain models learned from time series. 
Moreover, there are also other motivations limiting explanation consistency assessment to time series.
First, saliency methods mainly work for local predictions, so evaluating explanation consistency may not provide any global understanding of models, unless a huge number of data instances is analysed individually.
Moreover, the usefulness of comparing maps obtained by different methods can be questionable if end users have difficulties in understanding the high-level content hidden in the input sub-sequences showing the highest relevance. As a result, this section focuses on the most relevant XAI studies targeting tabular data and performing explanation consistency assessment.
Research works are also summarised in Table \ref{res_tab_5}.
\\
\cite{thimoteo2022explainable} compared post-hoc explanations for COVID-$19$ diagnosis with other glass-box AI approaches.
Specifically, they applied SHAP to SVM and RF classifiers trained on lab test data provided by the COVID-$19$ Data Sharing/BR (over $50$k suspected COVID-$19$ patients), then they compared feature relevance with both LR coefficients and feature importance scores provided by Explainable Boosting Machine (EBM) algorithm \cite{nori2019interpretml}.
All global explanations converged in indicating eosinophils and leukocytes, and in general white blood cell-related parameters among the essential features to help diagnose the infection from blood test and pathogen variables.
\\
\cite{alves2021explaining} proposed a \textit{Decision Tree-based eXplainer} (DTX) and applied it for COVID-$19$ diagnosis from lab test (i.e., blood, urine, and others) data.
This approach produces readable tree structure that provides classification rules to reflect the local behavior of complex models, and it can be considered similar to LIME method using decision trees as surrogate models instead of sparse linear models. 
In addition, DTX outcomes have been aggregated over many patients for the identification of global patterns, named criteria graphs.
DTX rules and criteria graphs have been extracted from a RF classifier trained on the same public COVID-$19$ dataset used in \cite{leung2021explainable}, then they have been compared with SHAP and LIME explanations at global and local stages, respectively.
Results showed a high level of overlap of the proposed method with respect to these well-established XAI techniques, in particular highlighting a correspondence between the $5$ largest nodes in the graph and the top-$5$ features in SHAP ranking.
\\ 
\cite{okay2021interpretable} developed an interpretable ML approach for early stage diabetes detection from sign and symptom data, which can be easily collected through patient questionnaires. 
They first trained RF and GBDT models on the Sylhet Diabetes dataset\footnote{https://archive.ics.uci.edu/ml/datasets/Early+stage+diabetes+risk+prediction+dataset.\label{sylhet}} ($520$ instances, $17$ categorical features), then they applied SHAP to compare global explanations.
Results indicated that the top-$3$ features (polyuria, polydipsia, and gender) are shared across the two models using SHAP, with also a high degree of overlap for the top-$10$ attributes.
LIME also provided similar feature rankings among RF and GBDT for the selected local predictions, but this is not enough to assert the convergence of its global explanations. 
\cite{oba2021interpretable} performed a similar study to analyse explanations related to diabetes aggravation detection from medical records that integrate patient interviews with lab tests and physiological measurements.
They used a medical check-up dataset collected by a Japanese health care center to train different tree ensemble models (i.e., XGBoost, LightGBM, and CatBoost) and a TabNet, then they compared SHAP values obtained from the former with attention weights generated by the network.
In this case, the top-$3$ features ranked by SHAP, namely current severity status of diabetes, blood sugar level, and glycated hemoglobin, were the same among all the models, and also equal to those learned by attentive transformers.
In addition, results obtained by TabNet pointed out many highly-ranked indicators that can be obtained by non-invasive tests and interviews, which are less burdensome and expensive to patients.
\\ 
\cite{elshawi2019interpretability} performed an extensive analysis for investigating the outcomes of ML models for hypertension prediction from cardio-respiratory fitness data obtained after a treadmill test.
The authors exploited data from $>23k$ patients and models coming from the FIT project \cite{sakr2018using}, then they selected the best performing RF classifier to compare a variety of XAI methods, including feature permutation, PDP, ICE plots, feature interaction with Friedman's H statistic \cite{friedman2008predictive}, and surrogate models for global explanations, as well as LIME and SHAP for local explanations.
Results obtained by this experiment suggested that integrating different global interpretations may allow to generalise the overall conditional distribution modeled by the trained response function, but local interpretations should be preferred for a better understanding of smaller variations in the conditional distribution for specific instances.
\\ 
Another automated and interpretable diagnostic application has been proposed by \cite{seedat2020automated} for voice pathology assessment from smartphone-based microphone recordings.
To this aim, they conducted a pilot study to collect and analyse voice recordings obtained from $33$ healthy and diseased subjects, then they trained several ML models using a set of handcrafted features extracted through audio signal processing.
By choosing ExtraTrees as the best performing model, they compared global explanations obtained through SHAP, Morris sensitivity analysis, and feature permutation.
All methods converged in identifying the most relevant features, and they also highlighted $6$ clinically used MFCC features as the top relevant ones.
\\ 
Assessing stability should be imperative for predictive models that are supposed to be deployed for survival analysis.
\cite{kapcia2021exmed} proposed \textit{ExMed}, a tool to enable XAI data analytics and visualisation for clinicians by supporting multiple ML models and feature attribution algorithms, and they tested it using lung cancer life expectancy prediction from EHR as main application.
By exploiting the public Simulacrum dataset\footnote{https://simulacrum.healthdatainsight.org.uk/\label{simulacrum}}, a cancer dataset provided by the National Cancer Registration and Analysis Service of Public Health England, they trained different ML models and then selected the best RF classifier to compare SHAP values with the global average of LIME local scores obtained over all the test patients.
This analysis show a very similar impact of almost all $20$ patient features used.
In particular, cancer grade and M-best (i.e., presence/absence of distant metastatic spread) are the two most relevant features with very close importance scores (also in line with the current clinical knowledge), while there is a disagreement on age attribution.
By using the same dataset, \cite{duell2021comparison}
presented an extended comparison of explanations obtained for lung cancer survival prediction. Specifically, they compared SHAP, LIME, and anchors methods applied to a XGBoost model, as well as the feature importance ranking derived from an EBM model.
Overall, all methods converge in identifying M-best as the most relevant feature globally, while the ranking for the remaining features differs between SHAP and LIME. For instance SHAP consider N-best (i.e., extent of involvement of regional lymph nodes) as the second most important feature, whereas LIME considers the behaviour of the tumour.
Given such discrepancy between LIME and SHAP, the authors studied the scale of their differences by analysing the first $1000$ instances of the test set individually to identify priority features, regardless of whether they are shared or not for every instance between the two methods.
Results highlight M-best, N-best, and T-best (i.e, size and extent of the primary tumor) as the three most important features, further supporting both consistency and clinical relevance of knowledge
representation.
\\
Similarly, \cite{moncada2021explainable} compared the performance of a conventional WCPH regression model against three different ML methods, namely Random Survival Forest \cite{ishwaran2008random}, Survival SVM \cite{polsterl2016efficient}, and XGBoost, for breast cancer survival prediction from patient, tumor, and treatment-related characteristics.
Models have been trained on a dataset built through a retrospective data collection from the Netherlands Cancer Registry between $2005$ and $2008$, then SHAP has been applied to investigate the differences between a reference WCPH model and the best performing XGBoost model.
This comparison resulted in a high degree of overlap between global explanations across the models, while XGBoost reached a considerably better accuracy. Therefore, this increase in model performance may be attributed to XGBoost’s ability to model non-linearities and complex interactions among input variables with respect to a simpler semi-parametric approach.
\\
\cite{ang2021interpretable} performed a study to compare the most salient features for ICU mortality risk prediction with different ML models, all of which have been trained on the benchmark MIMIC-III database. Specifically, the authors compared SHAP values obtained from MLP and RF classifiers, as well as Logistic Regression (LR) coefficients and Gini's feature importance scores derived from a decision tree.
Obtained results highlighted a high degree of commonality for age, blood urea nitrogen level, and patient dischargement from cardiac surgery as main mortality risk factors in ICU settings.
\\
\cite{song2020cross} proposed an alternative perspective for analysing explanation consistency, aimed at cross-site validation of an AKI prediction model from multi-center EHR data.
To this aim, they developed a GBDT classifier using a huge amount of input features collected from patients enrolled in the Greater Plain Collaborative \cite{waitman2014greater}, a research network including twelve healthcare systems in the US, including demographics, diagnoses, procedures, lab tests, medications, as well as vital signs.
To evaluate this approach, a reference site has been chosen for model training and internal validation, while data coming from other $6$ clinical sites have been used for external validation. 
System transportability has been assessed through the definition of a novel metric, the adjusted Maximum Mean Discrepancy (adjMMD), to infer performance deterioration (i.e., drop in AUC) between transported and refitted (i.e., retrained) models at each site.
adjMMD is a modified version of the traditional MMD metric, which is widely used in transfer learning for measuring the data distribution shift between source and target data \cite{pan2010domain}.
SHAP analysis has also been performed to investigate changes in the marginal effects of top-$10$ features in predicting moderate-to-severe AKI.
This work demonstrated that cross-site performance deterioration is likely and it is generally associated with significant disparities in feature importance, which may be caused by heterogeneity of risk factors across different populations.
As a result, a joint analysis of model explanations and performance could be used to estimate the transportability of AI models, which in turn can foster the adaptation process across different clinical settings (e.g., health institutions) and/or patient cohorts.
\\
In addition to cross-site transportability, another fundamental issue related to the deployment of predictive models in clinical settings is the possible occurrence of both data and concept drift over the time.
The former is a systematic shift in the underlying distributions of input parameters, while the latter represents a substantial change in the relationship between inputs and target clinical outcomes.
Both of them represent threats to model validity; therefore, a periodical monitoring of explanation consistency (together with predictive performance) may be used as model temporal validation, with particular reference to data drift detection. 
To the best of our knowledge, \cite{duckworth2021using} is the first study suggesting the analysis of explanations over the time to measure data drift, by exploiting COVID-$19$ as an exemplary case for monitoring hospital readmission risk prediction models.
Specifically, this work analyses attendance records of all adults to Southampton General Hospital’s ED between $2019$ and $2020$, including patient descriptors medical history, attendance characteristics, as well as any vital sign measured at the point of triage to perform temporal validation.
Specifically, a XGBoost model has been trained only on pre-pandemic attendances, then it has been evaluated in weekly bins over a test period ranging from October $2020$ to May $2021$ in terms of predictive performance (i.e, AUC) and explanations (i.e., average SHAP values).
Results outline a clear drop in model performance starting from March $2020$ (i.e., onset of COVID-$19$), which are also associated with significant changes in the importance of several features.
In particular, symptoms data that we now know to be common in infected subjects, such as shortness of breath and chest pain, show a higher relevance starting from COVID-$19$ spread.
Similarly, respiration rate, BT, and SpO$_2$ present peaks in importance scores starting from March $2020$, and also in December $2019$ (potentially reflecting the flu season).
These outcomes demonstrate that short-time and unpredictable changes in the data distribution, for instance related to a new emergent disease, can negatively affect model outcomes.
Therefore, periodical analysis of explanations may act as proxy measure of data drift, with the final aim to timely plan model retraining/updating interventions.
\\ 
To conclude this section, we report additional innovative health applications characterised by a lack of benchmark AI solutions, which have been also less investigated from a clinical standpoint. Here, explanation consistency assessment is used for preliminary model validation beyond predictive performance analysis.
\cite{tahmassebi2020interpretable} compared SHAP explanations for XGBoost and DNN models for eye state detection from multi-channel Electroencephalogram (EEG) signals.
The authors conducted a pilot study to build a dataset of approximately $15$K EEG recordings, then they trained both classifiers by considering each of the $14$ employed channels as a separate input feature in order to investigate the activation of the corresponding brain regions.
Results indicated that the top-$3$ important features are shared across models with slight variations in their ranking; however, the greatest non-linearity of DNN also provided higher-order interactions among features with respect to a XGBoost model with shallow trees as base learners, which in turn led to a higher contribution of low-impact regions for the final prediction.
\\
\cite{antoniadi2021prediction} proposed the first XAI framework for Quality of Life (QoL) assessment in Amyotrophic Lateral Sclerosis (ALS) patient caregiving from a heterogeneous set of EHR data, which include demographics, usage of health service, questionnaire data from both patients and their caregivers, as well as caregiving duties and patients' clinical attributes.
Data have been collected by $90$ ALS subjects (and their caregivers) who attended the National ALS/MND multidisciplinary Clinic at Beaumont Hospital, Dublin, then they have been used to compare performance and explanations of a XGBoost model with those of a LASSO-LR reference model, due to the absence of state-of-the-art methods for the target application.
In addition, consistency check of the top ranked features by SHAP has been performed across several XGBoost model configurations, each one built with a smaller feature subset chosen through ensemble feature selection. 
This analysis is geared towards the creation of a human-centered DSS to alert clinicians about worsening QoL conditions in ALS patient caregiving. In addition, the system should be able to reveal and explain the main risk factors within a limited set of data that can be easily and routinely collected by both patients and caregivers, in order to achieve a better usability and assessment.
\\
\cite{ward2021explainable} proposed a XAI approach for pharmacovigilance monitoring by analysing the impact of patient features related to drug history, comorbidities, and current drug dispensing in developing adverse health conditions.
To this aim, they collected data from patients aged $>65$yo from the Western Australian Department of Health, by considering acute coronary syndrome as adverse outcome to be predicted with different tree ensemble models (i.e., RF, XGBoost, and ExtraTrees).
Then, they compared feature importance scores obtained by different XAI techniques, namely LIME, SHAP, as well as MDI and MDA feature permutation.
To reduce cross-model variance, the authors first compared the average feature importance scores across all models, then they performed a per-model analysis of top ranked features across explanatory methods.
Overall results indicated that sex and age are ranked highly by all methods, an expected outcome as these are the most important auxiliary inputs in clinical/epidemiological studies.
Results also outlined repeating peaks of importance for all XAI methods falling within musculo-skeletal and cardio-vascular drug classes.
For what concerns per-model rankings, results indicated that highly ranked comorbidity and drug dispensing features are shared among models under SHAP and MDA analysis, while LIME outcomes are inconsistent, which may be due to a higher variance introduced by averaging LIME local explanations to obtain global patterns.
Finally, random integer features have been also included into the models as confounding factors, which should be ranked last as they cannot have any predictive power.
Nevertheless, MDI ranked some of them as relevant, unveiling a potential bias in this method as variables with high cardinality may be selected for many tree splits, thus appearing important to the model.

\begin{table}[!htbp]
\begin{minipage}{\textwidth}
\caption{List of XAI studies performing explanation consistency assessment.}\label{res_tab_5}
\begin{tabular*}{\textwidth}[l]{p{.05\textwidth}p{.05\textwidth} p{.2\textwidth}p{.1\textwidth}p{.1\textwidth}p{.15\textwidth}p{.15\textwidth}}
\toprule
\textbf{Ref.} & \textbf{\# Cit.}&\textbf{Application} & \textbf{Input Data} & \textbf{AI model(s)} & 
\textbf{XAI method(s)} &\textbf{Dataset(s)} \\
\midrule 
\cite{thimoteo2022explainable} ($2021$)&$2$&COVID-$19$ diagnosis& EHR& SVM, RF& SHAP& COVID-$19$ Data Sharing/BR\\
\midrule
\cite{alves2021explaining} ($2021$)&$36$&COVID-$19$ diagnosis&EHR& RF& DTX, Criteria graph, LIME, SHAP& COVID-$19$ dataset\footref{covid}\\
\midrule
\cite{okay2021interpretable} ($2021$)&$1$&Diabetes diagnosis& EHR& RF, GBDT& SHAP, LIME&Sylhet Diabetes dataset\footref{sylhet}\\
\midrule
\cite{oba2021interpretable} ($2021$)&$1$&Diabetes diagnosis&EHR& TabNet, XGBoost, LightGBM, CatBoost& SHAP (all), attention (TabNet)& Retrospective study\\
\midrule
\cite{elshawi2019interpretability} ($2019$)&$132$& Hypertension prediction& EHR& RF & feature permutation, PDP, ICE, global surrogate models, LIME, SHAP& Pilot study\\
\midrule
\cite{seedat2020automated} ($2020$)&$1$& Voice pathology assessment& Audio features&ExtraTrees& SHAP,Morris sensitivity analysis& Pilot study\\
\midrule
\cite{kapcia2021exmed} ($2021$)&$0$&Lung cancer life expectancy prediction&EHR&RF&LIME, SHAP& Simulacrum dataset\footref{simulacrum}\\
\midrule
\cite{duell2021comparison} ($2021$)&$9$&Lung cancer mortality prediction&EHR& XGBoost& LIME, SHAP, Anchors&Simulacrum dataset\\
\midrule
\cite{moncada2021explainable} ($2021$)&$43$&Breast cancer survival prediction& EHR& XGBoost&SHAP& Retrospective study\\
\midrule
\cite{ang2021interpretable} ($2021$)&$0$&ICU mortality risk prediction& EHR& RF, MLP& SHAP& MIMIC-III\\
\midrule
\cite{song2020cross} ($2020$)&$33$&AKI prediction&EHR&GBDT&SHAP& Retrospective study\\
\midrule
\cite{duckworth2021using} ($2021$)&$5$&Hospital readmission prediction&EHR&XGBoost&SHAP&Retrospective study\\
\midrule
\cite{tahmassebi2020interpretable} ($2020$)&$3$&Eye state detection&EEG & XGBoost, DNN& SHAP& Pilot study\\
\midrule
\cite{antoniadi2021prediction} ($2021$)&$7$&QoL assessment in ALS caregiving&EHR&XGBoost &SHAP&Retrospective study\\
\midrule
\cite{ward2021explainable} ($2021$)&$5$&Pharmaco-vigilance monitoring&EHR&RF, XGBoost, ExtraTrees& MDA, MDI, LIME, SHAP& Retrospective study\\
\bottomrule
\end{tabular*}
\end{minipage}
\end{table}

\subsection{Explanation quality assessment}\label{quality}
Clinical validation and consistency assessment are valuable but indirect ways to estimate the quality and relevance of the proposed explanations from different perspectives.
However, there are not well-established and globally recognised metrics or practices that enable a formal assessment and a systematic comparison of methods \cite{markus2021role}.
To fill this gap, quantitative metrics have been recently proposed for an initial and objective evaluation. However, their applicability is often limited to the target task or to the format of explanations generated (e.g., saliency maps, decision rules), while a standardised framework is still missing.
\\
Quantitative evaluations should be followed by human-centered assessment made by healthcare professionals through surveys, feedback, and ratings. 
Moreover, the collaboration with domain experts is also necessary for the design and development of novel analytics and visualisation tools that effectively incorporate visual/textual explanations of AI models into clinical decision making workflow, also highlighting which interactions with end users are needed to improve the interpretation of machine-based predictions.
Given the limited number of research works found in the literature, studies targeting tabular and time series data are grouped together in Table \ref{res_tab_6}.
\\
\cite{cheng2021vbridge} developed \textit{VBridge}, a novel visual analytics tool to address three key challenges related to XAI adoption in healthcare, namely clinicians’ unfamiliarity with ML features, lack of contextual information, and need for cohort-level evidence.
To this aim, the system first provides a hierarchical display of attribution-based feature explanations, by grouping the most relevant features semantically for a better analysis.
In addition, it includes enriched interactions to link relevant features with raw data (both static and time series), reference values from desired patient clusters, with the ultimate goal of providing a complete overview of each patient.
As running example for their study, they trained several ML models on the Pediatric Intensive Care (PIC) database \cite{zeng2020pic} for surgical risk prediction in pediatric patients, then they selected SHAP for feature explanation.
VBridge has been evaluated through interviews with four expert clinicians, by applying both forward and backward data analysis as separate case studies.
In the first case, clinicians inspect data following the data processing flow (i.e., from original EHR, to features, predictions and explanations), which is also similar to the current clinical practice; in the second case, the order is reversed by starting with the output of AI models.
Feedback obtained from both case studies demonstrate that visually associating model explanations with patients’ situational records can help clinicians to better interpret and use predictions for decision-making.
\\
\cite{kumarakulasinghe2020evaluating} proposed a framework to evaluate clinical relevance and quality of LIME local explanations, by using a RF classifier trained on the PhysioNet/Computing in Cardiology Challenge $2019$ dataset \cite{reyna2019early} for sepsis prediction in ICU patients as case study.
The framework enables a multidimensional and semi-quantitative evaluation. It first evaluates the percentage of model explanations accepted by clinicians and the percentage of overlap between LIME and clinicians for the top-$K$ ranked features (both general match and exact ranking match), then it includes a survey adapted from \cite{jian1998towards} for a qualitative assessment of trust and reliance on AI outcomes.
Additionally, physicians were also asked to report their level of satisfaction with respect to the textual and visual representation of LIME explanations.
As a first pilot study, it has been used by $10$ clinicians, each of them should have inspected $10$ separate local predictions to reach a total of $100$ cases. Obtained results indicated that LIME explanations have a very high level of agreement with those provided by physicians, with also trust and reliance scores fairly high.
\\
\cite{barda2020qualitative} conducted a multidisciplinary research study aimed at designing effective user-centered displays of ML explanations for healthcare applications.
To this aim, the first trained several ML models for in-hospital mortality risk prediction by using EHR data obtained from ICU pediatric patients admitted at Children’s Hospital of Pittsburgh between $2015$ and $2016$, then they applied SHAP to the best performing RF classifier.
SHAP explanations have been analysed through focus group session attended by a total of $21$ clinicians, in which different design options and rationales have been proposed for explanation presentation, such as display formats, risk representation formats, dimensionality, and many others.
Therefore, focus group sessions provided critical reviews and highlighted users' preferences, which in turn have been used to define a final version for explanations' content and display.
Specifically, the preferred visualisation tool integrates additional information, such as raw feature tables, time series plots, and contextual factors, to help a better understanding of model explanations.
In addition, feedback from focus group participants positively support the adoption of model-agnostic, instance-level explanations based on feature relevance (such as those obtained by SHAP) to understand AI predictions in healthcare.
\\
Explanation quality assessment may also be applied to novel XAI algorithms in order to compare them with benchmark techniques.
\cite{penafiel2020predicting} proposed a novel interpretable classifier to make predictions from incomplete/missing EHR data.
The proposed approach is able to integrate expert knowledge into the learning process through the a priori definition of rules by medical users, then it exploits the Dempster-Shafer (DS) mathematical framework \cite{shafer2016dempster} to reason with data uncertainty, along with a GD optimisation to derive a sub-optimal set of the best performing decision rules.
The DS-GD classifier has been trained on EHR data provided by the regional hospital in Okayama, Japan ($28$k records) to perform stroke prediction, providing superior performance with respect to the state-of-the-art ML models and clinical stroke assessment methods.
Moreover, the authors conducted an expert survey by presenting to different neurologists the global rules learned by the model that are associated with an increased stroke risk, including also less contributory and random rules to prevent biased answers.
For each rule, clinicians were asked to report whether they consider it true, false, or if there is no correlation. Results obtained by analysing the percentage of clinical adherence confirmed that almost all the rules discovered by the model are also the most accepted statements among experts.
\\
\cite{hatwell2020ada} proposed a novel approach to explain the AdaBoost (AB) classification algorithm, a black-box model widely used in the CAD literature.
The new algorithm, named Adaptive-Weighted High Importance Path Snippets (Ada-WHIPS), has been designed to provide accurate and highly-interpretable disease detection by extracting simple and logical rules from the AB model.
Specifically, Ada-WHIPS assigns unique weights among individual branch nodes of each tree learner within the ensemble, then it performs a simple heuristic search over the weighted nodes to find out a single rule that dominates the final decision.
This XAI algorithm has been compared with two state-of-the-art, rule-based explanatory techniques, namely Anchors and LORE, by conducting separate experiments over $9$ different CAD datasets (Cleveland, Breast Cancer, Thyroid, etc...) available at the UCI ML repository.
Explanation assessment has been conducted through non-parametric hypothesis tests in order to find significant differences in $4$ quality metrics, namely efficiency, precision, coverage, and stability.
Efficiency just refers to the computational time required to derive an explanation, whereas coverage measures the amount of test instances that could be predicted after looking at a given rule (i.e., rule generalisability).
On the other hand, precision represents the fraction of correct predictions that can be made by applying the rule (i.e., rule specificity). 
Finally, stability is a novel metric introduced by the authors for optimising rule-based explanation algorithms, as it represents a regularised version of precision including a trade-off with coverage to explicitly avoid over-fitting, thus preventing tautological rules that correctly applies only to a very limited set of instances.
Ada-WHIPS explanations result in better generalisation capabilities in a two-way comparison with Anchors for all datasets, while significant differences are less consistent in case of a three-way comparison.
Moreover, Ada-WHIPS remained competitive in terms of specificity with respect to state-of-the-art, also providing a comparable computational time.
\\
Explanation quality assessment is even more challenging in the time series field, as there are no gold standard metrics applicable to signal processing methods that can offer a global and objective evaluation of saliency/attention mechanisms generally used for explaining DNN models.
As a result, most of existing approaches are tied to the availability of extensive expert annotations for the comparison of local saliency with human ground truth over a significant number of data instances.
\cite{zhang2021explainability} proposed novel evaluation metrics for the validation of saliency maps with respect to expert annotations, which can be used for rigorous comparisons between DNN models.
Specifically, they include \textit{Congruence}, which measures the proportion of model saliency within the expert annotations, and \textit{Annotation Classification}, which determines the proportion of expert annotations covered by the saliency maps (i.e., the spread of model saliency).
To apply these metrics, the authors developed a \textit{ResNet} \cite{he2016deep} CNN architecture for photoplethysmography (PPG) signal/image quality assessment (i.e., normal VS. abnormal).
The network has been trained using PPG data from an ICU dataset used in \cite{drew2014insights} and tested over PPG data from a stroke dataset collected in \cite{pereira2019deep}.
Model saliency has been visualised for the whole test set using different methods, namely DeepSHAP, Integrated Gradients, and guided back-propagation, then it has been compared with manual expert annotations.
Obtained results demonstrate that signal-based models act in a more explainable fashion with respect to image-based ones, with guided saliency outperforming the other techniques in both evaluation metrics.
The comparison with model classification outcomes also highlights a weak-to-moderate positive correlation between explainability and performance metrics, such as accuracy and specificity, suggesting that the higher the saliency correctly allocated with respect to human ground truth, the higher the model performance.
\\
\cite{wickstrom2020uncertainty} proposed additional metrics to specifically address the lack of uncertainty measures of XAI techniques, with particular reference to saliency methods applied to DNN.
To this aim, they first developed a Fully Convolutional Network (FCN) ensemble, in which each CNN is trained independently. Then, uncertainty in the relevance scores has been estimated by taking the standard deviation across importance scores produced by each CNN through CAM, which in turn has been used to threshold the time steps that most models agree on to make explanations more reliable.
The proposed model has been tested for AMI detection from echocardiogram signals using the ECG$200$ dataset \cite{dau2019ucr}, and for Surgical Site Infection (SSI) detection from longitudinal blood measurements of CRP using a EHR dataset collected in \cite{mikalsen2016learning}.
To evaluate model saliency, \textit{Relevance Accuracy} metric has been computed to determine the amount of actually relevant data points within a sample time series that are recognised by the model, similarly to the congruence metric defined in \cite{zhang2021explainability}.
In addition, \textit{Relevance Consistency} metric has been also introduced by the authors to estimate how the ensemble correctly detects the most relevant time points across different network initialisation with respect to a single FCN model.
Results demonstrate that the proposed FCN model is accurate in locating relevant time points in both case studies, and it is also more consistent as it indicates mostly the same time steps as relevant for its predictions when trained with different random initialisation.
\\
\cite{slijepcevic2021explaining} proposed a complementary strategy to evaluate class-specific explanations for gait classification from $3$-D ground reaction force (GRF) sensors data, including both quantitative and qualitative analysis.
To this aim, they first trained CNN, SVM, and MLP classifiers on the GaitRec dataset \cite{horsak2020gaitrec}, a clinical database including bilateral GRF measurements from $132$ patients with $3$ classes of orthopedic gait disorders and from $62$ healthy controls, then LRP technique has been used to explain the most relevant signal characteristics learned by the models.
Explanation assessment has been performed through two interrelated approaches: a quantitative statistical analysis using Statistical Parametric Mapping (SPM) \cite{pataky2010generalized}, and a qualitative evaluation conducted by clinical experts.
In case of binary classification (i.e., normal VS. disordered gait), SPM analysis shows significant differences in signal regions that are also highly relevant according to LRP scores, supporting the premise that models based their predictions primarily on features that are also significantly different between the two classes. 
In addition, regions with high relevance according to LRP can be largely associated with gait analysis literature and are also plausible from a clinical standpoint according to domain experts' review.
\\ 
Preliminary approaches for quality assessment of attention mechanisms are quite similar to the evaluation methods proposed for CNN-based saliency maps.
To make an example, \cite{hsieh2021explainable} developed \textit{Locality-Aware eXplainable Convolutional ATtention network} (LAXCAT), a DNN architecture for MTS classification and forecasting consisting of a CNN encoder for feature extraction and a dual attention network for simultaneously learning informative variables and time intervals.
The network has been evaluated for both seizure detection and fist movement detection tasks, using multi-channel EEG signals as input data.
In the first case, the network has been trained and tested on a seizure dataset collected in \cite{schalk2004bci2000}, whereas the benchmark Physionet DB has been used in the second case.
Afterwards, the authors defined an \textit{Attention Allocation Measure} (AAM) metric to evaluate the average percentage of attention that is correctly detected by the model across all local predictions.
The proposed model outperformed other attention-based networks available in the literature in correctly detecting relevant signal time points, even if the global AAM value obtained in the best case is modest (approximately $25\%$). 

\begin{table}[!htbp]
\begin{minipage}{\textwidth}
\caption{List of XAI studies performing explanation quality assessment.}\label{res_tab_6}
\begin{tabular*}{\textwidth}[l]{p{.05\textwidth}p{.05\textwidth} p{.2\textwidth}p{.1\textwidth}p{.1\textwidth}p{.15\textwidth}p{.15\textwidth}}
\toprule
\textbf{Ref.} &\textbf{\# Cit.}& \textbf{Application} & \textbf{Input Data} & \textbf{AI model(s)} & 
\textbf{XAI method(s)} &\textbf{Dataset(s)} \\
\midrule
\cite{cheng2021vbridge} ($2021$)&$6$&Surgical complication risk prediction&EHR& Not found& SHAP& PIC \cite{zeng2020pic}\\
\midrule
\cite{kumarakulasinghe2020evaluating} ($2020$)&$13$&Sepsis detection&EHR&RF& LIME& PhysioNet $2019$ \cite{reyna2019early}\\
\midrule
\cite{barda2020qualitative} ($2020$)&$24$&Mortality risk prediction&EHR&RF&SHAP&Retrospective study\\
\midrule
\cite{penafiel2020predicting} ($2020$)&$4$&Stroke prediction&EHR&DS-GD& DS rules, LIME& Retrospective study\\
\midrule
\cite{hatwell2020ada} ($2020$)&$18$&CAD&EHR&Ada-WHIPS& Ada-WHIPS rules, Anchors, LORE& $9$ CAD datasets from UCI ML repository\\
\midrule
\cite{zhang2021explainability} ($2021$)&$8$&Signal quality assessment&PPG& CNN& SHAP, integrated gradients, guided saliency& Training: ICU dataset \cite{drew2014insights}; Test: stroke dataset \cite{pereira2019deep}\\
\midrule
\cite{wickstrom2020uncertainty} ($2020$)&$9$& AMI detection, SSI detection& Echo, CRP measures & FCN& CAM& ECG$200$ dataset \cite{dau2019ucr}, SSI dataset \cite{mikalsen2016learning}\\
\midrule
\cite{slijepcevic2021explaining} ($2021$)&$5$&Gait disorder detection& $3$-D GRF &CNN, SVM, MLP&LRP&GaitRec dataset \cite{horsak2020gaitrec}\\
\midrule
\cite{hsieh2021explainable} ($2021$)&$24$&Seizure detection, fist movement detection& EEG& CNN&Attention&Seizure dataset \cite{schalk2004bci2000}, Physionet DB\\
\bottomrule
\end{tabular*}
\end{minipage}
\end{table}

\section{Discussion}\label{discussion}
From the literature analysis, it may be noticed that learning from static or longitudinal data clearly determines the selection of candidate models, and hence it impacts on the application of XAI methods.
Static data have no time dimension by definition, and hence tree ensembles, such as RF and XGBoost, are generally the best-performing models to learn complex non-linearities between target and inputs.
Shallow MLP networks, ad-hoc networks (e.g., TabNet), as well as data mining approaches (e.g., fuzzy inference, evolutionary algorithms) are used only in limited cases.
Although less complex than DNN, tree ensembles are still black-boxes and demand for explainability. Post-hoc methods are generally preferred to explain these models, as they do not impose any trade-off between interpretability and predictive performance.
In terms of explanation output, feature attribution methods overcome decision rule-based algorithms like \textit{Anchors} and \textit{InTrees}, with LIME and SHAP being, by far, the most comprehensive and dominant methods across the literature to detect and visualise feature importance. In particular, SHAP carries several advantages with respect to LIME that makes it a reference explanatory technique:
\begin{itemize}
    \item {\it Solid theoretical foundations:} SHAP is backed by the game theoretically optimal Shapley values. In particular, SHAP satisfies three desirable properties of explanations: local accuracy, missingness, and consistency \cite{lundberg2017unified}, which contribute to generate trustworthy explanations. Differently, LIME is a more heuristic approach based on the assumption of a sparse linear model as local surrogate model, which may make explanations unfaithful if the underlying model is highly non-linear even in the locality of the target instance.
    \item {\it Scope of explanations:} SHAP provides both local and global explanations, while LIME is well suited only for local interpretability.
    As already mentioned in Section \ref{background_tabular}, simply averaging LIME across multiple local predictions is subject to high variance and may negatively impact on explanation stability.
    On the other hand, LIME-SP just provides a set of representative explanations for each class, which is definitely not enough to get insights into the inner working of a model as a whole.
    \item {\it Model-agnosticism:} SHAP provide different implementations to deal with any kind of black-box model.
    However, it is important to outline that only \textit{TreeSHAP} algorithm is used for tree-based models as it provides fast and exact results \cite{lundberg2020local2global}. 
    \item {\it Feature interactions:} SHAP can also break down input-target relationships into main effects and interaction effects, making possible to investigate how and to what extent interactions between each variable pair contribute to model predictions, both at local and global stages.
\end{itemize}
In the time series domain, RNN and CNN currently represent the best performing solutions to learn from raw data without the need of heavy signal pre-processing methods.
On the other hand, their complexity combined with the non intuitive nature and lower interpretability of time series data makes understanding these models even more complicated. By looking at the results of our literature survey, attention is the reference ante-hoc method to integrate explainability at variable and time levels when developing RNN, while gradient-based methods are usually applied to generate post-hoc explanations for CNN, with Grad-CAM and LRP being the most influential algorithms.
Example-based methods involving prototype learning and perturbation-based methods (e.g., occlusion) can also be found, as well as CNN equipped with attention modules in place of post-hoc methods.
Attention and saliency methods are generally used only for local interpretability, and they both provide heatmaps to be overlapped onto the input signal to highlight which specific components get the most attention from the model while classification is performed. However, they do not specify how the information contained in the most relevant signal sub-sequences is used to make a given prediction.
\\
Despite their increasing adoption to bring explainability to AI models developed to accomplish medical tasks, an extensive and multi-dimensional assessment of explanations encompassing clinical relevance, consistency, and quality is generally missing for most health applications, and especially in terms of quality. 
The main goal of quality evaluation is twofold: $1$) enabling a formal comparison among suitable methods in order to choose which one should be preferred; $2$) determining if the offered form of explainability meets end users' needs.
These objectives should not be in contrast, indeed they require integration to define the solution providing the best evidence of success.
\\
Even if some preliminary evaluation processes conducted in the revised works provide promising results, the reference XAI methods highlighted in our survey still carry out several limitations that may undermine their safety and reliability in real scenarios.
As a result, employing them to understand models deployed in high-stake health applications can be risky.
In the next section, we outline some of the main limitations and current challenges related to the application of XAI methods in healthcare, also highlighting potential yet under-developed research directions to address these issues.
Specifically, we begin with a focus on the challenges introduced by the evaluation process of explanations from both user-centered and user-independent perspectives, then we outline some technical/methodological flaws and potential advancements.

\subsection{XAI Challenges in Healthcare}

\subsubsection{Clinical Validation}
Evidence-based assessment is still a prevailing paradigm in healthcare with respect to data-driven knowledge, and hence it plays a central role in the decision making process. Therefore, clinical validation is a basic requirement for the acceptance and deployment of AI systems in real-life medical applications. 
However, clinical comparison is also valuable for other reasons. First, it can be used to mitigate {\it AI bias}. This can be achieved by detecting systematic deviations in the generated explanations with respect to the expected behaviour, such as counter-intuitive and clinically misaligned input-target relationships. However, this aspect is often missing or underestimated, as only the top input-target relationships that are in line with the current clinical knowledge are generally reported. Detecting and investigating such possible model failures is necessary as well, and even more before AI systems go through certification and approval processes as it may enable saving time, resources, and user efforts.
In addition, clinical comparison may also be used for {\it knowledge discovery}, by highlighting possible emergent biomarkers/risk factors that might not be considered (or less used) by domain experts, thus potentially enabling clinical utility of AI models.
In turn, medical AI systems may benefit of knowledge discovery for a better learning through iterative model design and training, involving both model selection and data collection/curation stages.

\subsubsection{Stability and robustness}
AI models should not be only clinically sound, but they should also be stable and robust.
Models should generate similar predictions and explanations for equal or close data points over the time. Evaluate consistency among explanations provided by multiple methods at global/local stage is a straightforward and inexpensive approach to get insights into model stability and robustness, but results must be handled cautiously.
Empirical and theoretical analysis demonstrated that the majority of popular feature importance and counterfactual explanation methods are non-robust \cite{mishra2021survey}.
In particular, most works focused on XAI methods that are specific to DNN models.
\cite{kindermans2019reliability} demonstrated that simply adding a constant shift to input data causes several gradient-based methods to attribute wrong saliency.
Similar results have been also shown by \cite{alvarez2018robustness} for model-agnostic methods such as LIME and SHAP, and the authors also claimed such methods are even more non-robust than gradient-based ones.
As a result, comparing the level of agreement of explainability brought by most of these methods cannot exhaustively assess the stability and robustness in AI models, whereas sensitivity analysis is currently the most targeted analytical approach to better expose model vulnerabilities.
For example, \cite{siddiqui2019tsviz} applied perturbation on the most salient part of the input sequences leading to a huge drop in classification, thus highlighting model sensitivity to noises.
Sensitivity analysis conducted in \cite{hartl2020explainability} highlighted the most salient features as the same with highest potential to cause misclassification, thus making the model easily susceptible to adversarial attacks.
\\
However, these aspects may be only partially investigated but not definitely solved through XAI, and they still require novel evaluation metrics and practices that will guarantee the \enquote{right} behavior of a model. 
Extensive training is currently the main solution to mitigate sensitivity against noise and adversarials, while extensive and external validation enables the detection of several issues. \cite{song2020cross} and \cite{duckworth2021using} are two pioneer research studies reviewed in this survey that proposed a joint analysis of performance degradation and explanation consistency to reveal model transportability and temporal validity concerns, respectively.
Moreover, an in-depth quality evaluation of collected data is equally important as it enables understanding the limitations of the resulting model.
However, such evaluations can be costly and consuming, and hence often impractical.
In addition, they might never be enough to guarantee robustness and stability, as random perturbations leading to wrong decisions may occur anyway. Therefore, a potential countermeasure to face with input perturbations could be evaluating how different is a given prediction with respect to other data points in the validation set, for instance by detecting samples that are far from the input data distribution (i.e., outliers).
As a result, every prediction should be associated with a score indicating how confident is the model in its decision, leaving end users the autonomy to choose whether to trust such prediction or not.

\subsubsection{Quality assessment}
When using XAI to build trust and confidence, evaluating the quality of explanations is key.
Quantitative proxy metrics are necessary for an initial and objective assessment as well as a formal comparison of explanation methods to identify the best-quality option regardless of end users' needs and preferences.
However, this research line is currently under development, and there are no standard evaluation frameworks available in the community yet.
This might be due to several factors, such as the different nature of explanations, the different models, as well as the different input data types.
As a result, most of existing quantitative evaluations are based on metrics that measure performance degradation through input perturbation analysis \cite{alvarez2018towards}. Other quantitative evaluations often perform a comparison with expert annotations.
This approach has been originally proposed for image-based applications to compare saliency maps with expert annotations \cite{mohseni2018human}, then it has been adapted for the time series domain, as it can be noticed in Section \ref{quality}.
However, one major flaw is related to the introduction of human bias in such evaluation process, which requires large-scale studies to be canceled or at least reduced. Moreover, the unintuitive nature of some time series (e.g., inertial measurements) poses additional challenges as even domain experts cannot provide an exact and reliable ground truth to compare saliency methods with.
\\
Although quantitative metrics may enable a formal evaluation, human-centered assessment is a fundamental yet missing step for tuning explanations in order they can effectively and successfully reach target users. In addition, user studies can also integrate both qualitative assessment related for instance to usefulness, satisfaction, and trust on explanations, as well as quantitative measurements of human-machine task performance.
As a result, quantitative evaluation framework should be complemented with user-centered assessment before employing AI systems in real-life. 
As a first note, the intended audience of XAI strongly impacts on the scope and purpose of explanations, as different users imply different responsibilities, and hence different needs and preferences.
AI experts as well as non-expert users added with responsibility in case of system failure, may be probably more interested in global explanations as they provide an overview of model behaviour that might help to decide whether using the model is reliable or too risky. On the other hand, consumers of the model (i.e., clinicians as well as patients) will rather look for local explanations related to a single prediction (or group of predictions) that affect them, while they cannot take any advantage from understanding the model as a whole.
Despite the huge differences that can be shown by different pools of humans, there is a lack of user studies exploring the needs of clinicians (and also patients) in the XAI literature. 
Non-expert users generally have difficulties due to the gap between algorithmic outputs and human-consumable explanations \cite{antoniadi2021current}. Therefore, pushing interdisciplinary collaboration moving forward is necessary not only to explains AI systems in a more transparent and interpretable manner, but also to determine how end users would like to receive information and share meaning with each other.

\subsubsection{Interactivity}
Human-AI interaction in healthcare is an open challenge especially in terms of trust, usability, and acceptability, both from physicians and patients, and this can be closely related to explainability.
A recent statistical study with more than $1300$ physicians demonstrated that $88\%$ positively evaluated AI systems accompanied by model-agnostic explanations \cite{diprose2020physician} in terms of trust and system understanding.
Improving user understanding of the decision making process is even more important in remote healthcare, as AI algorithms embedded in e-health and m-health apps often require a direct interaction with non-expert users (i.e., patients).
In a recent study conducted in \cite{su2020analyzing}, $40$ popular commercial m-health apps coming from different domains (e.g., fitness, mental health, nutrition) have been reviewed by $400$ consumers, and obtained results indicate that most of the involved users were not able to understand the motivation of AI outcomes due to the lack of information and explanations.
\\
However, simply associating model outcomes with the corresponding explanations is not sufficient to build trust and confidence in a system.
This should include specific interactions by design to correctly provide system outputs to end users.
According to \cite{wang2019designing}, designing effective human-centered XAI strategies should first bridge algorithm-generated explanations with user explanation goals and expected reasoning methods. An in-depth analysis of these relationships is essential to provide effective data structures and visualisations, as well as to mitigate common cognitive biases in the human decision making leading diagnostic errors in medicine. Then, more specific approaches are necessary to fill the gap between experimental implementations and solutions acceptable by physicians in daily practice. For instance, \cite{kovalchuk2022three} highlights the need of explicit references to existing clinical norms (e.g., scales and recommendations) and domain-specific explanation of the results, in order to improve decision making while keeping the problem-specific reasoning interface similar to well-known clinical tools.
\\
Moreover, a major finding highlighted both in \cite{wang2019designing} and \cite{kovalchuk2022three} is the need of integrating different XAI interactions and facilities to improve human interpretability, either by the joint design with healthcare professionals or through successive refinements following human feedback.
Some examples may include: linking explanations with raw and/or processed data to address clinicians' difficulties with ML features; clustering patient data to provide cohort-level evidence; adding context information to generate a more comprehensive overview of a patient. 
Moreover, clinicians often outlined the need of WHAT-IF analysis tools to probe AI systems when their output is not as expected, such as in case of wrong or \enquote{suspicious} decisions.
This is generally achieved through the investigation of \textit{hypothetical} questions (i.e., \textit{how can specific input updates change model outcomes?}) and/or \textit{counterfactual} questions (i.e., \textit{what minimum change should I apply to a given input to drive the model towards a desired different prediction?}) \cite{wexler2019if}.
In this way, clinicians can edit specific data instances and see how they changes or influences model outcomes. However, WHAT-IF analysis is generally missing in most XAI visual analytics tools proposed in the literature. Indeed, among the reviewed works, only \textit{VBridge} \cite{cheng2021vbridge} and \textit{RetainVis} \cite{kwon2018retainvis} systems include this functionality to assess how and to what extent abnormal feature values contribute to a given prediction.
Eventually, end users also require multiple explanation methods and/or format to be included in the same system. Outlining how different explanations can be best combined in a user interface and how these combined AI systems should be then evaluated is another open research problem.

\subsubsection{Human-consumable explanations for time series data}\label{challenges_5}
By looking at XAI methods, and especially in the time series field, one question arises: \textit{are explanations really explaining?}. Attention and saliency methods just highlight the most relevant input components for a given data instance, without providing any knowledge about how they are used by the inner model reasoning to derive the final decision. Whereas such explanations can be useful for image and text data as humans are generally able to understand their content at first sight, the situation is more challenging for time series because expert knowledge might be needed.
Prototypes learning-based approaches are even harder to be put in practice, as comparing target and exemplary sequences might result counter-intuitive.
However, saliency can still be considered a na\"ive approach to provide explanations and increase the trust in the system in some cases, as a user or a developer can rely more on a system if he/she knows that the model gives most attention on well-known parts of the input signal.
A practical example can be the usage of saliency methods to highlight relevant ECG patterns and/or waveforms for detecting heart anomalies, such as in \cite{ivaturi2021comprehensive}.
This approach does not differ so much from the current clinical practice based on the visual inspection of ECG traces, hence it can effectively assist cardiologists in their decision making.
Conversely, the trust brought by these methods can be questioned when using unintuitive and less inherently interpretable time series, which even domain expert may struggle to understand.
A typical example in healthcare is the analysis of human gait from inertial measurements (e.g., accelerometer, gyroscope, GRF sensor, etc...) to detect neurodegenerative diseases \cite{filtjens2021modelling}.
In such cases, some data mining methods should be applied to automatically extract meaningful information hidden within the highlighted signal components \cite{rojat2021explainable}. 
Moreover, novel algorithms may be needed for generating human-friendly explanations from time series. \textit{Concepts} have gained an increasing attention as a new class of XAI methods, which is able to deduce high-level and human-understandable contents (i.e., concepts) from lower-level sensor data and to use them as meta-explanations.
Some recent concept-based methods include \textit{Testing with Concept Activation Vectors} (T-CAV) \cite{kim2018interpretability}, \textit{Automatic Concept-based Explanations} (ACE) \cite{ghorbani2019towards}, \textit{Causal Concept Effect} (CaCE) \cite{goyal2019explaining}, as well as \textit{ConceptSHAP} \cite{yeh2020completeness}. By designing explanations as concepts, these methods allow integrating standard saliency-based local explanations with additional meta information on the most contributing factors.

\subsubsection{Understanding DNN models}
Beyond the underlying limitations of saliency methods highlighted in Section \ref{challenges_5}, with particular reference to the time series field, the adoption of DNN for decision making in healthcare is further limited by the lack of reference methods to provide global explanations and to investigate the inner functioning of such models.
As far as global explanations is concerned, saliency methods may provide useful information only if structured and meaningful data are passed as input (e.g. handcrafted features, clinical attributes), but this condition rarely occur while developing DNN models.
On the other hand, explaining features learned in the latent space might be useless as they often do not have a human-understandable meaning. 
Very few research works extended saliency methods to produce global explanations, but most of them have been originally designed for image data.
For instance, \cite{oviedo2019fast} averaged CAM saliency maps across all training examples to detect the main discriminating features for each class.
More formal algorithmic approaches can be found as well. Activation Maximization method is probably the most recognised explanation method for generating human-interpretable representations of different intermediate filters within a deep CNN architecture \cite{simonyan2013deep}.
However, the above approaches are less suitable for explaining signal-based methods as time series data are less intuitive than images, so their output is more difficult to understand by end users.
Other existing global explanation methods, such as Tree Regularization \cite{wu2018beyond} or Network Dissection \cite{bau2017network}, are also unfit for signal processing.
Indeed, approximating DNN with decision trees may become hard to interpret with high dimensional
data. On the other hand, the Network Dissection algorithm works by quantifying the network response to human understandable concepts contained in image data, such as colors, objects, and scenes. As a result, it requires densely labeled time series data to ideally match concepts with sensor data samples, which is often inapplicable.
\\
The literature contribution to global explanation methods for DNN models learned from time series data is even more limited. \cite{siddiqui2019tsviz} clustered CNN filters according to their activation pattern, based on the assumption that filters with similar activation patterns basically detect the same content.
\cite{cho2020interpretation} proposed an alternative clustering approach by grouping input sub-sequences that activate the same nodes, and associating each cluster to a representative example. 
These methods represent preliminary works to achieve some global insights, but much more effort is still required to generate human-consumable explanations, and especially to explain the latent space of CNN.
\\
In addition, no methods seem to exist to probe DNN in case of wrong or \enquote{strange} explanations.
Indeed, benchmark methods such as Grad-CAM and LRP are based on heuristic to produce locally interpretable information, but the internal process and computations cannot be discerned.
For example, changing one (or more weights) in the hidden layers does not easily reveal any useful clue about model functioning.
This clearly limits the interactions with DSS based on DNN models as clinicians cannot investigate how and why a certain outcome is reached. Trust and confidence in such systems is negatively affected as well.

\subsubsection{Causability}
Human assessment of post-hoc, model-agnostic methods often criticised that explanations mainly show correlations \cite{bruckert2020next}, while they cannot fulfil expectations when it comes to enhance medical decisions on tasks where there is a need to understand causal relationships.
Therefore, a consistent mapping of explainability with causability is crucial to design effective interactions for physicians and, consequently, successful user interfaces \cite{holzinger2021toward}.
To this aim, \cite{holzinger2020measuring} proposed a novel evaluation framework, namely System Causability Scale (SCS), which allows to quantify the level reached by an explanation in providing a cause-effect understanding with effectiveness, efficiency, and satisfaction by end users. However, simply measuring the degree of causality of explanation statements may not necessarily facilitate medical decisions, such as for detecting the impact of different health interventions on model outcomes.
The next level of data-driven decisions require understanding how the systems react to external stimulation and why the change occurs.
Answering to these questions through XAI may be tricky, as changing the statistical distribution of affected input variables may invalidate the basic assumptions on which the models were built, making results unreliable.
To this aim, \textit{Causal AI} is emerging as a novel AI paradigm combining traditional ML algorithms with the principles of causal reasoning to build causal inference systems \cite{scholkopf2021toward}.
As a result, we expect causal AI to bridge the gap between predictions and decision-making, with the potential to enable AI researchers and clinicians to jointly design and simulate an intervention and infer causality by relying on already available data.
This might also bring a huge innovation in evidence-based medicine, as it may support and complement clinical randomised controlled trials for measuring the impact of health interventions over the long term.

\section{Conclusions}
Tabular and time series data are widely used in clinical and remote AI-empowered health applications, enabling the accurate detection of a variety of diseases and conditions in a more responsive, less-invasive, and low-cost fashion.
However, they are not commonly analysed from XAI perspective with respect to biomedical images and medical text records. Nevertheless, XAI techniques are required to bring explainability to complex models learned from this data. An extensive evaluation of generated explanations including clinical relevance, consistency, and quality, is also a necessary (but often missing) step to ensure XAI methods effectively improve human understanding and confidence in AI decision making, thus representing a natural step towards {\it Trustworthy AI}.
However, it should be acknowledged that the list of potential XAI challenges is much broader than only building trustworthiness.
For instance, XAI may contribute to ensure the adherence to ethical principles and values.
Recently, \cite{muller2021ten} elaborated ten practical principles from the most recent and relevant works on the ethical application of AI, with particular reference to the medical field, addressing several issues such as clear identification of decisions, actions, and communications performed by an AI agent, accountability, lawfulness, and compliance with the state-of-the-art theories and practices. Explainability can play a central role in respecting most of these principles. As already discussed, explanations can be compared with medical knowledge for clinical validation, as well as for revealing hidden AI bias.
Moreover, XAI can contribute to ensure the fairness of AI algorithms (commandment $9$ in the proposed checklist) by verifying that machine decisions are made without any discrimination based on patient characteristics and/or groups, such as gender and ethnicity \cite{hardt2016equality}.
\\
XAI may also help make the system compliant with the patient-centered care paradigm. As patients are nowadays considered active partners in the care provisioning process, they have the right to choose and control their medical treatments and recommendations \cite{bjerring2021artificial}. Therefore, XAI can boost patient acceptance and comfort in undergoing AI-empowered medical practices as long as clinicians are able to comprehend and report the reasons behind a given decision.
From the medical perspective, XAI enables the definition of a shared meaning of the decision process, so that clinicians can support their decisions. Moreover, XAI may also facilitate the resolution of disagreements between human and machine-based decisions.
However, a clear definition of liability in case of wrong decisions is a core touch-point that is in current need of strong clarification by legislative bodies.
\\
By combining the above ethical, social, and medical implications with the technical challenges discussed in this survey, it can be easily noticed that we are far from reaching end-to-end XAI systems ready for large-scale deployment with minimal human supervision. Methodological improvements, user-centered studies, and also clear and full regulations are still necessary for such systems to be accepted and used in the medical practice.
Focusing only on the technical aspects, some may argue that explainable modelling is the unique solution for high-stake domains such as healthcare. However, the development of novel and inherently interpretable AI algorithms often imposes a trade-off in terms of predictive performance, which can raise further criticism as it may be exchanging better medical outcomes with an increased transparency.
On the other hand, post-hoc explanations can prioritise model accuracy in healthcare, but they are generally approximations of its inner reasoning. As a result, they cannot be completely trustable by definition. 
\\
Before XAI will be able to reach a robust way to handle interpretability, AI predictions will inevitably carry some risks and failures, as for any new technology, treatment, and drug we aim to introduce in healthcare. Until that moment, XAI should be considered as a complementary support and not a replacement of standard medical practice, and domain expert supervision is still necessary to make the final decision.
According to this perspective, XAI and evidence-based assessment can safely coexist and improve medical outcomes. However, their coexistence needs a careful orchestration to avoid a constant conflict between innovative and gold standard approaches, which can undermine the effective usage of medical AI systems with negative consequences on both patients and clinicians.

\backmatter



\section*{Declarations}
\begin{itemize}
\item Funding: This
work has been partially funded by the European Commission under
H2020-INFRAIA-2019-1SoBigData-PlusPlus project. Grant No.: 871042.
\item Competing interests: 
The authors have no relevant financial or non-financial interests to disclose.
\end{itemize}

\bibliography{references.bib}


\begin{thebibliography}{198}
\ifx \bisbn   \undefined \def \bisbn  #1{ISBN #1}\fi
\ifx \binits  \undefined \def \binits#1{#1}\fi
\ifx \bauthor  \undefined \def \bauthor#1{#1}\fi
\ifx \batitle  \undefined \def \batitle#1{#1}\fi
\ifx \bjtitle  \undefined \def \bjtitle#1{#1}\fi
\ifx \bvolume  \undefined \def \bvolume#1{\textbf{#1}}\fi
\ifx \byear  \undefined \def \byear#1{#1}\fi
\ifx \bissue  \undefined \def \bissue#1{#1}\fi
\ifx \bfpage  \undefined \def \bfpage#1{#1}\fi
\ifx \blpage  \undefined \def \blpage #1{#1}\fi
\ifx \burl  \undefined \def \burl#1{\textsf{#1}}\fi
\ifx \doiurl  \undefined \def \doiurl#1{\url{https://doi.org/#1}}\fi
\ifx \betal  \undefined \def \betal{\textit{et al.}}\fi
\ifx \binstitute  \undefined \def \binstitute#1{#1}\fi
\ifx \binstitutionaled  \undefined \def \binstitutionaled#1{#1}\fi
\ifx \bctitle  \undefined \def \bctitle#1{#1}\fi
\ifx \beditor  \undefined \def \beditor#1{#1}\fi
\ifx \bpublisher  \undefined \def \bpublisher#1{#1}\fi
\ifx \bbtitle  \undefined \def \bbtitle#1{#1}\fi
\ifx \bedition  \undefined \def \bedition#1{#1}\fi
\ifx \bseriesno  \undefined \def \bseriesno#1{#1}\fi
\ifx \blocation  \undefined \def \blocation#1{#1}\fi
\ifx \bsertitle  \undefined \def \bsertitle#1{#1}\fi
\ifx \bsnm \undefined \def \bsnm#1{#1}\fi
\ifx \bsuffix \undefined \def \bsuffix#1{#1}\fi
\ifx \bparticle \undefined \def \bparticle#1{#1}\fi
\ifx \barticle \undefined \def \barticle#1{#1}\fi
\bibcommenthead
\ifx \bconfdate \undefined \def \bconfdate #1{#1}\fi
\ifx \botherref \undefined \def \botherref #1{#1}\fi
\ifx \url \undefined \def \url#1{\textsf{#1}}\fi
\ifx \bchapter \undefined \def \bchapter#1{#1}\fi
\ifx \bbook \undefined \def \bbook#1{#1}\fi
\ifx \bcomment \undefined \def \bcomment#1{#1}\fi
\ifx \oauthor \undefined \def \oauthor#1{#1}\fi
\ifx \citeauthoryear \undefined \def \citeauthoryear#1{#1}\fi
\ifx \endbibitem  \undefined \def \endbibitem {}\fi
\ifx \bconflocation  \undefined \def \bconflocation#1{#1}\fi
\ifx \arxivurl  \undefined \def \arxivurl#1{\textsf{#1}}\fi
\csname PreBibitemsHook\endcsname

\bibitem{topol2019high}
\begin{barticle}
\bauthor{\bsnm{Topol}, \binits{E.J.}}:
\batitle{High-performance medicine: the convergence of human and artificial
  intelligence}.
\bjtitle{Nature medicine}
\bvolume{25}(\bissue{1}),
\bfpage{44}--\blpage{56}
(\byear{2019})
\end{barticle}
\endbibitem

\bibitem{das2020opportunities}
\begin{botherref}
\oauthor{\bsnm{Das}, \binits{A.}},
\oauthor{\bsnm{Rad}, \binits{P.}}:
Opportunities and challenges in explainable artificial intelligence (xai): A
  survey.
arXiv preprint arXiv:2006.11371
(2020)
\end{botherref}
\endbibitem

\bibitem{cina2022we}
\begin{botherref}
\oauthor{\bsnm{Cin{\`a}}, \binits{G.}},
\oauthor{\bsnm{R{\"o}ber}, \binits{T.}},
\oauthor{\bsnm{Goedhart}, \binits{R.}},
\oauthor{\bsnm{Birbil}, \binits{I.}}:
Why we do need explainable ai for healthcare.
arXiv preprint arXiv:2206.15363
(2022)
\end{botherref}
\endbibitem

\bibitem{tjoa2020survey}
\begin{barticle}
\bauthor{\bsnm{Tjoa}, \binits{E.}},
\bauthor{\bsnm{Guan}, \binits{C.}}:
\batitle{A survey on explainable artificial intelligence (xai): Toward medical
  xai}.
\bjtitle{IEEE transactions on neural networks and learning systems}
\bvolume{32}(\bissue{11}),
\bfpage{4793}--\blpage{4813}
(\byear{2020})
\end{barticle}
\endbibitem

\bibitem{gulum2021review}
\begin{barticle}
\bauthor{\bsnm{Gulum}, \binits{M.A.}},
\bauthor{\bsnm{Trombley}, \binits{C.M.}},
\bauthor{\bsnm{Kantardzic}, \binits{M.}}:
\batitle{A review of explainable deep learning cancer detection models in
  medical imaging}.
\bjtitle{Applied Sciences}
\bvolume{11}(\bissue{10}),
\bfpage{4573}
(\byear{2021})
\end{barticle}
\endbibitem

\bibitem{mondal2021xvitcos}
\begin{barticle}
\bauthor{\bsnm{Mondal}, \binits{A.K.}},
\bauthor{\bsnm{Bhattacharjee}, \binits{A.}},
\bauthor{\bsnm{Singla}, \binits{P.}},
\bauthor{\bsnm{Prathosh}, \binits{A.}}:
\batitle{xvitcos: Explainable vision transformer based covid-19 screening using
  radiography}.
\bjtitle{IEEE Journal of Translational Engineering in Health and Medicine}
\bvolume{10},
\bfpage{1}--\blpage{10}
(\byear{2021})
\end{barticle}
\endbibitem

\bibitem{faruk2021residualcovid}
\begin{bchapter}
\bauthor{\bsnm{Faruk}, \binits{M.F.}}:
\bctitle{Residualcovid-net: An interpretable deep network to screen covid-19
  utilizing chest ct images}.
In: \bbtitle{2021 3rd International Conference on Electrical \& Electronic
  Engineering (ICEEE)},
pp. \bfpage{69}--\blpage{72}
(\byear{2021}).
\bcomment{IEEE}
\end{bchapter}
\endbibitem

\bibitem{payrovnaziri2020explainable}
\begin{barticle}
\bauthor{\bsnm{Payrovnaziri}, \binits{S.N.}},
\bauthor{\bsnm{Chen}, \binits{Z.}},
\bauthor{\bsnm{Rengifo-Moreno}, \binits{P.}},
\bauthor{\bsnm{Miller}, \binits{T.}},
\bauthor{\bsnm{Bian}, \binits{J.}},
\bauthor{\bsnm{Chen}, \binits{J.H.}},
\bauthor{\bsnm{Liu}, \binits{X.}},
\bauthor{\bsnm{He}, \binits{Z.}}:
\batitle{Explainable artificial intelligence models using real-world electronic
  health record data: a systematic scoping review}.
\bjtitle{Journal of the American Medical Informatics Association}
\bvolume{27}(\bissue{7}),
\bfpage{1173}--\blpage{1185}
(\byear{2020})
\end{barticle}
\endbibitem

\bibitem{kok2022explainable}
\begin{botherref}
\oauthor{\bsnm{Kok}, \binits{I.}},
\oauthor{\bsnm{Okay}, \binits{F.Y.}},
\oauthor{\bsnm{Muyanli}, \binits{O.}},
\oauthor{\bsnm{Ozdemir}, \binits{S.}}:
Explainable artificial intelligence (xai) for internet of things: A survey.
arXiv preprint arXiv:2206.04800
(2022)
\end{botherref}
\endbibitem

\bibitem{sahakyan2021explainable}
\begin{barticle}
\bauthor{\bsnm{Sahakyan}, \binits{M.}},
\bauthor{\bsnm{Aung}, \binits{Z.}},
\bauthor{\bsnm{Rahwan}, \binits{T.}}:
\batitle{Explainable artificial intelligence for tabular data: A survey}.
\bjtitle{IEEE Access}
\bvolume{9},
\bfpage{135392}--\blpage{135422}
(\byear{2021})
\end{barticle}
\endbibitem

\bibitem{rojat2021explainable}
\begin{botherref}
\oauthor{\bsnm{Rojat}, \binits{T.}},
\oauthor{\bsnm{Puget}, \binits{R.}},
\oauthor{\bsnm{Filliat}, \binits{D.}},
\oauthor{\bsnm{Del~Ser}, \binits{J.}},
\oauthor{\bsnm{Gelin}, \binits{R.}},
\oauthor{\bsnm{D{\'\i}az-Rodr{\'\i}guez}, \binits{N.}}:
Explainable artificial intelligence (xai) on timeseries data: A survey.
arXiv preprint arXiv:2104.00950
(2021)
\end{botherref}
\endbibitem

\bibitem{markus2021role}
\begin{barticle}
\bauthor{\bsnm{Markus}, \binits{A.F.}},
\bauthor{\bsnm{Kors}, \binits{J.A.}},
\bauthor{\bsnm{Rijnbeek}, \binits{P.R.}}:
\batitle{The role of explainability in creating trustworthy artificial
  intelligence for health care: a comprehensive survey of the terminology,
  design choices, and evaluation strategies}.
\bjtitle{Journal of Biomedical Informatics}
\bvolume{113},
\bfpage{103655}
(\byear{2021})
\end{barticle}
\endbibitem

\bibitem{amann2020explainability}
\begin{barticle}
\bauthor{\bsnm{Amann}, \binits{J.}},
\bauthor{\bsnm{Blasimme}, \binits{A.}},
\bauthor{\bsnm{Vayena}, \binits{E.}},
\bauthor{\bsnm{Frey}, \binits{D.}},
\bauthor{\bsnm{Madai}, \binits{V.I.}}:
\batitle{Explainability for artificial intelligence in healthcare: a
  multidisciplinary perspective}.
\bjtitle{BMC Medical Informatics and Decision Making}
\bvolume{20}(\bissue{1}),
\bfpage{1}--\blpage{9}
(\byear{2020})
\end{barticle}
\endbibitem

\bibitem{linardatos2020explainable}
\begin{barticle}
\bauthor{\bsnm{Linardatos}, \binits{P.}},
\bauthor{\bsnm{Papastefanopoulos}, \binits{V.}},
\bauthor{\bsnm{Kotsiantis}, \binits{S.}}:
\batitle{Explainable ai: A review of machine learning interpretability
  methods}.
\bjtitle{Entropy}
\bvolume{23}(\bissue{1}),
\bfpage{18}
(\byear{2020})
\end{barticle}
\endbibitem

\bibitem{guidotti2018local}
\begin{botherref}
\oauthor{\bsnm{Guidotti}, \binits{R.}},
\oauthor{\bsnm{Monreale}, \binits{A.}},
\oauthor{\bsnm{Ruggieri}, \binits{S.}},
\oauthor{\bsnm{Pedreschi}, \binits{D.}},
\oauthor{\bsnm{Turini}, \binits{F.}},
\oauthor{\bsnm{Giannotti}, \binits{F.}}:
Local rule-based explanations of black box decision systems.
arXiv preprint arXiv:1805.10820
(2018)
\end{botherref}
\endbibitem

\bibitem{doshi2017towards}
\begin{botherref}
\oauthor{\bsnm{Doshi-Velez}, \binits{F.}},
\oauthor{\bsnm{Kim}, \binits{B.}}:
Towards a rigorous science of interpretable machine learning.
arXiv preprint arXiv:1702.08608
(2017)
\end{botherref}
\endbibitem

\bibitem{arrieta2020explainable}
\begin{barticle}
\bauthor{\bsnm{Arrieta}, \binits{A.B.}},
\bauthor{\bsnm{D{\'\i}az-Rodr{\'\i}guez}, \binits{N.}},
\bauthor{\bsnm{Del~Ser}, \binits{J.}},
\bauthor{\bsnm{Bennetot}, \binits{A.}},
\bauthor{\bsnm{Tabik}, \binits{S.}},
\bauthor{\bsnm{Barbado}, \binits{A.}},
\bauthor{\bsnm{Garc{\'\i}a}, \binits{S.}},
\bauthor{\bsnm{Gil-L{\'o}pez}, \binits{S.}},
\bauthor{\bsnm{Molina}, \binits{D.}},
\bauthor{\bsnm{Benjamins}, \binits{R.}}, \betal:
\batitle{Explainable artificial intelligence (xai): Concepts, taxonomies,
  opportunities and challenges toward responsible ai}.
\bjtitle{Information fusion}
\bvolume{58},
\bfpage{82}--\blpage{115}
(\byear{2020})
\end{barticle}
\endbibitem

\bibitem{lundberg2017unified}
\begin{botherref}
\oauthor{\bsnm{Lundberg}, \binits{S.M.}},
\oauthor{\bsnm{Lee}, \binits{S.-I.}}:
A unified approach to interpreting model predictions.
Advances in neural information processing systems
\textbf{30}
(2017)
\end{botherref}
\endbibitem

\bibitem{ribeiro2016should}
\begin{bchapter}
\bauthor{\bsnm{Ribeiro}, \binits{M.T.}},
\bauthor{\bsnm{Singh}, \binits{S.}},
\bauthor{\bsnm{Guestrin}, \binits{C.}}:
\bctitle{" why should i trust you?" explaining the predictions of any
  classifier}.
In: \bbtitle{Proceedings of the 22nd ACM SIGKDD International Conference on
  Knowledge Discovery and Data Mining},
pp. \bfpage{1135}--\blpage{1144}
(\byear{2016})
\end{bchapter}
\endbibitem

\bibitem{shankaranarayana2019alime}
\begin{bchapter}
\bauthor{\bsnm{Shankaranarayana}, \binits{S.M.}},
\bauthor{\bsnm{Runje}, \binits{D.}}:
\bctitle{Alime: Autoencoder based approach for local interpretability}.
In: \bbtitle{International Conference on Intelligent Data Engineering and
  Automated Learning},
pp. \bfpage{454}--\blpage{463}
(\byear{2019}).
\bcomment{Springer}
\end{bchapter}
\endbibitem

\bibitem{zafar2019dlime}
\begin{botherref}
\oauthor{\bsnm{Zafar}, \binits{M.R.}},
\oauthor{\bsnm{Khan}, \binits{N.M.}}:
Dlime: A deterministic local interpretable model-agnostic explanations approach
  for computer-aided diagnosis systems.
arXiv preprint arXiv:1906.10263
(2019)
\end{botherref}
\endbibitem

\bibitem{elshawi2019ilime}
\begin{bchapter}
\bauthor{\bsnm{ElShawi}, \binits{R.}},
\bauthor{\bsnm{Sherif}, \binits{Y.}},
\bauthor{\bsnm{Al-Mallah}, \binits{M.}},
\bauthor{\bsnm{Sakr}, \binits{S.}}:
\bctitle{Ilime: Local and global interpretable model-agnostic explainer of
  black-box decision}.
In: \bbtitle{European Conference on Advances in Databases and Information
  Systems},
pp. \bfpage{53}--\blpage{68}
(\byear{2019}).
\bcomment{Springer}
\end{bchapter}
\endbibitem

\bibitem{ribeiro2018anchors}
\begin{bchapter}
\bauthor{\bsnm{Ribeiro}, \binits{M.T.}},
\bauthor{\bsnm{Singh}, \binits{S.}},
\bauthor{\bsnm{Guestrin}, \binits{C.}}:
\bctitle{Anchors: High-precision model-agnostic explanations}.
In: \bbtitle{Proceedings of the AAAI Conference on Artificial Intelligence},
vol. \bseriesno{32}
(\byear{2018})
\end{bchapter}
\endbibitem

\bibitem{wachter2017counterfactual}
\begin{barticle}
\bauthor{\bsnm{Wachter}, \binits{S.}},
\bauthor{\bsnm{Mittelstadt}, \binits{B.}},
\bauthor{\bsnm{Russell}, \binits{C.}}:
\batitle{Counterfactual explanations without opening the black box: Automated
  decisions and the gdpr}.
\bjtitle{Harv. JL \& Tech.}
\bvolume{31},
\bfpage{841}
(\byear{2017})
\end{barticle}
\endbibitem

\bibitem{looveren2021interpretable}
\begin{bchapter}
\bauthor{\bsnm{Looveren}, \binits{A.V.}},
\bauthor{\bsnm{Klaise}, \binits{J.}}:
\bctitle{Interpretable counterfactual explanations guided by prototypes}.
In: \bbtitle{Joint European Conference on Machine Learning and Knowledge
  Discovery in Databases},
pp. \bfpage{650}--\blpage{665}
(\byear{2021}).
\bcomment{Springer}
\end{bchapter}
\endbibitem

\bibitem{morris1991factorial}
\begin{barticle}
\bauthor{\bsnm{Morris}, \binits{M.D.}}:
\batitle{Factorial sampling plans for preliminary computational experiments}.
\bjtitle{Technometrics}
\bvolume{33}(\bissue{2}),
\bfpage{161}--\blpage{174}
(\byear{1991})
\end{barticle}
\endbibitem

\bibitem{saltelli2010variance}
\begin{barticle}
\bauthor{\bsnm{Saltelli}, \binits{A.}},
\bauthor{\bsnm{Annoni}, \binits{P.}},
\bauthor{\bsnm{Azzini}, \binits{I.}},
\bauthor{\bsnm{Campolongo}, \binits{F.}},
\bauthor{\bsnm{Ratto}, \binits{M.}},
\bauthor{\bsnm{Tarantola}, \binits{S.}}:
\batitle{Variance based sensitivity analysis of model output. design and
  estimator for the total sensitivity index}.
\bjtitle{Computer physics communications}
\bvolume{181}(\bissue{2}),
\bfpage{259}--\blpage{270}
(\byear{2010})
\end{barticle}
\endbibitem

\bibitem{friedman2001greedy}
\begin{botherref}
\oauthor{\bsnm{Friedman}, \binits{J.H.}}:
Greedy function approximation: a gradient boosting machine.
Annals of statistics,
1189--1232
(2001)
\end{botherref}
\endbibitem

\bibitem{goldstein2015peeking}
\begin{barticle}
\bauthor{\bsnm{Goldstein}, \binits{A.}},
\bauthor{\bsnm{Kapelner}, \binits{A.}},
\bauthor{\bsnm{Bleich}, \binits{J.}},
\bauthor{\bsnm{Pitkin}, \binits{E.}}:
\batitle{Peeking inside the black box: Visualizing statistical learning with
  plots of individual conditional expectation}.
\bjtitle{journal of Computational and Graphical Statistics}
\bvolume{24}(\bissue{1}),
\bfpage{44}--\blpage{65}
(\byear{2015})
\end{barticle}
\endbibitem

\bibitem{apley2020visualizing}
\begin{barticle}
\bauthor{\bsnm{Apley}, \binits{D.W.}},
\bauthor{\bsnm{Zhu}, \binits{J.}}:
\batitle{Visualizing the effects of predictor variables in black box supervised
  learning models}.
\bjtitle{Journal of the Royal Statistical Society: Series B (Statistical
  Methodology)}
\bvolume{82}(\bissue{4}),
\bfpage{1059}--\blpage{1086}
(\byear{2020})
\end{barticle}
\endbibitem

\bibitem{deng2019interpreting}
\begin{barticle}
\bauthor{\bsnm{Deng}, \binits{H.}}:
\batitle{Interpreting tree ensembles with intrees}.
\bjtitle{International Journal of Data Science and Analytics}
\bvolume{7}(\bissue{4}),
\bfpage{277}--\blpage{287}
(\byear{2019})
\end{barticle}
\endbibitem

\bibitem{wu2018beyond}
\begin{bchapter}
\bauthor{\bsnm{Wu}, \binits{M.}},
\bauthor{\bsnm{Hughes}, \binits{M.}},
\bauthor{\bsnm{Parbhoo}, \binits{S.}},
\bauthor{\bsnm{Zazzi}, \binits{M.}},
\bauthor{\bsnm{Roth}, \binits{V.}},
\bauthor{\bsnm{Doshi-Velez}, \binits{F.}}:
\bctitle{Beyond sparsity: Tree regularization of deep models for
  interpretability}.
In: \bbtitle{Proceedings of the AAAI Conference on Artificial Intelligence},
vol. \bseriesno{32}
(\byear{2018})
\end{bchapter}
\endbibitem

\bibitem{barakat2007rule}
\begin{barticle}
\bauthor{\bsnm{Barakat}, \binits{N.H.}},
\bauthor{\bsnm{Bradley}, \binits{A.P.}}:
\batitle{Rule extraction from support vector machines: A sequential covering
  approach}.
\bjtitle{IEEE Transactions on Knowledge and Data Engineering}
\bvolume{19}(\bissue{6}),
\bfpage{729}--\blpage{741}
(\byear{2007})
\end{barticle}
\endbibitem

\bibitem{simonyan2013deep}
\begin{botherref}
\oauthor{\bsnm{Simonyan}, \binits{K.}},
\oauthor{\bsnm{Vedaldi}, \binits{A.}},
\oauthor{\bsnm{Zisserman}, \binits{A.}}:
Deep inside convolutional networks: Visualising image classification models and
  saliency maps.
arXiv preprint arXiv:1312.6034
(2013)
\end{botherref}
\endbibitem

\bibitem{sundararajan2017axiomatic}
\begin{bchapter}
\bauthor{\bsnm{Sundararajan}, \binits{M.}},
\bauthor{\bsnm{Taly}, \binits{A.}},
\bauthor{\bsnm{Yan}, \binits{Q.}}:
\bctitle{Axiomatic attribution for deep networks}.
In: \bbtitle{International Conference on Machine Learning},
pp. \bfpage{3319}--\blpage{3328}
(\byear{2017}).
\bcomment{PMLR}
\end{bchapter}
\endbibitem

\bibitem{shrikumar2017learning}
\begin{bchapter}
\bauthor{\bsnm{Shrikumar}, \binits{A.}},
\bauthor{\bsnm{Greenside}, \binits{P.}},
\bauthor{\bsnm{Kundaje}, \binits{A.}}:
\bctitle{Learning important features through propagating activation
  differences}.
In: \bbtitle{International Conference on Machine Learning},
pp. \bfpage{3145}--\blpage{3153}
(\byear{2017}).
\bcomment{PMLR}
\end{bchapter}
\endbibitem

\bibitem{zeiler2014visualizing}
\begin{bchapter}
\bauthor{\bsnm{Zeiler}, \binits{M.D.}},
\bauthor{\bsnm{Fergus}, \binits{R.}}:
\bctitle{Visualizing and understanding convolutional networks}.
In: \bbtitle{European Conference on Computer Vision},
pp. \bfpage{818}--\blpage{833}
(\byear{2014}).
\bcomment{Springer}
\end{bchapter}
\endbibitem

\bibitem{zeiler2011adaptive}
\begin{bchapter}
\bauthor{\bsnm{Zeiler}, \binits{M.D.}},
\bauthor{\bsnm{Taylor}, \binits{G.W.}},
\bauthor{\bsnm{Fergus}, \binits{R.}}:
\bctitle{Adaptive deconvolutional networks for mid and high level feature
  learning}.
In: \bbtitle{2011 International Conference on Computer Vision},
pp. \bfpage{2018}--\blpage{2025}
(\byear{2011}).
\bcomment{IEEE}
\end{bchapter}
\endbibitem

\bibitem{springenberg2014striving}
\begin{botherref}
\oauthor{\bsnm{Springenberg}, \binits{J.T.}},
\oauthor{\bsnm{Dosovitskiy}, \binits{A.}},
\oauthor{\bsnm{Brox}, \binits{T.}},
\oauthor{\bsnm{Riedmiller}, \binits{M.}}:
Striving for simplicity: The all convolutional net.
arXiv preprint arXiv:1412.6806
(2014)
\end{botherref}
\endbibitem

\bibitem{zhou2016learning}
\begin{bchapter}
\bauthor{\bsnm{Zhou}, \binits{B.}},
\bauthor{\bsnm{Khosla}, \binits{A.}},
\bauthor{\bsnm{Lapedriza}, \binits{A.}},
\bauthor{\bsnm{Oliva}, \binits{A.}},
\bauthor{\bsnm{Torralba}, \binits{A.}}:
\bctitle{Learning deep features for discriminative localization}.
In: \bbtitle{Proceedings of the IEEE Conference on Computer Vision and Pattern
  Recognition},
pp. \bfpage{2921}--\blpage{2929}
(\byear{2016})
\end{bchapter}
\endbibitem

\bibitem{selvaraju2017grad}
\begin{bchapter}
\bauthor{\bsnm{Selvaraju}, \binits{R.R.}},
\bauthor{\bsnm{Cogswell}, \binits{M.}},
\bauthor{\bsnm{Das}, \binits{A.}},
\bauthor{\bsnm{Vedantam}, \binits{R.}},
\bauthor{\bsnm{Parikh}, \binits{D.}},
\bauthor{\bsnm{Batra}, \binits{D.}}:
\bctitle{Grad-cam: Visual explanations from deep networks via gradient-based
  localization}.
In: \bbtitle{Proceedings of the IEEE International Conference on Computer
  Vision},
pp. \bfpage{618}--\blpage{626}
(\byear{2017})
\end{bchapter}
\endbibitem

\bibitem{bach2015pixel}
\begin{barticle}
\bauthor{\bsnm{Bach}, \binits{S.}},
\bauthor{\bsnm{Binder}, \binits{A.}},
\bauthor{\bsnm{Montavon}, \binits{G.}},
\bauthor{\bsnm{Klauschen}, \binits{F.}},
\bauthor{\bsnm{M{\"u}ller}, \binits{K.-R.}},
\bauthor{\bsnm{Samek}, \binits{W.}}:
\batitle{On pixel-wise explanations for non-linear classifier decisions by
  layer-wise relevance propagation}.
\bjtitle{PloS one}
\bvolume{10}(\bissue{7}),
\bfpage{0130140}
(\byear{2015})
\end{barticle}
\endbibitem

\bibitem{bahdanau2014neural}
\begin{botherref}
\oauthor{\bsnm{Bahdanau}, \binits{D.}},
\oauthor{\bsnm{Cho}, \binits{K.}},
\oauthor{\bsnm{Bengio}, \binits{Y.}}:
Neural machine translation by jointly learning to align and translate.
arXiv preprint arXiv:1409.0473
(2014)
\end{botherref}
\endbibitem

\bibitem{vaswani2017attention}
\begin{botherref}
\oauthor{\bsnm{Vaswani}, \binits{A.}},
\oauthor{\bsnm{Shazeer}, \binits{N.}},
\oauthor{\bsnm{Parmar}, \binits{N.}},
\oauthor{\bsnm{Uszkoreit}, \binits{J.}},
\oauthor{\bsnm{Jones}, \binits{L.}},
\oauthor{\bsnm{Gomez}, \binits{A.N.}},
\oauthor{\bsnm{Kaiser}, \binits{{\L}.}},
\oauthor{\bsnm{Polosukhin}, \binits{I.}}:
Attention is all you need.
Advances in neural information processing systems
\textbf{30}
(2017)
\end{botherref}
\endbibitem

\bibitem{devlin2018bert}
\begin{botherref}
\oauthor{\bsnm{Devlin}, \binits{J.}},
\oauthor{\bsnm{Chang}, \binits{M.-W.}},
\oauthor{\bsnm{Lee}, \binits{K.}},
\oauthor{\bsnm{Toutanova}, \binits{K.}}:
Bert: Pre-training of deep bidirectional transformers for language
  understanding.
arXiv preprint arXiv:1810.04805
(2018)
\end{botherref}
\endbibitem

\bibitem{radford2018improving}
\begin{botherref}
\oauthor{\bsnm{Radford}, \binits{A.}},
\oauthor{\bsnm{Narasimhan}, \binits{K.}},
\oauthor{\bsnm{Salimans}, \binits{T.}},
\oauthor{\bsnm{Sutskever}, \binits{I.}}:
Improving language understanding by generative pre-training
(2018)
\end{botherref}
\endbibitem

\bibitem{lin2003symbolic}
\begin{bchapter}
\bauthor{\bsnm{Lin}, \binits{J.}},
\bauthor{\bsnm{Keogh}, \binits{E.}},
\bauthor{\bsnm{Lonardi}, \binits{S.}},
\bauthor{\bsnm{Chiu}, \binits{B.}}:
\bctitle{A symbolic representation of time series, with implications for
  streaming algorithms}.
In: \bbtitle{Proceedings of the 8th ACM SIGMOD Workshop on Research Issues in
  Data Mining and Knowledge Discovery},
pp. \bfpage{2}--\blpage{11}
(\byear{2003})
\end{bchapter}
\endbibitem

\bibitem{ye2009time}
\begin{bchapter}
\bauthor{\bsnm{Ye}, \binits{L.}},
\bauthor{\bsnm{Keogh}, \binits{E.}}:
\bctitle{Time series shapelets: a new primitive for data mining}.
In: \bbtitle{Proceedings of the 15th ACM SIGKDD International Conference on
  Knowledge Discovery and Data Mining},
pp. \bfpage{947}--\blpage{956}
(\byear{2009})
\end{bchapter}
\endbibitem

\bibitem{du2019techniques}
\begin{barticle}
\bauthor{\bsnm{Du}, \binits{M.}},
\bauthor{\bsnm{Liu}, \binits{N.}},
\bauthor{\bsnm{Hu}, \binits{X.}}:
\batitle{Techniques for interpretable machine learning}.
\bjtitle{Communications of the ACM}
\bvolume{63}(\bissue{1}),
\bfpage{68}--\blpage{77}
(\byear{2019})
\end{barticle}
\endbibitem

\bibitem{beebe2021efficient}
\begin{barticle}
\bauthor{\bsnm{Beebe-Wang}, \binits{N.}},
\bauthor{\bsnm{Okeson}, \binits{A.}},
\bauthor{\bsnm{Althoff}, \binits{T.}},
\bauthor{\bsnm{Lee}, \binits{S.-I.}}:
\batitle{Efficient and explainable risk assessments for imminent dementia in an
  aging cohort study}.
\bjtitle{IEEE Journal of Biomedical and Health Informatics}
\bvolume{25}(\bissue{7}),
\bfpage{2409}--\blpage{2420}
(\byear{2021})
\end{barticle}
\endbibitem

\bibitem{a2012overview}
\begin{barticle}
\bauthor{\bsnm{A~Bennett}, \binits{D.}},
\bauthor{\bsnm{A~Schneider}, \binits{J.}},
\bauthor{\bsnm{S~Buchman}, \binits{A.}},
\bauthor{\bsnm{L~Barnes}, \binits{L.}},
\bauthor{\bsnm{A~Boyle}, \binits{P.}},
\bauthor{\bsnm{S~Wilson}, \binits{R.}}:
\batitle{Overview and findings from the rush memory and aging project}.
\bjtitle{Current Alzheimer Research}
\bvolume{9}(\bissue{6}),
\bfpage{646}--\blpage{663}
(\byear{2012})
\end{barticle}
\endbibitem

\bibitem{sha2021smile}
\begin{barticle}
\bauthor{\bsnm{Sha}, \binits{C.}},
\bauthor{\bsnm{Cuperlovic-Culf}, \binits{M.}},
\bauthor{\bsnm{Hu}, \binits{T.}}:
\batitle{Smile: systems metabolomics using interpretable learning and
  evolution}.
\bjtitle{BMC bioinformatics}
\bvolume{22}(\bissue{1}),
\bfpage{1}--\blpage{17}
(\byear{2021})
\end{barticle}
\endbibitem

\bibitem{wang2014plasma}
\begin{barticle}
\bauthor{\bsnm{Wang}, \binits{G.}},
\bauthor{\bsnm{Zhou}, \binits{Y.}},
\bauthor{\bsnm{Huang}, \binits{F.-J.}},
\bauthor{\bsnm{Tang}, \binits{H.-D.}},
\bauthor{\bsnm{Xu}, \binits{X.-H.}},
\bauthor{\bsnm{Liu}, \binits{J.-J.}},
\bauthor{\bsnm{Wang}, \binits{Y.}},
\bauthor{\bsnm{Deng}, \binits{Y.-L.}},
\bauthor{\bsnm{Ren}, \binits{R.-J.}},
\bauthor{\bsnm{Xu}, \binits{W.}}, \betal:
\batitle{Plasma metabolite profiles of alzheimer’s disease and mild cognitive
  impairment}.
\bjtitle{Journal of Proteome Research}
\bvolume{13}(\bissue{5}),
\bfpage{2649}--\blpage{2658}
(\byear{2014})
\end{barticle}
\endbibitem

\bibitem{kim2021interpretable}
\begin{barticle}
\bauthor{\bsnm{Kim}, \binits{S.-H.}},
\bauthor{\bsnm{Jeon}, \binits{E.-T.}},
\bauthor{\bsnm{Yu}, \binits{S.}},
\bauthor{\bsnm{Oh}, \binits{K.}},
\bauthor{\bsnm{Kim}, \binits{C.K.}},
\bauthor{\bsnm{Song}, \binits{T.-J.}},
\bauthor{\bsnm{Kim}, \binits{Y.-J.}},
\bauthor{\bsnm{Heo}, \binits{S.H.}},
\bauthor{\bsnm{Park}, \binits{K.-Y.}},
\bauthor{\bsnm{Kim}, \binits{J.-M.}}, \betal:
\batitle{Interpretable machine learning for early neurological deterioration
  prediction in atrial fibrillation-related stroke}.
\bjtitle{Scientific reports}
\bvolume{11}(\bissue{1}),
\bfpage{1}--\blpage{9}
(\byear{2021})
\end{barticle}
\endbibitem

\bibitem{jung2019long}
\begin{barticle}
\bauthor{\bsnm{Jung}, \binits{J.-M.}},
\bauthor{\bsnm{Kim}, \binits{Y.-H.}},
\bauthor{\bsnm{Yu}, \binits{S.}},
\bauthor{\bsnm{Kyungmi}, \binits{O.}},
\bauthor{\bsnm{Kim}, \binits{C.K.}},
\bauthor{\bsnm{Song}, \binits{T.-J.}},
\bauthor{\bsnm{Kim}, \binits{Y.-J.}},
\bauthor{\bsnm{Kim}, \binits{B.J.}},
\bauthor{\bsnm{Heo}, \binits{S.H.}},
\bauthor{\bsnm{Park}, \binits{K.-Y.}}, \betal:
\batitle{Long-term outcomes of real-world korean patients with
  atrial-fibrillation-related stroke and severely decreased ejection fraction}.
\bjtitle{Journal of Clinical Neurology}
\bvolume{15}(\bissue{4}),
\bfpage{545}--\blpage{554}
(\byear{2019})
\end{barticle}
\endbibitem

\bibitem{rashed2021clinically}
\begin{barticle}
\bauthor{\bsnm{Rashed-Al-Mahfuz}, \binits{M.}},
\bauthor{\bsnm{Haque}, \binits{A.}},
\bauthor{\bsnm{Azad}, \binits{A.}},
\bauthor{\bsnm{Alyami}, \binits{S.A.}},
\bauthor{\bsnm{Quinn}, \binits{J.M.}},
\bauthor{\bsnm{Moni}, \binits{M.A.}}:
\batitle{Clinically applicable machine learning approaches to identify
  attributes of chronic kidney disease (ckd) for use in low-cost diagnostic
  screening}.
\bjtitle{IEEE Journal of Translational Engineering in Health and Medicine}
\bvolume{9},
\bfpage{1}--\blpage{11}
(\byear{2021})
\end{barticle}
\endbibitem

\bibitem{rubini2015uci}
\begin{botherref}
\oauthor{\bsnm{Rubini}, \binits{L.J.}},
\oauthor{\bsnm{Eswaran}, \binits{P.}}:
Uci chronic kidney disease.
School Inf. Comput. Sci., Univ. California, Irvine, Irvine, CA, USA
(2015)
\end{botherref}
\endbibitem

\bibitem{pang2019understanding}
\begin{bchapter}
\bauthor{\bsnm{Pang}, \binits{X.}},
\bauthor{\bsnm{Forrest}, \binits{C.B.}},
\bauthor{\bsnm{L{\^e}-Scherban}, \binits{F.}},
\bauthor{\bsnm{Masino}, \binits{A.J.}}:
\bctitle{Understanding early childhood obesity via interpretation of machine
  learning model predictions}.
In: \bbtitle{2019 18th IEEE International Conference On Machine Learning And
  Applications (ICMLA)},
pp. \bfpage{1438}--\blpage{1443}
(\byear{2019}).
\bcomment{IEEE}
\end{bchapter}
\endbibitem

\bibitem{zeng2021explainable}
\begin{barticle}
\bauthor{\bsnm{Zeng}, \binits{X.}},
\bauthor{\bsnm{Hu}, \binits{Y.}},
\bauthor{\bsnm{Shu}, \binits{L.}},
\bauthor{\bsnm{Li}, \binits{J.}},
\bauthor{\bsnm{Duan}, \binits{H.}},
\bauthor{\bsnm{Shu}, \binits{Q.}},
\bauthor{\bsnm{Li}, \binits{H.}}:
\batitle{Explainable machine-learning predictions for complications after
  pediatric congenital heart surgery}.
\bjtitle{Scientific reports}
\bvolume{11}(\bissue{1}),
\bfpage{1}--\blpage{11}
(\byear{2021})
\end{barticle}
\endbibitem

\bibitem{zhang2021explainable}
\begin{barticle}
\bauthor{\bsnm{Zhang}, \binits{Y.}},
\bauthor{\bsnm{Yang}, \binits{D.}},
\bauthor{\bsnm{Liu}, \binits{Z.}},
\bauthor{\bsnm{Chen}, \binits{C.}},
\bauthor{\bsnm{Ge}, \binits{M.}},
\bauthor{\bsnm{Li}, \binits{X.}},
\bauthor{\bsnm{Luo}, \binits{T.}},
\bauthor{\bsnm{Wu}, \binits{Z.}},
\bauthor{\bsnm{Shi}, \binits{C.}},
\bauthor{\bsnm{Wang}, \binits{B.}}, \betal:
\batitle{An explainable supervised machine learning predictor of acute kidney
  injury after adult deceased donor liver transplantation}.
\bjtitle{Journal of translational medicine}
\bvolume{19}(\bissue{1}),
\bfpage{1}--\blpage{15}
(\byear{2021})
\end{barticle}
\endbibitem

\bibitem{shashikumar2021deepaise}
\begin{barticle}
\bauthor{\bsnm{Shashikumar}, \binits{S.P.}},
\bauthor{\bsnm{Josef}, \binits{C.S.}},
\bauthor{\bsnm{Sharma}, \binits{A.}},
\bauthor{\bsnm{Nemati}, \binits{S.}}:
\batitle{Deepaise--an interpretable and recurrent neural survival model for
  early prediction of sepsis}.
\bjtitle{Artificial Intelligence in Medicine}
\bvolume{113},
\bfpage{102036}
(\byear{2021})
\end{barticle}
\endbibitem

\bibitem{cox1992regression}
\begin{barticle}
\bauthor{\bsnm{Cox}, \binits{D.R.}}:
\batitle{Regression models and life-tables. breakthroughs in statistics}.
\bjtitle{Stat. Soc}
\bvolume{372},
\bfpage{527}--\blpage{541}
(\byear{1992})
\end{barticle}
\endbibitem

\bibitem{johnson2016mimic}
\begin{barticle}
\bauthor{\bsnm{Johnson}, \binits{A.E.}},
\bauthor{\bsnm{Pollard}, \binits{T.J.}},
\bauthor{\bsnm{Shen}, \binits{L.}},
\bauthor{\bsnm{Lehman}, \binits{L.-w.H.}},
\bauthor{\bsnm{Feng}, \binits{M.}},
\bauthor{\bsnm{Ghassemi}, \binits{M.}},
\bauthor{\bsnm{Moody}, \binits{B.}},
\bauthor{\bsnm{Szolovits}, \binits{P.}},
\bauthor{\bsnm{Anthony~Celi}, \binits{L.}},
\bauthor{\bsnm{Mark}, \binits{R.G.}}:
\batitle{Mimic-iii, a freely accessible critical care database}.
\bjtitle{Scientific data}
\bvolume{3}(\bissue{1}),
\bfpage{1}--\blpage{9}
(\byear{2016})
\end{barticle}
\endbibitem

\bibitem{sun2021attention}
\begin{barticle}
\bauthor{\bsnm{Sun}, \binits{Z.}},
\bauthor{\bsnm{Dong}, \binits{W.}},
\bauthor{\bsnm{Shi}, \binits{J.}},
\bauthor{\bsnm{He}, \binits{K.}},
\bauthor{\bsnm{Huang}, \binits{Z.}}:
\batitle{Attention-based deep recurrent model for survival prediction}.
\bjtitle{ACM Transactions on Computing for Healthcare}
\bvolume{2}(\bissue{4}),
\bfpage{1}--\blpage{18}
(\byear{2021})
\end{barticle}
\endbibitem

\bibitem{lemeshow2011applied}
\begin{bbook}
\bauthor{\bsnm{Lemeshow}, \binits{S.}},
\bauthor{\bsnm{May}, \binits{S.}},
\bauthor{\bsnm{Hosmer~Jr}, \binits{D.W.}}:
\bbtitle{Applied Survival Analysis: Regression Modeling of Time-to-Event Data}.
\bpublisher{John Wiley \& Sons}, \blocation{???}
(\byear{2011})
\end{bbook}
\endbibitem

\bibitem{knaus1995support}
\begin{barticle}
\bauthor{\bsnm{Knaus}, \binits{W.A.}},
\bauthor{\bsnm{Harrell}, \binits{F.E.}},
\bauthor{\bsnm{Lynn}, \binits{J.}},
\bauthor{\bsnm{Goldman}, \binits{L.}},
\bauthor{\bsnm{Phillips}, \binits{R.S.}},
\bauthor{\bsnm{Connors}, \binits{A.F.}},
\bauthor{\bsnm{Dawson}, \binits{N.V.}},
\bauthor{\bsnm{Fulkerson}, \binits{W.J.}},
\bauthor{\bsnm{Califf}, \binits{R.M.}},
\bauthor{\bsnm{Desbiens}, \binits{N.}}, \betal:
\batitle{The support prognostic model: Objective estimates of survival for
  seriously ill hospitalized adults}.
\bjtitle{Annals of internal medicine}
\bvolume{122}(\bissue{3}),
\bfpage{191}--\blpage{203}
(\byear{1995})
\end{barticle}
\endbibitem

\bibitem{curtis2012genomic}
\begin{barticle}
\bauthor{\bsnm{Curtis}, \binits{C.}},
\bauthor{\bsnm{Shah}, \binits{S.P.}},
\bauthor{\bsnm{Chin}, \binits{S.-F.}},
\bauthor{\bsnm{Turashvili}, \binits{G.}},
\bauthor{\bsnm{Rueda}, \binits{O.M.}},
\bauthor{\bsnm{Dunning}, \binits{M.J.}},
\bauthor{\bsnm{Speed}, \binits{D.}},
\bauthor{\bsnm{Lynch}, \binits{A.G.}},
\bauthor{\bsnm{Samarajiwa}, \binits{S.}},
\bauthor{\bsnm{Yuan}, \binits{Y.}}, \betal:
\batitle{The genomic and transcriptomic architecture of 2,000 breast tumours
  reveals novel subgroups}.
\bjtitle{Nature}
\bvolume{486}(\bissue{7403}),
\bfpage{346}--\blpage{352}
(\byear{2012})
\end{barticle}
\endbibitem

\bibitem{zheng2020tracer}
\begin{bchapter}
\bauthor{\bsnm{Zheng}, \binits{K.}},
\bauthor{\bsnm{Cai}, \binits{S.}},
\bauthor{\bsnm{Chua}, \binits{H.R.}},
\bauthor{\bsnm{Wang}, \binits{W.}},
\bauthor{\bsnm{Ngiam}, \binits{K.Y.}},
\bauthor{\bsnm{Ooi}, \binits{B.C.}}:
\bctitle{Tracer: A framework for facilitating accurate and interpretable
  analytics for high stakes applications}.
In: \bbtitle{Proceedings of the 2020 ACM SIGMOD International Conference on
  Management of Data},
pp. \bfpage{1747}--\blpage{1763}
(\byear{2020})
\end{bchapter}
\endbibitem

\bibitem{perez2018film}
\begin{bchapter}
\bauthor{\bsnm{Perez}, \binits{E.}},
\bauthor{\bsnm{Strub}, \binits{F.}},
\bauthor{\bsnm{De~Vries}, \binits{H.}},
\bauthor{\bsnm{Dumoulin}, \binits{V.}},
\bauthor{\bsnm{Courville}, \binits{A.}}:
\bctitle{Film: Visual reasoning with a general conditioning layer}.
In: \bbtitle{Proceedings of the AAAI Conference on Artificial Intelligence},
vol. \bseriesno{32}
(\byear{2018})
\end{bchapter}
\endbibitem

\bibitem{kwon2018retainvis}
\begin{barticle}
\bauthor{\bsnm{Kwon}, \binits{B.C.}},
\bauthor{\bsnm{Choi}, \binits{M.-J.}},
\bauthor{\bsnm{Kim}, \binits{J.T.}},
\bauthor{\bsnm{Choi}, \binits{E.}},
\bauthor{\bsnm{Kim}, \binits{Y.B.}},
\bauthor{\bsnm{Kwon}, \binits{S.}},
\bauthor{\bsnm{Sun}, \binits{J.}},
\bauthor{\bsnm{Choo}, \binits{J.}}:
\batitle{Retainvis: Visual analytics with interpretable and interactive
  recurrent neural networks on electronic medical records}.
\bjtitle{IEEE transactions on visualization and computer graphics}
\bvolume{25}(\bissue{1}),
\bfpage{299}--\blpage{309}
(\byear{2018})
\end{barticle}
\endbibitem

\bibitem{choi2016retain}
\begin{botherref}
\oauthor{\bsnm{Choi}, \binits{E.}},
\oauthor{\bsnm{Bahadori}, \binits{M.T.}},
\oauthor{\bsnm{Sun}, \binits{J.}},
\oauthor{\bsnm{Kulas}, \binits{J.}},
\oauthor{\bsnm{Schuetz}, \binits{A.}},
\oauthor{\bsnm{Stewart}, \binits{W.}}:
Retain: An interpretable predictive model for healthcare using reverse time
  attention mechanism.
Advances in neural information processing systems
\textbf{29}
(2016)
\end{botherref}
\endbibitem

\bibitem{kim2014guide}
\begin{botherref}
\oauthor{\bsnm{Kim}, \binits{L.}},
\oauthor{\bsnm{Kim}, \binits{J.-A.}},
\oauthor{\bsnm{Kim}, \binits{S.}}:
A guide for the utilization of health insurance review and assessment service
  national patient samples.
Epidemiology and health
\textbf{36}
(2014)
\end{botherref}
\endbibitem

\bibitem{lu2020explainable}
\begin{barticle}
\bauthor{\bsnm{Lu}, \binits{J.}},
\bauthor{\bsnm{Jin}, \binits{R.}},
\bauthor{\bsnm{Song}, \binits{E.}},
\bauthor{\bsnm{Alrashoud}, \binits{M.}},
\bauthor{\bsnm{Al-Mutib}, \binits{K.N.}},
\bauthor{\bsnm{Al-Rakhami}, \binits{M.S.}}:
\batitle{An explainable system for diagnosis and prognosis of covid-19}.
\bjtitle{IEEE Internet of Things Journal}
\bvolume{8}(\bissue{21}),
\bfpage{15839}--\blpage{15846}
(\byear{2020})
\end{barticle}
\endbibitem

\bibitem{yan2020interpretable}
\begin{barticle}
\bauthor{\bsnm{Yan}, \binits{L.}},
\bauthor{\bsnm{Zhang}, \binits{H.-T.}},
\bauthor{\bsnm{Goncalves}, \binits{J.}},
\bauthor{\bsnm{Xiao}, \binits{Y.}},
\bauthor{\bsnm{Wang}, \binits{M.}},
\bauthor{\bsnm{Guo}, \binits{Y.}},
\bauthor{\bsnm{Sun}, \binits{C.}},
\bauthor{\bsnm{Tang}, \binits{X.}},
\bauthor{\bsnm{Jing}, \binits{L.}},
\bauthor{\bsnm{Zhang}, \binits{M.}}, \betal:
\batitle{An interpretable mortality prediction model for covid-19 patients}.
\bjtitle{Nature machine intelligence}
\bvolume{2}(\bissue{5}),
\bfpage{283}--\blpage{288}
(\byear{2020})
\end{barticle}
\endbibitem

\bibitem{weiss2010fundamentals}
\begin{botherref}
\oauthor{\bsnm{Weiss}, \binits{S.M.}},
\oauthor{\bsnm{Indurkhya}, \binits{N.}},
\oauthor{\bsnm{Zhang}, \binits{T.}}:
Fundamentals of predictive text mining: Springer science \& business media
(2010)
\end{botherref}
\endbibitem

\bibitem{gupta2021interpretable}
\begin{bchapter}
\bauthor{\bsnm{Gupta}, \binits{A.}},
\bauthor{\bsnm{Jain}, \binits{J.}},
\bauthor{\bsnm{Poundrik}, \binits{S.}},
\bauthor{\bsnm{Shetty}, \binits{M.K.}},
\bauthor{\bsnm{Girish}, \binits{M.}},
\bauthor{\bsnm{Gupta}, \binits{M.D.}}:
\bctitle{Interpretable ai model-based predictions of ecg changes in
  covid-recovered patients}.
In: \bbtitle{2021 4th International Conference on Bio-Engineering for Smart
  Technologies (BioSMART)},
pp. \bfpage{1}--\blpage{5}
(\byear{2021}).
\bcomment{IEEE}
\end{bchapter}
\endbibitem

\bibitem{bonaz2020targeting}
\begin{barticle}
\bauthor{\bsnm{Bonaz}, \binits{B.}},
\bauthor{\bsnm{Sinniger}, \binits{V.}},
\bauthor{\bsnm{Pellissier}, \binits{S.}}:
\batitle{Targeting the cholinergic anti-inflammatory pathway with vagus nerve
  stimulation in patients with covid-19?}
\bjtitle{Bioelectronic medicine}
\bvolume{6}(\bissue{1}),
\bfpage{1}--\blpage{7}
(\byear{2020})
\end{barticle}
\endbibitem

\bibitem{pal2021pay}
\begin{bchapter}
\bauthor{\bsnm{Pal}, \binits{A.}},
\bauthor{\bsnm{Sankarasubbu}, \binits{M.}}:
\bctitle{Pay attention to the cough: Early diagnosis of covid-19 using
  interpretable symptoms embeddings with cough sound signal processing}.
In: \bbtitle{Proceedings of the 36th Annual ACM Symposium on Applied
  Computing},
pp. \bfpage{620}--\blpage{628}
(\byear{2021})
\end{bchapter}
\endbibitem

\bibitem{arik2021tabnet}
\begin{bchapter}
\bauthor{\bsnm{Ar{\i}k}, \binits{S.O.}},
\bauthor{\bsnm{Pfister}, \binits{T.}}:
\bctitle{Tabnet: Attentive interpretable tabular learning}.
In: \bbtitle{AAAI},
vol. \bseriesno{35},
pp. \bfpage{6679}--\blpage{6687}
(\byear{2021})
\end{bchapter}
\endbibitem

\bibitem{van2008visualizing}
\begin{botherref}
\oauthor{\bparticle{Van~der} \bsnm{Maaten}, \binits{L.}},
\oauthor{\bsnm{Hinton}, \binits{G.}}:
Visualizing data using t-sne.
Journal of machine learning research
\textbf{9}(11)
(2008)
\end{botherref}
\endbibitem

\bibitem{ivaturi2021comprehensive}
\begin{barticle}
\bauthor{\bsnm{Ivaturi}, \binits{P.}},
\bauthor{\bsnm{Gadaleta}, \binits{M.}},
\bauthor{\bsnm{Pandey}, \binits{A.C.}},
\bauthor{\bsnm{Pazzani}, \binits{M.}},
\bauthor{\bsnm{Steinhubl}, \binits{S.R.}},
\bauthor{\bsnm{Quer}, \binits{G.}}:
\batitle{A comprehensive explanation framework for biomedical time series
  classification}.
\bjtitle{IEEE Journal of Biomedical and Health Informatics}
\bvolume{25}(\bissue{7}),
\bfpage{2398}--\blpage{2408}
(\byear{2021})
\end{barticle}
\endbibitem

\bibitem{howard2017mobilenets}
\begin{botherref}
\oauthor{\bsnm{Howard}, \binits{A.G.}},
\oauthor{\bsnm{Zhu}, \binits{M.}},
\oauthor{\bsnm{Chen}, \binits{B.}},
\oauthor{\bsnm{Kalenichenko}, \binits{D.}},
\oauthor{\bsnm{Wang}, \binits{W.}},
\oauthor{\bsnm{Weyand}, \binits{T.}},
\oauthor{\bsnm{Andreetto}, \binits{M.}},
\oauthor{\bsnm{Adam}, \binits{H.}}:
Mobilenets: Efficient convolutional neural networks for mobile vision
  applications.
arXiv preprint arXiv:1704.04861
(2017)
\end{botherref}
\endbibitem

\bibitem{clifford2017af}
\begin{bchapter}
\bauthor{\bsnm{Clifford}, \binits{G.D.}},
\bauthor{\bsnm{Liu}, \binits{C.}},
\bauthor{\bsnm{Moody}, \binits{B.}},
\bauthor{\bsnm{Li-wei}, \binits{H.L.}},
\bauthor{\bsnm{Silva}, \binits{I.}},
\bauthor{\bsnm{Li}, \binits{Q.}},
\bauthor{\bsnm{Johnson}, \binits{A.}},
\bauthor{\bsnm{Mark}, \binits{R.G.}}:
\bctitle{Af classification from a short single lead ecg recording: The
  physionet/computing in cardiology challenge 2017}.
In: \bbtitle{2017 Computing in Cardiology (CinC)},
pp. \bfpage{1}--\blpage{4}
(\byear{2017}).
\bcomment{IEEE}
\end{bchapter}
\endbibitem

\bibitem{mousavi2020han}
\begin{barticle}
\bauthor{\bsnm{Mousavi}, \binits{S.}},
\bauthor{\bsnm{Afghah}, \binits{F.}},
\bauthor{\bsnm{Acharya}, \binits{U.R.}}:
\batitle{Han-ecg: An interpretable atrial fibrillation detection model using
  hierarchical attention networks}.
\bjtitle{Computers in biology and medicine}
\bvolume{127},
\bfpage{104057}
(\byear{2020})
\end{barticle}
\endbibitem

\bibitem{dissanayake2020robust}
\begin{barticle}
\bauthor{\bsnm{Dissanayake}, \binits{T.}},
\bauthor{\bsnm{Fernando}, \binits{T.}},
\bauthor{\bsnm{Denman}, \binits{S.}},
\bauthor{\bsnm{Sridharan}, \binits{S.}},
\bauthor{\bsnm{Ghaemmaghami}, \binits{H.}},
\bauthor{\bsnm{Fookes}, \binits{C.}}:
\batitle{A robust interpretable deep learning classifier for heart anomaly
  detection without segmentation}.
\bjtitle{IEEE Journal of Biomedical and Health Informatics}
\bvolume{25}(\bissue{6}),
\bfpage{2162}--\blpage{2171}
(\byear{2020})
\end{barticle}
\endbibitem

\bibitem{clifford2017recent}
\begin{barticle}
\bauthor{\bsnm{Clifford}, \binits{G.D.}},
\bauthor{\bsnm{Liu}, \binits{C.}},
\bauthor{\bsnm{Moody}, \binits{B.E.}},
\bauthor{\bsnm{Roig}, \binits{J.M.}},
\bauthor{\bsnm{Schmidt}, \binits{S.E.}},
\bauthor{\bsnm{Li}, \binits{Q.}},
\bauthor{\bsnm{Silva}, \binits{I.}},
\bauthor{\bsnm{Mark}, \binits{R.G.}}:
\batitle{Recent advances in heart sound analysis}.
\bjtitle{Physiological measurement}
\bvolume{38},
\bfpage{10}--\blpage{25}
(\byear{2017})
\end{barticle}
\endbibitem

\bibitem{goldberger2000physiobank}
\begin{barticle}
\bauthor{\bsnm{Goldberger}, \binits{A.L.}},
\bauthor{\bsnm{Amaral}, \binits{L.A.}},
\bauthor{\bsnm{Glass}, \binits{L.}},
\bauthor{\bsnm{Hausdorff}, \binits{J.M.}},
\bauthor{\bsnm{Ivanov}, \binits{P.C.}},
\bauthor{\bsnm{Mark}, \binits{R.G.}},
\bauthor{\bsnm{Mietus}, \binits{J.E.}},
\bauthor{\bsnm{Moody}, \binits{G.B.}},
\bauthor{\bsnm{Peng}, \binits{C.-K.}},
\bauthor{\bsnm{Stanley}, \binits{H.E.}}:
\batitle{Physiobank, physiotoolkit, and physionet: components of a new research
  resource for complex physiologic signals}.
\bjtitle{circulation}
\bvolume{101}(\bissue{23}),
\bfpage{215}--\blpage{220}
(\byear{2000})
\end{barticle}
\endbibitem

\bibitem{arrotta2022dexar}
\begin{barticle}
\bauthor{\bsnm{Arrotta}, \binits{L.}},
\bauthor{\bsnm{Civitarese}, \binits{G.}},
\bauthor{\bsnm{Bettini}, \binits{C.}}:
\batitle{Dexar: Deep explainable sensor-based activity recognition in
  smart-home environments}.
\bjtitle{Proceedings of the ACM on Interactive, Mobile, Wearable and Ubiquitous
  Technologies}
\bvolume{6}(\bissue{1}),
\bfpage{1}--\blpage{30}
(\byear{2022})
\end{barticle}
\endbibitem

\bibitem{khodabandehloo2021healthxai}
\begin{barticle}
\bauthor{\bsnm{Khodabandehloo}, \binits{E.}},
\bauthor{\bsnm{Riboni}, \binits{D.}},
\bauthor{\bsnm{Alimohammadi}, \binits{A.}}:
\batitle{Healthxai: Collaborative and explainable ai for supporting early
  diagnosis of cognitive decline}.
\bjtitle{Future Generation Computer Systems}
\bvolume{116},
\bfpage{168}--\blpage{189}
(\byear{2021})
\end{barticle}
\endbibitem

\bibitem{filtjens2021modelling}
\begin{barticle}
\bauthor{\bsnm{Filtjens}, \binits{B.}},
\bauthor{\bsnm{Ginis}, \binits{P.}},
\bauthor{\bsnm{Nieuwboer}, \binits{A.}},
\bauthor{\bsnm{Afzal}, \binits{M.R.}},
\bauthor{\bsnm{Spildooren}, \binits{J.}},
\bauthor{\bsnm{Vanrumste}, \binits{B.}},
\bauthor{\bsnm{Slaets}, \binits{P.}}:
\batitle{Modelling and identification of characteristic kinematic features
  preceding freezing of gait with convolutional neural networks and layer-wise
  relevance propagation}.
\bjtitle{BMC medical informatics and decision making}
\bvolume{21}(\bissue{1}),
\bfpage{1}--\blpage{11}
(\byear{2021})
\end{barticle}
\endbibitem

\bibitem{spildooren2010freezing}
\begin{barticle}
\bauthor{\bsnm{Spildooren}, \binits{J.}},
\bauthor{\bsnm{Vercruysse}, \binits{S.}},
\bauthor{\bsnm{Desloovere}, \binits{K.}},
\bauthor{\bsnm{Vandenberghe}, \binits{W.}},
\bauthor{\bsnm{Kerckhofs}, \binits{E.}},
\bauthor{\bsnm{Nieuwboer}, \binits{A.}}:
\batitle{Freezing of gait in parkinson's disease: the impact of dual-tasking
  and turning}.
\bjtitle{Movement Disorders}
\bvolume{25}(\bissue{15}),
\bfpage{2563}--\blpage{2570}
(\byear{2010})
\end{barticle}
\endbibitem

\bibitem{thimoteo2022explainable}
\begin{botherref}
\oauthor{\bsnm{Thimoteo}, \binits{L.M.}},
\oauthor{\bsnm{Vellasco}, \binits{M.M.}},
\oauthor{\bsnm{Amaral}, \binits{J.}},
\oauthor{\bsnm{Figueiredo}, \binits{K.}},
\oauthor{\bsnm{Yokoyama}, \binits{C.L.}},
\oauthor{\bsnm{Marques}, \binits{E.}}:
Explainable artificial intelligence for covid-19 diagnosis through blood test
  variables.
Journal of Control, Automation and Electrical Systems,
1--20
(2022)
\end{botherref}
\endbibitem

\bibitem{nori2019interpretml}
\begin{botherref}
\oauthor{\bsnm{Nori}, \binits{H.}},
\oauthor{\bsnm{Jenkins}, \binits{S.}},
\oauthor{\bsnm{Koch}, \binits{P.}},
\oauthor{\bsnm{Caruana}, \binits{R.}}:
Interpretml: A unified framework for machine learning interpretability.
arXiv preprint arXiv:1909.09223
(2019)
\end{botherref}
\endbibitem

\bibitem{alves2021explaining}
\begin{barticle}
\bauthor{\bsnm{Alves}, \binits{M.A.}},
\bauthor{\bsnm{Castro}, \binits{G.Z.}},
\bauthor{\bsnm{Oliveira}, \binits{B.A.S.}},
\bauthor{\bsnm{Ferreira}, \binits{L.A.}},
\bauthor{\bsnm{Ram{\'\i}rez}, \binits{J.A.}},
\bauthor{\bsnm{Silva}, \binits{R.}},
\bauthor{\bsnm{Guimar{\~a}es}, \binits{F.G.}}:
\batitle{Explaining machine learning based diagnosis of covid-19 from routine
  blood tests with decision trees and criteria graphs}.
\bjtitle{Computers in Biology and Medicine}
\bvolume{132},
\bfpage{104335}
(\byear{2021})
\end{barticle}
\endbibitem

\bibitem{leung2021explainable}
\begin{bchapter}
\bauthor{\bsnm{Leung}, \binits{C.K.}},
\bauthor{\bsnm{Fung}, \binits{D.L.}},
\bauthor{\bsnm{Mai}, \binits{D.}},
\bauthor{\bsnm{Wen}, \binits{Q.}},
\bauthor{\bsnm{Tran}, \binits{J.}},
\bauthor{\bsnm{Souza}, \binits{J.}}:
\bctitle{Explainable data analytics for disease and healthcare informatics}.
In: \bbtitle{25th International Database Engineering \& Applications
  Symposium},
pp. \bfpage{65}--\blpage{74}
(\byear{2021})
\end{bchapter}
\endbibitem

\bibitem{okay2021interpretable}
\begin{botherref}
\oauthor{\bsnm{Okay}, \binits{F.Y.}},
\oauthor{\bsnm{Y{\i}ld{\i}r{\i}m}, \binits{M.}},
\oauthor{\bsnm{{\"O}zdemir}, \binits{S.}}:
Interpretable machine learning: A case study of healthcare.
In: 2021 International Symposium on Networks, Computers and Communications
  (ISNCC),
pp. 1--6.
IEEE
\end{botherref}
\endbibitem

\bibitem{oba2021interpretable}
\begin{bchapter}
\bauthor{\bsnm{Oba}, \binits{Y.}},
\bauthor{\bsnm{Tezuka}, \binits{T.}},
\bauthor{\bsnm{Sanuki}, \binits{M.}},
\bauthor{\bsnm{Wagatsuma}, \binits{Y.}}:
\bctitle{Interpretable prediction of diabetes from tabular health screening
  records using an attentional neural network}.
In: \bbtitle{2021 IEEE 8th International Conference on Data Science and
  Advanced Analytics (DSAA)},
pp. \bfpage{1}--\blpage{11}
(\byear{2021}).
\bcomment{IEEE}
\end{bchapter}
\endbibitem

\bibitem{elshawi2019interpretability}
\begin{barticle}
\bauthor{\bsnm{Elshawi}, \binits{R.}},
\bauthor{\bsnm{Al-Mallah}, \binits{M.H.}},
\bauthor{\bsnm{Sakr}, \binits{S.}}:
\batitle{On the interpretability of machine learning-based model for predicting
  hypertension}.
\bjtitle{BMC medical informatics and decision making}
\bvolume{19}(\bissue{1}),
\bfpage{1}--\blpage{32}
(\byear{2019})
\end{barticle}
\endbibitem

\bibitem{sakr2018using}
\begin{barticle}
\bauthor{\bsnm{Sakr}, \binits{S.}},
\bauthor{\bsnm{Elshawi}, \binits{R.}},
\bauthor{\bsnm{Ahmed}, \binits{A.}},
\bauthor{\bsnm{Qureshi}, \binits{W.T.}},
\bauthor{\bsnm{Brawner}, \binits{C.}},
\bauthor{\bsnm{Keteyian}, \binits{S.}},
\bauthor{\bsnm{Blaha}, \binits{M.J.}},
\bauthor{\bsnm{Al-Mallah}, \binits{M.H.}}:
\batitle{Using machine learning on cardiorespiratory fitness data for
  predicting hypertension: The henry ford exercise testing (fit) project}.
\bjtitle{PLoS One}
\bvolume{13}(\bissue{4}),
\bfpage{0195344}
(\byear{2018})
\end{barticle}
\endbibitem

\bibitem{friedman2008predictive}
\begin{botherref}
\oauthor{\bsnm{Friedman}, \binits{J.H.}},
\oauthor{\bsnm{Popescu}, \binits{B.E.}}:
Predictive learning via rule ensembles.
The annals of applied statistics,
916--954
(2008)
\end{botherref}
\endbibitem

\bibitem{seedat2020automated}
\begin{bchapter}
\bauthor{\bsnm{Seedat}, \binits{N.}},
\bauthor{\bsnm{Aharonson}, \binits{V.}},
\bauthor{\bsnm{Hamzany}, \binits{Y.}}:
\bctitle{Automated and interpretable m-health discrimination of vocal cord
  pathology enabled by machine learning}.
In: \bbtitle{2020 IEEE Asia-Pacific Conference on Computer Science and Data
  Engineering (CSDE)},
pp. \bfpage{1}--\blpage{6}
(\byear{2020}).
\bcomment{IEEE}
\end{bchapter}
\endbibitem

\bibitem{kapcia2021exmed}
\begin{bchapter}
\bauthor{\bsnm{Kapcia}, \binits{M.}},
\bauthor{\bsnm{Eshkiki}, \binits{H.}},
\bauthor{\bsnm{Duell}, \binits{J.}},
\bauthor{\bsnm{Fan}, \binits{X.}},
\bauthor{\bsnm{Zhou}, \binits{S.}},
\bauthor{\bsnm{Mora}, \binits{B.}}:
\bctitle{Exmed: An ai tool for experimenting explainable ai techniques on
  medical data analytics}.
In: \bbtitle{2021 IEEE 33rd International Conference on Tools with Artificial
  Intelligence (ICTAI)},
pp. \bfpage{841}--\blpage{845}
(\byear{2021}).
\bcomment{IEEE}
\end{bchapter}
\endbibitem

\bibitem{duell2021comparison}
\begin{bchapter}
\bauthor{\bsnm{Duell}, \binits{J.}},
\bauthor{\bsnm{Fan}, \binits{X.}},
\bauthor{\bsnm{Burnett}, \binits{B.}},
\bauthor{\bsnm{Aarts}, \binits{G.}},
\bauthor{\bsnm{Zhou}, \binits{S.-M.}}:
\bctitle{A comparison of explanations given by explainable artificial
  intelligence methods on analysing electronic health records}.
In: \bbtitle{2021 IEEE EMBS International Conference on Biomedical and Health
  Informatics (BHI)},
pp. \bfpage{1}--\blpage{4}
(\byear{2021}).
\bcomment{IEEE}
\end{bchapter}
\endbibitem

\bibitem{moncada2021explainable}
\begin{barticle}
\bauthor{\bsnm{Moncada-Torres}, \binits{A.}},
\bauthor{\bparticle{van} \bsnm{Maaren}, \binits{M.C.}},
\bauthor{\bsnm{Hendriks}, \binits{M.P.}},
\bauthor{\bsnm{Siesling}, \binits{S.}},
\bauthor{\bsnm{Geleijnse}, \binits{G.}}:
\batitle{Explainable machine learning can outperform cox regression predictions
  and provide insights in breast cancer survival}.
\bjtitle{Scientific reports}
\bvolume{11}(\bissue{1}),
\bfpage{1}--\blpage{13}
(\byear{2021})
\end{barticle}
\endbibitem

\bibitem{ishwaran2008random}
\begin{barticle}
\bauthor{\bsnm{Ishwaran}, \binits{H.}},
\bauthor{\bsnm{Kogalur}, \binits{U.B.}},
\bauthor{\bsnm{Blackstone}, \binits{E.H.}},
\bauthor{\bsnm{Lauer}, \binits{M.S.}}:
\batitle{Random survival forests}.
\bjtitle{The annals of applied statistics}
\bvolume{2}(\bissue{3}),
\bfpage{841}--\blpage{860}
(\byear{2008})
\end{barticle}
\endbibitem

\bibitem{polsterl2016efficient}
\begin{botherref}
\oauthor{\bsnm{P{\"o}lsterl}, \binits{S.}},
\oauthor{\bsnm{Navab}, \binits{N.}},
\oauthor{\bsnm{Katouzian}, \binits{A.}}:
An efficient training algorithm for kernel survival support vector machines.
arXiv preprint arXiv:1611.07054
(2016)
\end{botherref}
\endbibitem

\bibitem{ang2021interpretable}
\begin{bchapter}
\bauthor{\bsnm{Ang}, \binits{E.T.}},
\bauthor{\bsnm{Nambiar}, \binits{M.}},
\bauthor{\bsnm{Soh}, \binits{Y.S.}},
\bauthor{\bsnm{Tan}, \binits{V.Y.}}:
\bctitle{An interpretable intensive care unit mortality risk calculator}.
In: \bbtitle{2021 43rd Annual International Conference of the IEEE Engineering
  in Medicine \& Biology Society (EMBC)},
pp. \bfpage{4152}--\blpage{4158}
(\byear{2021}).
\bcomment{IEEE}
\end{bchapter}
\endbibitem

\bibitem{song2020cross}
\begin{barticle}
\bauthor{\bsnm{Song}, \binits{X.}},
\bauthor{\bsnm{Yu}, \binits{A.S.}},
\bauthor{\bsnm{Kellum}, \binits{J.A.}},
\bauthor{\bsnm{Waitman}, \binits{L.R.}},
\bauthor{\bsnm{Matheny}, \binits{M.E.}},
\bauthor{\bsnm{Simpson}, \binits{S.Q.}},
\bauthor{\bsnm{Hu}, \binits{Y.}},
\bauthor{\bsnm{Liu}, \binits{M.}}:
\batitle{Cross-site transportability of an explainable artificial intelligence
  model for acute kidney injury prediction}.
\bjtitle{Nature communications}
\bvolume{11}(\bissue{1}),
\bfpage{1}--\blpage{12}
(\byear{2020})
\end{barticle}
\endbibitem

\bibitem{waitman2014greater}
\begin{barticle}
\bauthor{\bsnm{Waitman}, \binits{L.R.}},
\bauthor{\bsnm{Aaronson}, \binits{L.S.}},
\bauthor{\bsnm{Nadkarni}, \binits{P.M.}},
\bauthor{\bsnm{Connolly}, \binits{D.W.}},
\bauthor{\bsnm{Campbell}, \binits{J.R.}}:
\batitle{The greater plains collaborative: a pcornet clinical research data
  network}.
\bjtitle{Journal of the American Medical Informatics Association}
\bvolume{21}(\bissue{4}),
\bfpage{637}--\blpage{641}
(\byear{2014})
\end{barticle}
\endbibitem

\bibitem{pan2010domain}
\begin{barticle}
\bauthor{\bsnm{Pan}, \binits{S.J.}},
\bauthor{\bsnm{Tsang}, \binits{I.W.}},
\bauthor{\bsnm{Kwok}, \binits{J.T.}},
\bauthor{\bsnm{Yang}, \binits{Q.}}:
\batitle{Domain adaptation via transfer component analysis}.
\bjtitle{IEEE transactions on neural networks}
\bvolume{22}(\bissue{2}),
\bfpage{199}--\blpage{210}
(\byear{2010})
\end{barticle}
\endbibitem

\bibitem{duckworth2021using}
\begin{barticle}
\bauthor{\bsnm{Duckworth}, \binits{C.}},
\bauthor{\bsnm{Chmiel}, \binits{F.P.}},
\bauthor{\bsnm{Burns}, \binits{D.K.}},
\bauthor{\bsnm{Zlatev}, \binits{Z.D.}},
\bauthor{\bsnm{White}, \binits{N.M.}},
\bauthor{\bsnm{Daniels}, \binits{T.W.}},
\bauthor{\bsnm{Kiuber}, \binits{M.}},
\bauthor{\bsnm{Boniface}, \binits{M.J.}}:
\batitle{Using explainable machine learning to characterise data drift and
  detect emergent health risks for emergency department admissions during
  covid-19}.
\bjtitle{Scientific reports}
\bvolume{11}(\bissue{1}),
\bfpage{1}--\blpage{10}
(\byear{2021})
\end{barticle}
\endbibitem

\bibitem{tahmassebi2020interpretable}
\begin{bchapter}
\bauthor{\bsnm{Tahmassebi}, \binits{A.}},
\bauthor{\bsnm{Martin}, \binits{J.}},
\bauthor{\bsnm{Meyer-Baese}, \binits{A.}},
\bauthor{\bsnm{Gandomi}, \binits{A.H.}}:
\bctitle{An interpretable deep learning framework for health monitoring
  systems: A case study of eye state detection using eeg signals}.
In: \bbtitle{2020 IEEE Symposium Series on Computational Intelligence (SSCI)},
pp. \bfpage{211}--\blpage{218}
(\byear{2020}).
\bcomment{IEEE}
\end{bchapter}
\endbibitem

\bibitem{antoniadi2021prediction}
\begin{barticle}
\bauthor{\bsnm{Antoniadi}, \binits{A.M.}},
\bauthor{\bsnm{Galvin}, \binits{M.}},
\bauthor{\bsnm{Heverin}, \binits{M.}},
\bauthor{\bsnm{Hardiman}, \binits{O.}},
\bauthor{\bsnm{Mooney}, \binits{C.}}:
\batitle{Prediction of caregiver quality of life in amyotrophic lateral
  sclerosis using explainable machine learning}.
\bjtitle{Scientific Reports}
\bvolume{11}(\bissue{1}),
\bfpage{1}--\blpage{13}
(\byear{2021})
\end{barticle}
\endbibitem

\bibitem{ward2021explainable}
\begin{barticle}
\bauthor{\bsnm{Ward}, \binits{I.R.}},
\bauthor{\bsnm{Wang}, \binits{L.}},
\bauthor{\bsnm{Lu}, \binits{J.}},
\bauthor{\bsnm{Bennamoun}, \binits{M.}},
\bauthor{\bsnm{Dwivedi}, \binits{G.}},
\bauthor{\bsnm{Sanfilippo}, \binits{F.M.}}:
\batitle{Explainable artificial intelligence for pharmacovigilance: What
  features are important when predicting adverse outcomes?}
\bjtitle{Computer Methods and Programs in Biomedicine}
\bvolume{212},
\bfpage{106415}
(\byear{2021})
\end{barticle}
\endbibitem

\bibitem{cheng2021vbridge}
\begin{barticle}
\bauthor{\bsnm{Cheng}, \binits{F.}},
\bauthor{\bsnm{Liu}, \binits{D.}},
\bauthor{\bsnm{Du}, \binits{F.}},
\bauthor{\bsnm{Lin}, \binits{Y.}},
\bauthor{\bsnm{Zytek}, \binits{A.}},
\bauthor{\bsnm{Li}, \binits{H.}},
\bauthor{\bsnm{Qu}, \binits{H.}},
\bauthor{\bsnm{Veeramachaneni}, \binits{K.}}:
\batitle{Vbridge: Connecting the dots between features and data to explain
  healthcare models}.
\bjtitle{IEEE Transactions on Visualization and Computer Graphics}
\bvolume{28}(\bissue{1}),
\bfpage{378}--\blpage{388}
(\byear{2021})
\end{barticle}
\endbibitem

\bibitem{zeng2020pic}
\begin{barticle}
\bauthor{\bsnm{Zeng}, \binits{X.}},
\bauthor{\bsnm{Yu}, \binits{G.}},
\bauthor{\bsnm{Lu}, \binits{Y.}},
\bauthor{\bsnm{Tan}, \binits{L.}},
\bauthor{\bsnm{Wu}, \binits{X.}},
\bauthor{\bsnm{Shi}, \binits{S.}},
\bauthor{\bsnm{Duan}, \binits{H.}},
\bauthor{\bsnm{Shu}, \binits{Q.}},
\bauthor{\bsnm{Li}, \binits{H.}}:
\batitle{Pic, a paediatric-specific intensive care database}.
\bjtitle{Scientific data}
\bvolume{7}(\bissue{1}),
\bfpage{1}--\blpage{8}
(\byear{2020})
\end{barticle}
\endbibitem

\bibitem{kumarakulasinghe2020evaluating}
\begin{bchapter}
\bauthor{\bsnm{Kumarakulasinghe}, \binits{N.B.}},
\bauthor{\bsnm{Blomberg}, \binits{T.}},
\bauthor{\bsnm{Liu}, \binits{J.}},
\bauthor{\bsnm{Leao}, \binits{A.S.}},
\bauthor{\bsnm{Papapetrou}, \binits{P.}}:
\bctitle{Evaluating local interpretable model-agnostic explanations on clinical
  machine learning classification models}.
In: \bbtitle{2020 IEEE 33rd International Symposium on Computer-Based Medical
  Systems (CBMS)},
pp. \bfpage{7}--\blpage{12}
(\byear{2020}).
\bcomment{IEEE}
\end{bchapter}
\endbibitem

\bibitem{reyna2019early}
\begin{bchapter}
\bauthor{\bsnm{Reyna}, \binits{M.A.}},
\bauthor{\bsnm{Josef}, \binits{C.}},
\bauthor{\bsnm{Seyedi}, \binits{S.}},
\bauthor{\bsnm{Jeter}, \binits{R.}},
\bauthor{\bsnm{Shashikumar}, \binits{S.P.}},
\bauthor{\bsnm{Westover}, \binits{M.B.}},
\bauthor{\bsnm{Sharma}, \binits{A.}},
\bauthor{\bsnm{Nemati}, \binits{S.}},
\bauthor{\bsnm{Clifford}, \binits{G.D.}}:
\bctitle{Early prediction of sepsis from clinical data: the physionet/computing
  in cardiology challenge 2019}.
In: \bbtitle{2019 Computing in Cardiology (CinC)},
p. \bfpage{1}
(\byear{2019}).
\bcomment{IEEE}
\end{bchapter}
\endbibitem

\bibitem{jian1998towards}
\begin{bchapter}
\bauthor{\bsnm{Jian}, \binits{J.-Y.}},
\bauthor{\bsnm{Bisantz}, \binits{A.M.}},
\bauthor{\bsnm{Drury}, \binits{C.G.}}:
\bctitle{Towards an empirically determined scale of trust in computerized
  systems: distinguishing concepts and types of trust}.
In: \bbtitle{Proceedings of the Human Factors and Ergonomics Society Annual
  Meeting},
vol. \bseriesno{42},
pp. \bfpage{501}--\blpage{505}
(\byear{1998}).
\bcomment{SAGE Publications Sage CA: Los Angeles, CA}
\end{bchapter}
\endbibitem

\bibitem{barda2020qualitative}
\begin{barticle}
\bauthor{\bsnm{Barda}, \binits{A.J.}},
\bauthor{\bsnm{Horvat}, \binits{C.M.}},
\bauthor{\bsnm{Hochheiser}, \binits{H.}}:
\batitle{A qualitative research framework for the design of user-centered
  displays of explanations for machine learning model predictions in
  healthcare}.
\bjtitle{BMC medical informatics and decision making}
\bvolume{20}(\bissue{1}),
\bfpage{1}--\blpage{16}
(\byear{2020})
\end{barticle}
\endbibitem

\bibitem{penafiel2020predicting}
\begin{barticle}
\bauthor{\bsnm{Penafiel}, \binits{S.}},
\bauthor{\bsnm{Baloian}, \binits{N.}},
\bauthor{\bsnm{Sanson}, \binits{H.}},
\bauthor{\bsnm{Pino}, \binits{J.A.}}:
\batitle{Predicting stroke risk with an interpretable classifier}.
\bjtitle{IEEE Access}
\bvolume{9},
\bfpage{1154}--\blpage{1166}
(\byear{2020})
\end{barticle}
\endbibitem

\bibitem{shafer2016dempster}
\begin{barticle}
\bauthor{\bsnm{Shafer}, \binits{G.}}:
\batitle{Dempster's rule of combination}.
\bjtitle{International Journal of Approximate Reasoning}
\bvolume{79},
\bfpage{26}--\blpage{40}
(\byear{2016})
\end{barticle}
\endbibitem

\bibitem{hatwell2020ada}
\begin{barticle}
\bauthor{\bsnm{Hatwell}, \binits{J.}},
\bauthor{\bsnm{Gaber}, \binits{M.M.}},
\bauthor{\bsnm{Atif~Azad}, \binits{R.M.}}:
\batitle{Ada-whips: explaining adaboost classification with applications in the
  health sciences}.
\bjtitle{BMC Medical Informatics and Decision Making}
\bvolume{20}(\bissue{1}),
\bfpage{1}--\blpage{25}
(\byear{2020})
\end{barticle}
\endbibitem

\bibitem{zhang2021explainability}
\begin{barticle}
\bauthor{\bsnm{Zhang}, \binits{O.}},
\bauthor{\bsnm{Ding}, \binits{C.}},
\bauthor{\bsnm{Pereira}, \binits{T.}},
\bauthor{\bsnm{Xiao}, \binits{R.}},
\bauthor{\bsnm{Gadhoumi}, \binits{K.}},
\bauthor{\bsnm{Meisel}, \binits{K.}},
\bauthor{\bsnm{Lee}, \binits{R.J.}},
\bauthor{\bsnm{Chen}, \binits{Y.}},
\bauthor{\bsnm{Hu}, \binits{X.}}:
\batitle{Explainability metrics of deep convolutional networks for
  photoplethysmography quality assessment}.
\bjtitle{IEEE Access}
\bvolume{9},
\bfpage{29736}--\blpage{29745}
(\byear{2021})
\end{barticle}
\endbibitem

\bibitem{he2016deep}
\begin{bchapter}
\bauthor{\bsnm{He}, \binits{K.}},
\bauthor{\bsnm{Zhang}, \binits{X.}},
\bauthor{\bsnm{Ren}, \binits{S.}},
\bauthor{\bsnm{Sun}, \binits{J.}}:
\bctitle{Deep residual learning for image recognition}.
In: \bbtitle{Proceedings of the IEEE Conference on Computer Vision and Pattern
  Recognition},
pp. \bfpage{770}--\blpage{778}
(\byear{2016})
\end{bchapter}
\endbibitem

\bibitem{drew2014insights}
\begin{barticle}
\bauthor{\bsnm{Drew}, \binits{B.J.}},
\bauthor{\bsnm{Harris}, \binits{P.}},
\bauthor{\bsnm{Z{\`e}gre-Hemsey}, \binits{J.K.}},
\bauthor{\bsnm{Mammone}, \binits{T.}},
\bauthor{\bsnm{Schindler}, \binits{D.}},
\bauthor{\bsnm{Salas-Boni}, \binits{R.}},
\bauthor{\bsnm{Bai}, \binits{Y.}},
\bauthor{\bsnm{Tinoco}, \binits{A.}},
\bauthor{\bsnm{Ding}, \binits{Q.}},
\bauthor{\bsnm{Hu}, \binits{X.}}:
\batitle{Insights into the problem of alarm fatigue with physiologic monitor
  devices: a comprehensive observational study of consecutive intensive care
  unit patients}.
\bjtitle{PloS one}
\bvolume{9}(\bissue{10}),
\bfpage{110274}
(\byear{2014})
\end{barticle}
\endbibitem

\bibitem{pereira2019deep}
\begin{barticle}
\bauthor{\bsnm{Pereira}, \binits{T.}},
\bauthor{\bsnm{Ding}, \binits{C.}},
\bauthor{\bsnm{Gadhoumi}, \binits{K.}},
\bauthor{\bsnm{Tran}, \binits{N.}},
\bauthor{\bsnm{Colorado}, \binits{R.A.}},
\bauthor{\bsnm{Meisel}, \binits{K.}},
\bauthor{\bsnm{Hu}, \binits{X.}}:
\batitle{Deep learning approaches for plethysmography signal quality assessment
  in the presence of atrial fibrillation}.
\bjtitle{Physiological measurement}
\bvolume{40}(\bissue{12}),
\bfpage{125002}
(\byear{2019})
\end{barticle}
\endbibitem

\bibitem{wickstrom2020uncertainty}
\begin{barticle}
\bauthor{\bsnm{Wickstr{\o}m}, \binits{K.}},
\bauthor{\bsnm{Mikalsen}, \binits{K.{\O}.}},
\bauthor{\bsnm{Kampffmeyer}, \binits{M.}},
\bauthor{\bsnm{Revhaug}, \binits{A.}},
\bauthor{\bsnm{Jenssen}, \binits{R.}}:
\batitle{Uncertainty-aware deep ensembles for reliable and explainable
  predictions of clinical time series}.
\bjtitle{IEEE Journal of Biomedical and Health Informatics}
\bvolume{25}(\bissue{7}),
\bfpage{2435}--\blpage{2444}
(\byear{2020})
\end{barticle}
\endbibitem

\bibitem{dau2019ucr}
\begin{barticle}
\bauthor{\bsnm{Dau}, \binits{H.A.}},
\bauthor{\bsnm{Bagnall}, \binits{A.}},
\bauthor{\bsnm{Kamgar}, \binits{K.}},
\bauthor{\bsnm{Yeh}, \binits{C.-C.M.}},
\bauthor{\bsnm{Zhu}, \binits{Y.}},
\bauthor{\bsnm{Gharghabi}, \binits{S.}},
\bauthor{\bsnm{Ratanamahatana}, \binits{C.A.}},
\bauthor{\bsnm{Keogh}, \binits{E.}}:
\batitle{The ucr time series archive}.
\bjtitle{IEEE/CAA Journal of Automatica Sinica}
\bvolume{6}(\bissue{6}),
\bfpage{1293}--\blpage{1305}
(\byear{2019})
\end{barticle}
\endbibitem

\bibitem{mikalsen2016learning}
\begin{botherref}
\oauthor{\bsnm{Mikalsen}, \binits{K.{\O}.}},
\oauthor{\bsnm{Bianchi}, \binits{F.M.}},
\oauthor{\bsnm{Soguero-Ruiz}, \binits{C.}},
\oauthor{\bsnm{Skr{\o}vseth}, \binits{S.O.}},
\oauthor{\bsnm{Lindsetmo}, \binits{R.-O.}},
\oauthor{\bsnm{Revhaug}, \binits{A.}},
\oauthor{\bsnm{Jenssen}, \binits{R.}}:
Learning similarities between irregularly sampled short multivariate time
  series from ehrs
(2016)
\end{botherref}
\endbibitem

\bibitem{slijepcevic2021explaining}
\begin{barticle}
\bauthor{\bsnm{Slijepcevic}, \binits{D.}},
\bauthor{\bsnm{Horst}, \binits{F.}},
\bauthor{\bsnm{Lapuschkin}, \binits{S.}},
\bauthor{\bsnm{Horsak}, \binits{B.}},
\bauthor{\bsnm{Raberger}, \binits{A.-M.}},
\bauthor{\bsnm{Kranzl}, \binits{A.}},
\bauthor{\bsnm{Samek}, \binits{W.}},
\bauthor{\bsnm{Breiteneder}, \binits{C.}},
\bauthor{\bsnm{Sch{\"o}llhorn}, \binits{W.I.}},
\bauthor{\bsnm{Zeppelzauer}, \binits{M.}}:
\batitle{Explaining machine learning models for clinical gait analysis}.
\bjtitle{ACM Transactions on Computing for Healthcare (HEALTH)}
\bvolume{3}(\bissue{2}),
\bfpage{1}--\blpage{27}
(\byear{2021})
\end{barticle}
\endbibitem

\bibitem{horsak2020gaitrec}
\begin{barticle}
\bauthor{\bsnm{Horsak}, \binits{B.}},
\bauthor{\bsnm{Slijepcevic}, \binits{D.}},
\bauthor{\bsnm{Raberger}, \binits{A.-M.}},
\bauthor{\bsnm{Schwab}, \binits{C.}},
\bauthor{\bsnm{Worisch}, \binits{M.}},
\bauthor{\bsnm{Zeppelzauer}, \binits{M.}}:
\batitle{Gaitrec, a large-scale ground reaction force dataset of healthy and
  impaired gait}.
\bjtitle{Scientific Data}
\bvolume{7}(\bissue{1}),
\bfpage{1}--\blpage{8}
(\byear{2020})
\end{barticle}
\endbibitem

\bibitem{pataky2010generalized}
\begin{barticle}
\bauthor{\bsnm{Pataky}, \binits{T.C.}}:
\batitle{Generalized n-dimensional biomechanical field analysis using
  statistical parametric mapping}.
\bjtitle{Journal of biomechanics}
\bvolume{43}(\bissue{10}),
\bfpage{1976}--\blpage{1982}
(\byear{2010})
\end{barticle}
\endbibitem

\bibitem{hsieh2021explainable}
\begin{bchapter}
\bauthor{\bsnm{Hsieh}, \binits{T.-Y.}},
\bauthor{\bsnm{Wang}, \binits{S.}},
\bauthor{\bsnm{Sun}, \binits{Y.}},
\bauthor{\bsnm{Honavar}, \binits{V.}}:
\bctitle{Explainable multivariate time series classification: A deep neural
  network which learns to attend to important variables as well as time
  intervals}.
In: \bbtitle{Proceedings of the 14th ACM International Conference on Web Search
  and Data Mining},
pp. \bfpage{607}--\blpage{615}
(\byear{2021})
\end{bchapter}
\endbibitem

\bibitem{schalk2004bci2000}
\begin{barticle}
\bauthor{\bsnm{Schalk}, \binits{G.}},
\bauthor{\bsnm{McFarland}, \binits{D.J.}},
\bauthor{\bsnm{Hinterberger}, \binits{T.}},
\bauthor{\bsnm{Birbaumer}, \binits{N.}},
\bauthor{\bsnm{Wolpaw}, \binits{J.R.}}:
\batitle{Bci2000: a general-purpose brain-computer interface (bci) system}.
\bjtitle{IEEE Transactions on biomedical engineering}
\bvolume{51}(\bissue{6}),
\bfpage{1034}--\blpage{1043}
(\byear{2004})
\end{barticle}
\endbibitem

\bibitem{lundberg2020local2global}
\begin{barticle}
\bauthor{\bsnm{Lundberg}, \binits{S.M.}},
\bauthor{\bsnm{Erion}, \binits{G.}},
\bauthor{\bsnm{Chen}, \binits{H.}},
\bauthor{\bsnm{DeGrave}, \binits{A.}},
\bauthor{\bsnm{Prutkin}, \binits{J.M.}},
\bauthor{\bsnm{Nair}, \binits{B.}},
\bauthor{\bsnm{Katz}, \binits{R.}},
\bauthor{\bsnm{Himmelfarb}, \binits{J.}},
\bauthor{\bsnm{Bansal}, \binits{N.}},
\bauthor{\bsnm{Lee}, \binits{S.-I.}}:
\batitle{From local explanations to global understanding with explainable ai
  for trees}.
\bjtitle{Nature Machine Intelligence}
\bvolume{2}(\bissue{1}),
\bfpage{2522}--\blpage{5839}
(\byear{2020})
\end{barticle}
\endbibitem

\bibitem{mishra2021survey}
\begin{botherref}
\oauthor{\bsnm{Mishra}, \binits{S.}},
\oauthor{\bsnm{Dutta}, \binits{S.}},
\oauthor{\bsnm{Long}, \binits{J.}},
\oauthor{\bsnm{Magazzeni}, \binits{D.}}:
A survey on the robustness of feature importance and counterfactual
  explanations.
arXiv preprint arXiv:2111.00358
(2021)
\end{botherref}
\endbibitem

\bibitem{kindermans2019reliability}
\begin{bchapter}
\bauthor{\bsnm{Kindermans}, \binits{P.-J.}},
\bauthor{\bsnm{Hooker}, \binits{S.}},
\bauthor{\bsnm{Adebayo}, \binits{J.}},
\bauthor{\bsnm{Alber}, \binits{M.}},
\bauthor{\bsnm{Sch{\"u}tt}, \binits{K.T.}},
\bauthor{\bsnm{D{\"a}hne}, \binits{S.}},
\bauthor{\bsnm{Erhan}, \binits{D.}},
\bauthor{\bsnm{Kim}, \binits{B.}}:
\bctitle{The (un) reliability of saliency methods}.
In: \bbtitle{Explainable AI: Interpreting, Explaining and Visualizing Deep
  Learning},
pp. \bfpage{267}--\blpage{280}.
\bpublisher{Springer}, \blocation{???}
(\byear{2019})
\end{bchapter}
\endbibitem

\bibitem{alvarez2018robustness}
\begin{botherref}
\oauthor{\bsnm{Alvarez-Melis}, \binits{D.}},
\oauthor{\bsnm{Jaakkola}, \binits{T.S.}}:
On the robustness of interpretability methods.
arXiv preprint arXiv:1806.08049
(2018)
\end{botherref}
\endbibitem

\bibitem{siddiqui2019tsviz}
\begin{barticle}
\bauthor{\bsnm{Siddiqui}, \binits{S.A.}},
\bauthor{\bsnm{Mercier}, \binits{D.}},
\bauthor{\bsnm{Munir}, \binits{M.}},
\bauthor{\bsnm{Dengel}, \binits{A.}},
\bauthor{\bsnm{Ahmed}, \binits{S.}}:
\batitle{Tsviz: Demystification of deep learning models for time-series
  analysis}.
\bjtitle{IEEE Access}
\bvolume{7},
\bfpage{67027}--\blpage{67040}
(\byear{2019})
\end{barticle}
\endbibitem

\bibitem{hartl2020explainability}
\begin{bchapter}
\bauthor{\bsnm{Hartl}, \binits{A.}},
\bauthor{\bsnm{Bachl}, \binits{M.}},
\bauthor{\bsnm{Fabini}, \binits{J.}},
\bauthor{\bsnm{Zseby}, \binits{T.}}:
\bctitle{Explainability and adversarial robustness for rnns}.
In: \bbtitle{2020 IEEE Sixth International Conference on Big Data Computing
  Service and Applications (BigDataService)},
pp. \bfpage{148}--\blpage{156}
(\byear{2020}).
\bcomment{IEEE}
\end{bchapter}
\endbibitem

\bibitem{alvarez2018towards}
\begin{botherref}
\oauthor{\bsnm{Alvarez~Melis}, \binits{D.}},
\oauthor{\bsnm{Jaakkola}, \binits{T.}}:
Towards robust interpretability with self-explaining neural networks.
Advances in neural information processing systems
\textbf{31}
(2018)
\end{botherref}
\endbibitem

\bibitem{mohseni2018human}
\begin{botherref}
\oauthor{\bsnm{Mohseni}, \binits{S.}},
\oauthor{\bsnm{Block}, \binits{J.E.}},
\oauthor{\bsnm{Ragan}, \binits{E.D.}}:
A human-grounded evaluation benchmark for local explanations of machine
  learning.
arXiv preprint arXiv:1801.05075
(2018)
\end{botherref}
\endbibitem

\bibitem{antoniadi2021current}
\begin{barticle}
\bauthor{\bsnm{Antoniadi}, \binits{A.M.}},
\bauthor{\bsnm{Du}, \binits{Y.}},
\bauthor{\bsnm{Guendouz}, \binits{Y.}},
\bauthor{\bsnm{Wei}, \binits{L.}},
\bauthor{\bsnm{Mazo}, \binits{C.}},
\bauthor{\bsnm{Becker}, \binits{B.A.}},
\bauthor{\bsnm{Mooney}, \binits{C.}}:
\batitle{Current challenges and future opportunities for xai in machine
  learning-based clinical decision support systems: a systematic review}.
\bjtitle{Applied Sciences}
\bvolume{11}(\bissue{11}),
\bfpage{5088}
(\byear{2021})
\end{barticle}
\endbibitem

\bibitem{diprose2020physician}
\begin{barticle}
\bauthor{\bsnm{Diprose}, \binits{W.K.}},
\bauthor{\bsnm{Buist}, \binits{N.}},
\bauthor{\bsnm{Hua}, \binits{N.}},
\bauthor{\bsnm{Thurier}, \binits{Q.}},
\bauthor{\bsnm{Shand}, \binits{G.}},
\bauthor{\bsnm{Robinson}, \binits{R.}}:
\batitle{Physician understanding, explainability, and trust in a hypothetical
  machine learning risk calculator}.
\bjtitle{Journal of the American Medical Informatics Association}
\bvolume{27}(\bissue{4}),
\bfpage{592}--\blpage{600}
(\byear{2020})
\end{barticle}
\endbibitem

\bibitem{su2020analyzing}
\begin{bchapter}
\bauthor{\bsnm{Su}, \binits{Z.}},
\bauthor{\bsnm{Figueiredo}, \binits{M.C.}},
\bauthor{\bsnm{Jo}, \binits{J.}},
\bauthor{\bsnm{Zheng}, \binits{K.}},
\bauthor{\bsnm{Chen}, \binits{Y.}}:
\bctitle{Analyzing description, user understanding and expectations of ai in
  mobile health applications}.
In: \bbtitle{AMIA Annual Symposium Proceedings},
vol. \bseriesno{2020},
p. \bfpage{1170}
(\byear{2020}).
\bcomment{American Medical Informatics Association}
\end{bchapter}
\endbibitem

\bibitem{wang2019designing}
\begin{bchapter}
\bauthor{\bsnm{Wang}, \binits{D.}},
\bauthor{\bsnm{Yang}, \binits{Q.}},
\bauthor{\bsnm{Abdul}, \binits{A.}},
\bauthor{\bsnm{Lim}, \binits{B.Y.}}:
\bctitle{Designing theory-driven user-centric explainable ai}.
In: \bbtitle{Proceedings of the 2019 CHI Conference on Human Factors in
  Computing Systems},
pp. \bfpage{1}--\blpage{15}
(\byear{2019})
\end{bchapter}
\endbibitem

\bibitem{kovalchuk2022three}
\begin{barticle}
\bauthor{\bsnm{Kovalchuk}, \binits{S.V.}},
\bauthor{\bsnm{Kopanitsa}, \binits{G.D.}},
\bauthor{\bsnm{Derevitskii}, \binits{I.V.}},
\bauthor{\bsnm{Matveev}, \binits{G.A.}},
\bauthor{\bsnm{Savitskaya}, \binits{D.A.}}:
\batitle{Three-stage intelligent support of clinical decision making for higher
  trust, validity, and explainability}.
\bjtitle{Journal of Biomedical Informatics}
\bvolume{127},
\bfpage{104013}
(\byear{2022})
\end{barticle}
\endbibitem

\bibitem{wexler2019if}
\begin{barticle}
\bauthor{\bsnm{Wexler}, \binits{J.}},
\bauthor{\bsnm{Pushkarna}, \binits{M.}},
\bauthor{\bsnm{Bolukbasi}, \binits{T.}},
\bauthor{\bsnm{Wattenberg}, \binits{M.}},
\bauthor{\bsnm{Vi{\'e}gas}, \binits{F.}},
\bauthor{\bsnm{Wilson}, \binits{J.}}:
\batitle{The what-if tool: Interactive probing of machine learning models}.
\bjtitle{IEEE transactions on visualization and computer graphics}
\bvolume{26}(\bissue{1}),
\bfpage{56}--\blpage{65}
(\byear{2019})
\end{barticle}
\endbibitem

\bibitem{kim2018interpretability}
\begin{bchapter}
\bauthor{\bsnm{Kim}, \binits{B.}},
\bauthor{\bsnm{Wattenberg}, \binits{M.}},
\bauthor{\bsnm{Gilmer}, \binits{J.}},
\bauthor{\bsnm{Cai}, \binits{C.}},
\bauthor{\bsnm{Wexler}, \binits{J.}},
\bauthor{\bsnm{Viegas}, \binits{F.}}, \betal:
\bctitle{Interpretability beyond feature attribution: Quantitative testing with
  concept activation vectors (tcav)}.
In: \bbtitle{International Conference on Machine Learning},
pp. \bfpage{2668}--\blpage{2677}
(\byear{2018}).
\bcomment{PMLR}
\end{bchapter}
\endbibitem

\bibitem{ghorbani2019towards}
\begin{botherref}
\oauthor{\bsnm{Ghorbani}, \binits{A.}},
\oauthor{\bsnm{Wexler}, \binits{J.}},
\oauthor{\bsnm{Zou}, \binits{J.Y.}},
\oauthor{\bsnm{Kim}, \binits{B.}}:
Towards automatic concept-based explanations.
Advances in Neural Information Processing Systems
\textbf{32}
(2019)
\end{botherref}
\endbibitem

\bibitem{goyal2019explaining}
\begin{botherref}
\oauthor{\bsnm{Goyal}, \binits{Y.}},
\oauthor{\bsnm{Feder}, \binits{A.}},
\oauthor{\bsnm{Shalit}, \binits{U.}},
\oauthor{\bsnm{Kim}, \binits{B.}}:
Explaining classifiers with causal concept effect (cace).
arXiv preprint arXiv:1907.07165
(2019)
\end{botherref}
\endbibitem

\bibitem{yeh2020completeness}
\begin{barticle}
\bauthor{\bsnm{Yeh}, \binits{C.-K.}},
\bauthor{\bsnm{Kim}, \binits{B.}},
\bauthor{\bsnm{Arik}, \binits{S.}},
\bauthor{\bsnm{Li}, \binits{C.-L.}},
\bauthor{\bsnm{Pfister}, \binits{T.}},
\bauthor{\bsnm{Ravikumar}, \binits{P.}}:
\batitle{On completeness-aware concept-based explanations in deep neural
  networks}.
\bjtitle{Advances in Neural Information Processing Systems}
\bvolume{33},
\bfpage{20554}--\blpage{20565}
(\byear{2020})
\end{barticle}
\endbibitem

\bibitem{oviedo2019fast}
\begin{barticle}
\bauthor{\bsnm{Oviedo}, \binits{F.}},
\bauthor{\bsnm{Ren}, \binits{Z.}},
\bauthor{\bsnm{Sun}, \binits{S.}},
\bauthor{\bsnm{Settens}, \binits{C.}},
\bauthor{\bsnm{Liu}, \binits{Z.}},
\bauthor{\bsnm{Hartono}, \binits{N.T.P.}},
\bauthor{\bsnm{Ramasamy}, \binits{S.}},
\bauthor{\bsnm{DeCost}, \binits{B.L.}},
\bauthor{\bsnm{Tian}, \binits{S.I.}},
\bauthor{\bsnm{Romano}, \binits{G.}}, \betal:
\batitle{Fast and interpretable classification of small x-ray diffraction
  datasets using data augmentation and deep neural networks}.
\bjtitle{npj Computational Materials}
\bvolume{5}(\bissue{1}),
\bfpage{1}--\blpage{9}
(\byear{2019})
\end{barticle}
\endbibitem

\bibitem{bau2017network}
\begin{bchapter}
\bauthor{\bsnm{Bau}, \binits{D.}},
\bauthor{\bsnm{Zhou}, \binits{B.}},
\bauthor{\bsnm{Khosla}, \binits{A.}},
\bauthor{\bsnm{Oliva}, \binits{A.}},
\bauthor{\bsnm{Torralba}, \binits{A.}}:
\bctitle{Network dissection: Quantifying interpretability of deep visual
  representations}.
In: \bbtitle{Proceedings of the IEEE Conference on Computer Vision and Pattern
  Recognition},
pp. \bfpage{6541}--\blpage{6549}
(\byear{2017})
\end{bchapter}
\endbibitem

\bibitem{cho2020interpretation}
\begin{botherref}
\oauthor{\bsnm{Cho}, \binits{S.}},
\oauthor{\bsnm{Lee}, \binits{G.}},
\oauthor{\bsnm{Chang}, \binits{W.}},
\oauthor{\bsnm{Choi}, \binits{J.}}:
Interpretation of deep temporal representations by selective visualization of
  internally activated nodes.
arXiv preprint arXiv:2004.12538
(2020)
\end{botherref}
\endbibitem

\bibitem{bruckert2020next}
\begin{barticle}
\bauthor{\bsnm{Bruckert}, \binits{S.}},
\bauthor{\bsnm{Finzel}, \binits{B.}},
\bauthor{\bsnm{Schmid}, \binits{U.}}:
\batitle{The next generation of medical decision support: A roadmap toward
  transparent expert companions}.
\bjtitle{Frontiers in artificial intelligence}
\bvolume{3},
\bfpage{507973}
(\byear{2020})
\end{barticle}
\endbibitem

\bibitem{holzinger2021toward}
\begin{barticle}
\bauthor{\bsnm{Holzinger}, \binits{A.T.}},
\bauthor{\bsnm{Muller}, \binits{H.}}:
\batitle{Toward human--ai interfaces to support explainability and causability
  in medical ai}.
\bjtitle{Computer}
\bvolume{54}(\bissue{10}),
\bfpage{78}--\blpage{86}
(\byear{2021})
\end{barticle}
\endbibitem

\bibitem{holzinger2020measuring}
\begin{barticle}
\bauthor{\bsnm{Holzinger}, \binits{A.}},
\bauthor{\bsnm{Carrington}, \binits{A.}},
\bauthor{\bsnm{M{\"u}ller}, \binits{H.}}:
\batitle{Measuring the quality of explanations: the system causability scale
  (scs)}.
\bjtitle{KI-K{\"u}nstliche Intelligenz}
\bvolume{34}(\bissue{2}),
\bfpage{193}--\blpage{198}
(\byear{2020})
\end{barticle}
\endbibitem

\bibitem{scholkopf2021toward}
\begin{barticle}
\bauthor{\bsnm{Sch{\"o}lkopf}, \binits{B.}},
\bauthor{\bsnm{Locatello}, \binits{F.}},
\bauthor{\bsnm{Bauer}, \binits{S.}},
\bauthor{\bsnm{Ke}, \binits{N.R.}},
\bauthor{\bsnm{Kalchbrenner}, \binits{N.}},
\bauthor{\bsnm{Goyal}, \binits{A.}},
\bauthor{\bsnm{Bengio}, \binits{Y.}}:
\batitle{Toward causal representation learning}.
\bjtitle{Proceedings of the IEEE}
\bvolume{109}(\bissue{5}),
\bfpage{612}--\blpage{634}
(\byear{2021})
\end{barticle}
\endbibitem

\bibitem{muller2021ten}
\begin{barticle}
\bauthor{\bsnm{Muller}, \binits{H.}},
\bauthor{\bsnm{Mayrhofer}, \binits{M.T.}},
\bauthor{\bsnm{Van~Veen}, \binits{E.-B.}},
\bauthor{\bsnm{Holzinger}, \binits{A.}}:
\batitle{The ten commandments of ethical medical ai}.
\bjtitle{Computer}
\bvolume{54}(\bissue{07}),
\bfpage{119}--\blpage{123}
(\byear{2021})
\end{barticle}
\endbibitem

\bibitem{hardt2016equality}
\begin{botherref}
\oauthor{\bsnm{Hardt}, \binits{M.}},
\oauthor{\bsnm{Price}, \binits{E.}},
\oauthor{\bsnm{Srebro}, \binits{N.}}:
Equality of opportunity in supervised learning.
Advances in neural information processing systems
\textbf{29}
(2016)
\end{botherref}
\endbibitem

\bibitem{bjerring2021artificial}
\begin{barticle}
\bauthor{\bsnm{Bjerring}, \binits{J.C.}},
\bauthor{\bsnm{Busch}, \binits{J.}}:
\batitle{Artificial intelligence and patient-centered decision-making}.
\bjtitle{Philosophy \& Technology}
\bvolume{34}(\bissue{2}),
\bfpage{349}--\blpage{371}
(\byear{2021})
\end{barticle}
\endbibitem

\bibitem{davagdorj2021explainable}
\begin{barticle}
\bauthor{\bsnm{Davagdorj}, \binits{K.}},
\bauthor{\bsnm{Bae}, \binits{J.-W.}},
\bauthor{\bsnm{Pham}, \binits{V.-H.}},
\bauthor{\bsnm{Theera-Umpon}, \binits{N.}},
\bauthor{\bsnm{Ryu}, \binits{K.H.}}:
\batitle{Explainable artificial intelligence based framework for
  non-communicable diseases prediction}.
\bjtitle{IEEE Access}
\bvolume{9},
\bfpage{123672}--\blpage{123688}
(\byear{2021})
\end{barticle}
\endbibitem

\bibitem{prentzas2019integrating}
\begin{bchapter}
\bauthor{\bsnm{Prentzas}, \binits{N.}},
\bauthor{\bsnm{Nicolaides}, \binits{A.}},
\bauthor{\bsnm{Kyriacou}, \binits{E.}},
\bauthor{\bsnm{Kakas}, \binits{A.}},
\bauthor{\bsnm{Pattichis}, \binits{C.}}:
\bctitle{Integrating machine learning with symbolic reasoning to build an
  explainable ai model for stroke prediction}.
In: \bbtitle{2019 IEEE 19th International Conference on Bioinformatics and
  Bioengineering (BIBE)},
pp. \bfpage{817}--\blpage{821}
(\byear{2019}).
\bcomment{IEEE}
\end{bchapter}
\endbibitem

\bibitem{nicolaides2010asymptomatic}
\begin{barticle}
\bauthor{\bsnm{Nicolaides}, \binits{A.N.}},
\bauthor{\bsnm{Kakkos}, \binits{S.K.}},
\bauthor{\bsnm{Kyriacou}, \binits{E.}},
\bauthor{\bsnm{Griffin}, \binits{M.}},
\bauthor{\bsnm{Sabetai}, \binits{M.}},
\bauthor{\bsnm{Thomas}, \binits{D.J.}},
\bauthor{\bsnm{Tegos}, \binits{T.}},
\bauthor{\bsnm{Geroulakos}, \binits{G.}},
\bauthor{\bsnm{Labropoulos}, \binits{N.}},
\bauthor{\bsnm{Dor{\'e}}, \binits{C.J.}}, \betal:
\batitle{Asymptomatic internal carotid artery stenosis and cerebrovascular risk
  stratification}.
\bjtitle{Journal of vascular surgery}
\bvolume{52}(\bissue{6}),
\bfpage{1486}--\blpage{1496}
(\byear{2010})
\end{barticle}
\endbibitem

\bibitem{lisboa2020explaining}
\begin{bchapter}
\bauthor{\bsnm{Lisboa}, \binits{P.J.}},
\bauthor{\bsnm{Ortega-Martorell}, \binits{S.}},
\bauthor{\bsnm{Olier}, \binits{I.}}:
\bctitle{Explaining the neural network: A case study to model the incidence of
  cervical cancer}.
In: \bbtitle{International Conference on Information Processing and Management
  of Uncertainty in Knowledge-Based Systems},
pp. \bfpage{585}--\blpage{598}
(\byear{2020}).
\bcomment{Springer}
\end{bchapter}
\endbibitem

\bibitem{cavaliere2020parkinson}
\begin{bchapter}
\bauthor{\bsnm{Cavaliere}, \binits{F.}},
\bauthor{\bsnm{Della~Cioppa}, \binits{A.}},
\bauthor{\bsnm{Marcelli}, \binits{A.}},
\bauthor{\bsnm{Parziale}, \binits{A.}},
\bauthor{\bsnm{Senatore}, \binits{R.}}:
\bctitle{Parkinson’s disease diagnosis: towards grammar-based explainable
  artificial intelligence}.
In: \bbtitle{2020 IEEE Symposium on Computers and Communications (ISCC)},
pp. \bfpage{1}--\blpage{6}
(\byear{2020}).
\bcomment{IEEE}
\end{bchapter}
\endbibitem

\bibitem{pereira2015step}
\begin{bchapter}
\bauthor{\bsnm{Pereira}, \binits{C.R.}},
\bauthor{\bsnm{Pereira}, \binits{D.R.}},
\bauthor{\bsnm{Da~Silva}, \binits{F.A.}},
\bauthor{\bsnm{Hook}, \binits{C.}},
\bauthor{\bsnm{Weber}, \binits{S.A.}},
\bauthor{\bsnm{Pereira}, \binits{L.A.}},
\bauthor{\bsnm{Papa}, \binits{J.P.}}:
\bctitle{A step towards the automated diagnosis of parkinson's disease:
  Analyzing handwriting movements}.
In: \bbtitle{2015 IEEE 28th International Symposium on Computer-based Medical
  Systems},
pp. \bfpage{171}--\blpage{176}
(\byear{2015}).
\bcomment{IEEE}
\end{bchapter}
\endbibitem

\bibitem{ibrahim2020explainable}
\begin{barticle}
\bauthor{\bsnm{Ibrahim}, \binits{L.}},
\bauthor{\bsnm{Mesinovic}, \binits{M.}},
\bauthor{\bsnm{Yang}, \binits{K.-W.}},
\bauthor{\bsnm{Eid}, \binits{M.A.}}:
\batitle{Explainable prediction of acute myocardial infarction using machine
  learning and shapley values}.
\bjtitle{IEEE Access}
\bvolume{8},
\bfpage{210410}--\blpage{210417}
(\byear{2020})
\end{barticle}
\endbibitem

\bibitem{sadhukhan2018automated}
\begin{barticle}
\bauthor{\bsnm{Sadhukhan}, \binits{D.}},
\bauthor{\bsnm{Pal}, \binits{S.}},
\bauthor{\bsnm{Mitra}, \binits{M.}}:
\batitle{Automated identification of myocardial infarction using harmonic phase
  distribution pattern of ecg data}.
\bjtitle{IEEE Transactions on Instrumentation and Measurement}
\bvolume{67}(\bissue{10}),
\bfpage{2303}--\blpage{2313}
(\byear{2018})
\end{barticle}
\endbibitem

\bibitem{krishnakumar2020towards}
\begin{bchapter}
\bauthor{\bsnm{Krishnakumar}, \binits{S.}},
\bauthor{\bsnm{Abdou}, \binits{T.}}:
\bctitle{Towards interpretable and maintainable supervised learning using
  shapley values in arrhythmia}.
In: \bbtitle{Proceedings of the 30th Annual International Conference on
  Computer Science and Software Engineering},
pp. \bfpage{23}--\blpage{32}
(\byear{2020})
\end{bchapter}
\endbibitem

\bibitem{guvenir1997supervised}
\begin{bchapter}
\bauthor{\bsnm{Guvenir}, \binits{H.A.}},
\bauthor{\bsnm{Acar}, \binits{B.}},
\bauthor{\bsnm{Demiroz}, \binits{G.}},
\bauthor{\bsnm{Cekin}, \binits{A.}}:
\bctitle{A supervised machine learning algorithm for arrhythmia analysis}.
In: \bbtitle{Computers in Cardiology 1997},
pp. \bfpage{433}--\blpage{436}
(\byear{1997}).
\bcomment{IEEE}
\end{bchapter}
\endbibitem

\bibitem{moreno2020development}
\begin{bchapter}
\bauthor{\bsnm{Moreno-Sanchez}, \binits{P.A.}}:
\bctitle{Development of an explainable prediction model of heart failure
  survival by using ensemble trees}.
In: \bbtitle{2020 IEEE International Conference on Big Data (Big Data)},
pp. \bfpage{4902}--\blpage{4910}
(\byear{2020}).
\bcomment{IEEE}
\end{bchapter}
\endbibitem

\bibitem{ahmad2017survival}
\begin{barticle}
\bauthor{\bsnm{Ahmad}, \binits{T.}},
\bauthor{\bsnm{Munir}, \binits{A.}},
\bauthor{\bsnm{Bhatti}, \binits{S.H.}},
\bauthor{\bsnm{Aftab}, \binits{M.}},
\bauthor{\bsnm{Raza}, \binits{M.A.}}:
\batitle{Survival analysis of heart failure patients: A case study}.
\bjtitle{PloS one}
\bvolume{12}(\bissue{7}),
\bfpage{0181001}
(\byear{2017})
\end{barticle}
\endbibitem

\bibitem{chmiel2021using}
\begin{barticle}
\bauthor{\bsnm{Chmiel}, \binits{F.}},
\bauthor{\bsnm{Burns}, \binits{D.}},
\bauthor{\bsnm{Azor}, \binits{M.}},
\bauthor{\bsnm{Borca}, \binits{F.}},
\bauthor{\bsnm{Boniface}, \binits{M.}},
\bauthor{\bsnm{Zlatev}, \binits{Z.}},
\bauthor{\bsnm{White}, \binits{N.}},
\bauthor{\bsnm{Daniels}, \binits{T.}},
\bauthor{\bsnm{Kiuber}, \binits{M.}}:
\batitle{Using explainable machine learning to identify patients at risk of
  reattendance at discharge from emergency departments}.
\bjtitle{Scientific reports}
\bvolume{11}(\bissue{1}),
\bfpage{1}--\blpage{11}
(\byear{2021})
\end{barticle}
\endbibitem

\bibitem{chen2020interpretable}
\begin{barticle}
\bauthor{\bsnm{Chen}, \binits{P.}},
\bauthor{\bsnm{Dong}, \binits{W.}},
\bauthor{\bsnm{Wang}, \binits{J.}},
\bauthor{\bsnm{Lu}, \binits{X.}},
\bauthor{\bsnm{Kaymak}, \binits{U.}},
\bauthor{\bsnm{Huang}, \binits{Z.}}:
\batitle{Interpretable clinical prediction via attention-based neural network}.
\bjtitle{BMC Medical Informatics and Decision Making}
\bvolume{20}(\bissue{3}),
\bfpage{1}--\blpage{9}
(\byear{2020})
\end{barticle}
\endbibitem

\bibitem{el2020hospital}
\begin{barticle}
\bauthor{\bsnm{El-Bouri}, \binits{R.}},
\bauthor{\bsnm{Eyre}, \binits{D.W.}},
\bauthor{\bsnm{Watkinson}, \binits{P.}},
\bauthor{\bsnm{Zhu}, \binits{T.}},
\bauthor{\bsnm{Clifton}, \binits{D.A.}}:
\batitle{Hospital admission location prediction via deep interpretable networks
  for the year-round improvement of emergency patient care}.
\bjtitle{IEEE Journal of Biomedical and Health Informatics}
\bvolume{25}(\bissue{1}),
\bfpage{289}--\blpage{300}
(\byear{2020})
\end{barticle}
\endbibitem

\bibitem{ochoa2021medical}
\begin{barticle}
\bauthor{\bsnm{Ochoa}, \binits{J.G.D.}},
\bauthor{\bsnm{Csisz{\'a}r}, \binits{O.}},
\bauthor{\bsnm{Schimper}, \binits{T.}}:
\batitle{Medical recommender systems based on continuous-valued logic and
  multi-criteria decision operators, using interpretable neural networks}.
\bjtitle{BMC medical informatics and decision making}
\bvolume{21}(\bissue{1}),
\bfpage{1}--\blpage{15}
(\byear{2021})
\end{barticle}
\endbibitem

\bibitem{costa2021predicting}
\begin{barticle}
\bauthor{\bsnm{Costa}, \binits{A.B.D.}},
\bauthor{\bsnm{Moreira}, \binits{L.}},
\bauthor{\bsnm{Andrade}, \binits{D.C.D.}},
\bauthor{\bsnm{Veloso}, \binits{A.}},
\bauthor{\bsnm{Ziviani}, \binits{N.}}:
\batitle{Predicting the evolution of pain relief: Ensemble learning by
  diversifying model explanations}.
\bjtitle{ACM Transactions on Computing for Healthcare}
\bvolume{2}(\bissue{4}),
\bfpage{1}--\blpage{28}
(\byear{2021})
\end{barticle}
\endbibitem

\bibitem{dwivedi2021diagnosing}
\begin{bchapter}
\bauthor{\bsnm{Dwivedi}, \binits{P.}},
\bauthor{\bsnm{Khan}, \binits{A.A.}},
\bauthor{\bsnm{Mugde}, \binits{S.}},
\bauthor{\bsnm{Sharma}, \binits{G.}}:
\bctitle{Diagnosing the major contributing factors in the classification of the
  fetal health status using cardiotocography measurements: An automl and xai
  approach}.
In: \bbtitle{2021 13th International Conference on Electronics, Computers and
  Artificial Intelligence (ECAI)},
pp. \bfpage{1}--\blpage{6}
(\byear{2021}).
\bcomment{IEEE}
\end{bchapter}
\endbibitem

\bibitem{zhai2020making}
\begin{barticle}
\bauthor{\bsnm{Zhai}, \binits{B.}},
\bauthor{\bsnm{Perez-Pozuelo}, \binits{I.}},
\bauthor{\bsnm{Clifton}, \binits{E.A.}},
\bauthor{\bsnm{Palotti}, \binits{J.}},
\bauthor{\bsnm{Guan}, \binits{Y.}}:
\batitle{Making sense of sleep: Multimodal sleep stage classification in a
  large, diverse population using movement and cardiac sensing}.
\bjtitle{Proceedings of the ACM on Interactive, Mobile, Wearable and Ubiquitous
  Technologies}
\bvolume{4}(\bissue{2}),
\bfpage{1}--\blpage{33}
(\byear{2020})
\end{barticle}
\endbibitem

\bibitem{zhang2018national}
\begin{barticle}
\bauthor{\bsnm{Zhang}, \binits{G.-Q.}},
\bauthor{\bsnm{Cui}, \binits{L.}},
\bauthor{\bsnm{Mueller}, \binits{R.}},
\bauthor{\bsnm{Tao}, \binits{S.}},
\bauthor{\bsnm{Kim}, \binits{M.}},
\bauthor{\bsnm{Rueschman}, \binits{M.}},
\bauthor{\bsnm{Mariani}, \binits{S.}},
\bauthor{\bsnm{Mobley}, \binits{D.}},
\bauthor{\bsnm{Redline}, \binits{S.}}:
\batitle{The national sleep research resource: towards a sleep data commons}.
\bjtitle{Journal of the American Medical Informatics Association}
\bvolume{25}(\bissue{10}),
\bfpage{1351}--\blpage{1358}
(\byear{2018})
\end{barticle}
\endbibitem

\bibitem{gullapalli2021opitrack}
\begin{barticle}
\bauthor{\bsnm{Gullapalli}, \binits{B.T.}},
\bauthor{\bsnm{Carreiro}, \binits{S.}},
\bauthor{\bsnm{Chapman}, \binits{B.P.}},
\bauthor{\bsnm{Ganesan}, \binits{D.}},
\bauthor{\bsnm{Sjoquist}, \binits{J.}},
\bauthor{\bsnm{Rahman}, \binits{T.}}:
\batitle{Opitrack: A wearable-based clinical opioid use tracker with temporal
  convolutional attention networks}.
\bjtitle{Proceedings of the ACM on Interactive, Mobile, Wearable and Ubiquitous
  Technologies}
\bvolume{5}(\bissue{3}),
\bfpage{1}--\blpage{29}
(\byear{2021})
\end{barticle}
\endbibitem

\bibitem{sun2021interpretable}
\begin{barticle}
\bauthor{\bsnm{Sun}, \binits{C.}},
\bauthor{\bsnm{Dui}, \binits{H.}},
\bauthor{\bsnm{Li}, \binits{H.}}:
\batitle{Interpretable time-aware and co-occurrence-aware network for medical
  prediction}.
\bjtitle{BMC medical informatics and decision making}
\bvolume{21}(\bissue{1}),
\bfpage{1}--\blpage{12}
(\byear{2021})
\end{barticle}
\endbibitem

\bibitem{thorsen2020dynamic}
\begin{barticle}
\bauthor{\bsnm{Thorsen-Meyer}, \binits{H.-C.}},
\bauthor{\bsnm{Nielsen}, \binits{A.B.}},
\bauthor{\bsnm{Nielsen}, \binits{A.P.}},
\bauthor{\bsnm{Kaas-Hansen}, \binits{B.S.}},
\bauthor{\bsnm{Toft}, \binits{P.}},
\bauthor{\bsnm{Schierbeck}, \binits{J.}},
\bauthor{\bsnm{Str{\o}m}, \binits{T.}},
\bauthor{\bsnm{Chmura}, \binits{P.J.}},
\bauthor{\bsnm{Heimann}, \binits{M.}},
\bauthor{\bsnm{Dybdahl}, \binits{L.}}, \betal:
\batitle{Dynamic and explainable machine learning prediction of mortality in
  patients in the intensive care unit: a retrospective study of high-frequency
  data in electronic patient records}.
\bjtitle{The Lancet Digital Health}
\bvolume{2}(\bissue{4}),
\bfpage{179}--\blpage{191}
(\byear{2020})
\end{barticle}
\endbibitem

\bibitem{shamout2019deep}
\begin{barticle}
\bauthor{\bsnm{Shamout}, \binits{F.E.}},
\bauthor{\bsnm{Zhu}, \binits{T.}},
\bauthor{\bsnm{Sharma}, \binits{P.}},
\bauthor{\bsnm{Watkinson}, \binits{P.J.}},
\bauthor{\bsnm{Clifton}, \binits{D.A.}}:
\batitle{Deep interpretable early warning system for the detection of clinical
  deterioration}.
\bjtitle{IEEE journal of biomedical and health informatics}
\bvolume{24}(\bissue{2}),
\bfpage{437}--\blpage{446}
(\byear{2019})
\end{barticle}
\endbibitem

\bibitem{lauritsen2020explainable}
\begin{barticle}
\bauthor{\bsnm{Lauritsen}, \binits{S.M.}},
\bauthor{\bsnm{Kristensen}, \binits{M.}},
\bauthor{\bsnm{Olsen}, \binits{M.V.}},
\bauthor{\bsnm{Larsen}, \binits{M.S.}},
\bauthor{\bsnm{Lauritsen}, \binits{K.M.}},
\bauthor{\bsnm{J{\o}rgensen}, \binits{M.J.}},
\bauthor{\bsnm{Lange}, \binits{J.}},
\bauthor{\bsnm{Thiesson}, \binits{B.}}:
\batitle{Explainable artificial intelligence model to predict acute critical
  illness from electronic health records}.
\bjtitle{Nature communications}
\bvolume{11}(\bissue{1}),
\bfpage{1}--\blpage{11}
(\byear{2020})
\end{barticle}
\endbibitem

\bibitem{jiang2021explainable}
\begin{bchapter}
\bauthor{\bsnm{Jiang}, \binits{J.}},
\bauthor{\bsnm{Hewner}, \binits{S.}},
\bauthor{\bsnm{Chandola}, \binits{V.}}:
\bctitle{Explainable deep learning for readmission prediction with tree-glove
  embedding}.
In: \bbtitle{2021 IEEE 9th International Conference on Healthcare Informatics
  (ICHI)},
pp. \bfpage{138}--\blpage{147}
(\byear{2021}).
\bcomment{IEEE}
\end{bchapter}
\endbibitem

\bibitem{luo2020hitanet}
\begin{bchapter}
\bauthor{\bsnm{Luo}, \binits{J.}},
\bauthor{\bsnm{Ye}, \binits{M.}},
\bauthor{\bsnm{Xiao}, \binits{C.}},
\bauthor{\bsnm{Ma}, \binits{F.}}:
\bctitle{Hitanet: Hierarchical time-aware attention networks for risk
  prediction on electronic health records}.
In: \bbtitle{Proceedings of the 26th ACM SIGKDD International Conference on
  Knowledge Discovery \& Data Mining},
pp. \bfpage{647}--\blpage{656}
(\byear{2020})
\end{bchapter}
\endbibitem

\bibitem{maweu2021cefes}
\begin{barticle}
\bauthor{\bsnm{Maweu}, \binits{B.M.}},
\bauthor{\bsnm{Dakshit}, \binits{S.}},
\bauthor{\bsnm{Shamsuddin}, \binits{R.}},
\bauthor{\bsnm{Prabhakaran}, \binits{B.}}:
\batitle{Cefes: A cnn explainable framework for ecg signals}.
\bjtitle{Artificial Intelligence in Medicine}
\bvolume{115},
\bfpage{102059}
(\byear{2021})
\end{barticle}
\endbibitem

\bibitem{plawiak2017ecg}
\begin{botherref}
\oauthor{\bsnm{Plawiak}, \binits{P.}}:
Ecg signals (1000 fragments).
Mendeley Data, v3
(2017)
\end{botherref}
\endbibitem

\bibitem{ma2020interpretable}
\begin{bchapter}
\bauthor{\bsnm{Ma}, \binits{D.}},
\bauthor{\bsnm{Wang}, \binits{Z.}},
\bauthor{\bsnm{Xie}, \binits{J.}},
\bauthor{\bsnm{Guo}, \binits{B.}},
\bauthor{\bsnm{Yu}, \binits{Z.}}:
\bctitle{Interpretable multivariate time series classification based on
  prototype learning}.
In: \bbtitle{International Conference on Green, Pervasive, and Cloud
  Computing},
pp. \bfpage{205}--\blpage{216}
(\byear{2020}).
\bcomment{Springer}
\end{bchapter}
\endbibitem

\bibitem{bois2020interpreting}
\begin{bchapter}
\bauthor{\bsnm{Bois}, \binits{M.D.}},
\bauthor{\bsnm{El~Yacoubi}, \binits{M.A.}},
\bauthor{\bsnm{Ammi}, \binits{M.}}:
\bctitle{Interpreting deep glucose predictive models for diabetic people using
  retain}.
In: \bbtitle{International Conference on Pattern Recognition and Artificial
  Intelligence},
pp. \bfpage{685}--\blpage{694}
(\byear{2020}).
\bcomment{Springer}
\end{bchapter}
\endbibitem

\bibitem{zhang2020adversarial}
\begin{barticle}
\bauthor{\bsnm{Zhang}, \binits{X.}},
\bauthor{\bsnm{Yao}, \binits{L.}},
\bauthor{\bsnm{Dong}, \binits{M.}},
\bauthor{\bsnm{Liu}, \binits{Z.}},
\bauthor{\bsnm{Zhang}, \binits{Y.}},
\bauthor{\bsnm{Li}, \binits{Y.}}:
\batitle{Adversarial representation learning for robust patient-independent
  epileptic seizure detection}.
\bjtitle{IEEE journal of biomedical and health informatics}
\bvolume{24}(\bissue{10}),
\bfpage{2852}--\blpage{2859}
(\byear{2020})
\end{barticle}
\endbibitem

\bibitem{obeid2016temple}
\begin{barticle}
\bauthor{\bsnm{Obeid}, \binits{I.}},
\bauthor{\bsnm{Picone}, \binits{J.}}:
\batitle{The temple university hospital eeg data corpus}.
\bjtitle{Frontiers in neuroscience}
\bvolume{10},
\bfpage{196}
(\byear{2016})
\end{barticle}
\endbibitem

\bibitem{li2020extraction}
\begin{barticle}
\bauthor{\bsnm{Li}, \binits{B.}},
\bauthor{\bsnm{Sano}, \binits{A.}}:
\batitle{Extraction and interpretation of deep autoencoder-based temporal
  features from wearables for forecasting personalized mood, health, and
  stress}.
\bjtitle{Proceedings of the ACM on Interactive, Mobile, Wearable and Ubiquitous
  Technologies}
\bvolume{4}(\bissue{2}),
\bfpage{1}--\blpage{26}
(\byear{2020})
\end{barticle}
\endbibitem

\bibitem{he2021multi}
\begin{barticle}
\bauthor{\bsnm{He}, \binits{L.}},
\bauthor{\bsnm{Liu}, \binits{H.}},
\bauthor{\bsnm{Yang}, \binits{Y.}},
\bauthor{\bsnm{Wang}, \binits{B.}}:
\batitle{A multi-attention collaborative deep learning approach for blood
  pressure prediction}.
\bjtitle{ACM Transactions on Management Information Systems (TMIS)}
\bvolume{13}(\bissue{2}),
\bfpage{1}--\blpage{20}
(\byear{2021})
\end{barticle}
\endbibitem

\end{thebibliography}
\pagebreak

\begin{appendices}
\section{Acronyms}\label{secA1}
\begin{longtable}[l]{p{.2\textwidth} p{.85\textwidth}}
\caption{List of acronyms sorted in alphabetic order.\label{tab_acronyms}}\\
  \hline
 \textbf{Acronym} & \textbf{Definition}\\
 \hline\hline
 \endfirsthead
 \hline
 \multicolumn{2}{|c|}{Table \ref{tab_acronyms} (Continued)}\\
 \hline
\textbf{Acronym} & \textbf{Definition}\\
 \hline
 \endhead
AAM&Attention Allocation Measure\\
\hline
AB& AdaBoost\\
\hline
ACE&Automatic Concept-based Explanations\\
\hline
ACSRS&Asymptomatic Carotid Stenosis and Risk of Stroke\\
\hline
AD& Alzheimer Disease\\
\hline
Ada-WHIPS&Adaptive-Weighted High Importance Path Snippets\\
\hline
adjMMD& Adjusted Maximum Mean Discrepancy\\
\hline
AF & Atrial Fibrillation\\
\hline
AI& Artificial Intelligence\\
\hline
AKI& Acute Kidney Injury\\
\hline
ALE& Accumulated Local Effects\\
\hline
ALS&Amyotrophic Lateral Sclerosis\\
\hline
AMI& Acute Myocardial Infarction\\
\hline
ANN& Artificial Neural Network\\
\hline
AUC&Area Under the ROC Curve\\
\hline
BGC& Blood Glucose Concentration\\
\hline
Bi-LSTM& Bidirectional Long Short-Term Memory\\
\hline
BP& Blood Pressure\\
\hline
BT& Body Temperature\\
\hline
CaCE&Causal Concept Effect\\
\hline
CAD& Computer-aided Diagnosis\\
\hline
CAM& Class Activation Mapping\\
\hline
CHO& Carbohydrates\\
\hline
CNN& Convolutional Neural Network\\
\hline
COPD&Chronic Obstructive Pulmonary Disease\\
\hline
CRP&C-Reactive Protein\\
\hline
CTA-TCN&Channel Temporal Attention-Temporal Convolutional Network\\
\hline
CTG& Carditocogram\\
\hline
DeepAISE&Deep
Artificial Intelligence Sepsis Expert\\
\hline
DL& Deep Learning\\
\hline
DNN& Deep Neural Network\\
\hline
DS&Dempster-Shafer\\
\hline
DSS & Decision Support System\\
\hline
DTX& Decision Tree eXplainer\\
\hline
EBM& Explainable Boosting Machine\\
\hline
ECG& Electrocardiogram\\
\hline
ED& Emergency Department\\
\hline
EDA& Electrodermal Activity\\
\hline
EEG& Electroencephalogram\\
\hline
EHR& Electronic Health Record\\
\hline
END& Early Neurological Deterioration\\
\hline
FCN&Fully Convolutional Network \\
\hline
FHS & Fetal Health Status\\
\hline
FiLM& Feature-wise Linear Modulation\\
\hline
FoG& Freeze of Gait\\
\hline
FSL& Few-Shot Learning\\
\hline
GBDT & Gradient Boosting Decision Tree\\
\hline
GD& Gradient Descent\\
\hline
GE& Grammar Evolution\\
\hline
GNN& Graph Neural Network\\
\hline
Grad-CAM& Gradient-weighted Class Activation Mapping \\
\hline
GRF& Ground Reaction Force\\
\hline
GRU& Gated Recurrent Unit\\
\hline
HAR& Human Activity Recognition\\
\hline
HF& Heart Failure\\
\hline
HiTANet&Hierarchical Time-aware Attention Network\\
\hline
HR& Heart Rate\\
\hline
HRV & Heart Rate Variability\\
\hline
ICE&Individual Conditional Expectation\\
\hline
ICU & Intensive Care Unit\\
\hline
IML & Interpretable Machine Learning\\
\hline
k-NN&k-Nearest Neighbors\\
\hline
LASSO & Least Absolute Shrinkage and Selection Operator\\
\hline
LAXCAT&Locality-Aware eXplainable Convolutional ATtention network\\
\hline
LC-LSTM-DAE&Locally Connected Long Short-Term Memory Denoising Auto-Encoder\\
\hline
LGP& Linear Genetic Programming\\
\hline
LightGBM& Light Gradient Boosting Machine\\
\hline
LIME& Local Interpretable Model-agnostic Explanations\\
\hline
LIME-SP& LIME Submodular Pick\\
\hline
LONN& Logic-Operator Neuronal Network\\
\hline
LORE&Local Rule-Based Explanations of black-box decision systems\\
\hline
LRP&Layer-wise Relevant Propagation\\
\hline
LSTM& Long Short-Term Memory\\
\hline
LT& Liver Transplantation\\
\hline
MAC-LSTM&Multi-Attention Collaborative Long Short-Term Memory\\
\hline
MCI& Mild Cognitive Impairment\\
\hline
MDA& Mean Decrease in Accuracy\\
\hline
MDI& Mean Decrease in Impurity\\
\hline
MDW&Medicaid Data Warehouse\\
\hline
MESA&Multi-Ethnic Study of Atherosclerosis\\
\hline
MFCC&Mel-Frequency Cepstral
Coefficient\\
\hline
MIMIC&Medical Information Mart for Intensive Care\\
\hline
ML& Machine Learning\\
\hline
MLP&Multi-Layer Perceptron\\
\hline
MMD& Maximum Mean Discrepancy\\
\hline
MTS& Multivariate Time Series\\
\hline
NB& Na\"ive Bayes\\
\hline
NCD & Non-Communicable Disease\\
\hline
NLP& Natural Language Processing\\
\hline
PCG& Phonocardiogram\\
\hline
PD& Parkinson Disease\\
\hline
PDP& Partial Dependence Plot\\
\hline
PIC& Pediatric Intensive Care\\
\hline
PL& Prototype Learning\\
\hline
PPG&Photoplethysmogram\\
\hline
PRN& Partial Response Network\\
\hline
QoL& Quality of Life\\
\hline
RETAIN&REverse Time AttentIoN \\
\hline
RF& Random Forest\\
\hline
RNN &Recurrent Neural Network\\
\hline
ROSMAP&Religious Orders Study and Rush Memory and Aging Project\\
\hline
SAX&Symbolic Aggregate ApproXimation\\
\hline
SCS& System Causability Scale\\
\hline
SHAP&Shapley Additive Explanations\\
\hline
SMILE&Systems Metabolomics using Interpretable Learning and Evolution\\
\hline
SPM&Statistical Parametric Mapping\\
\hline
SpO$_2$& Pulse Oximetry\\
\hline
SSI&Surgical Site Infection\\
\hline
ST& Skin Temperature\\
\hline
SVM & Support Vector Machine\\
\hline
T-CAV&Testing with Concept Activation Vectors\\
\hline
T-LSTM& Time-aware Long Short-Term Memory\\
\hline
t-SNE& t-distributed Stochastic Neighbor Embedding\\
\hline
TCN& Temporal Convolutional Network\\
\hline
TCoN&Time-aware and Co-occurrence-aware deep learning Network\\
\hline
TF-IDF& Term Frequency Inverse Document Frequency\\
\hline
TITV&Time-Invariant and Time-Variant\\
\hline
WCPH&Weibull Cox Proportional Hazards\\
\hline
XAI & Explainable Artificial Intelligence\\
\hline
XGBoost & eXtreme Gradient Boosting\\
\hline
\end{longtable}

\section{Preliminary works}
\label{secB}
\begin{table}[!htbp]
\footnotesize
\begin{minipage}{\textwidth}
\caption{Preliminary studies applying XAI to tabular data.}\label{res_tab_1}
\begin{tabular*}{\textwidth}[l]{p{.05\textwidth}p{.05\textwidth} p{.2\textwidth}p{.1\textwidth}p{.1\textwidth}p{.15\textwidth}p{.15\textwidth}}
\toprule
\textbf{Ref.} &\textbf{\# Cit.}&\textbf{Application} & \textbf{Input Data} & \textbf{AI model(s)} & 
\textbf{XAI method(s)} &\textbf{Dataset(s)} \\
\midrule
\cite{davagdorj2021explainable} ($2021$)&$5$&NCD prediction& EHR& DNN & SHAP & NHANES dataset\footnote{https://www.cdc.gov/nchs/nhanes}\\
\midrule
\cite{prentzas2019integrating} ($2019$)&$25$&Stroke prediction& EHR & RF &InTrees,\newline symbolic AI& ACSRS dataset \cite{nicolaides2010asymptomatic}\\
\midrule
\cite{lisboa2020explaining} ($2020$)&$2$& Cervical cancer diagnosis&EHR&MLP&PRN&Retrospective study\\
\midrule
\cite{leung2021explainable} ($2021$)&$11$&COVID-$19$ diagnosis& EHR&RF, FSL-ANN&SHAP&COVID-$19$ dataset\footnote{https://www.kaggle.com/einsteindata4u/covid19\label{covid}}\\
\midrule
\cite{cavaliere2020parkinson} ($2020$)&$10$&PD diagnosis&Handmade writings& Evolutionary algorithms&GE&HandPD dataset \cite{pereira2015step}\\ 
\midrule
\cite{ibrahim2020explainable} ($2020$)&$15$&AMI prediction& ECG features& XGBoost&SHAP&ECG-ViEW II DB \cite{sadhukhan2018automated}\\
\midrule
\cite{krishnakumar2020towards} ($2020$)&$1$&Arrhythmia detection& ECG features& stacked NB-k-NN-RF ensemble& SHAP & Arrhythmia dataset \cite{guvenir1997supervised}\\
\midrule
\cite{moreno2020development} ($2020$)&$6$& Survival prediction in HF patients& EHR&XGBoost&SHAP&HF dataset \cite{ahmad2017survival}\\
\midrule
\cite{chmiel2021using} ($2021$)&$0$& Hospital reattendance prediction& EHR& XGBoost&SHAP&Retrospective study\\
\midrule
\cite{chen2020interpretable} ($2020$)&$15$&Hospital readmission prediction in HF patients&EHR&MLP&Attention&Retrospective study\\
\midrule
\cite{el2020hospital} ($2020$)&$7$&Hospital admission location prediction&EHR&ANN& Saliency maps&Retrospective study\\
\midrule
\cite{ochoa2021medical} ($2021$)&$7$&Therapy recommendation&EHR& LONN&Fuzzy rules&Pakistan database\footnote{https:// archive. ics.uci. edu/ml/datasets/ Heart+ failure+ clinical+records}\\
\midrule
\cite{costa2021predicting} ($2021$)&$1$&Pain relief evolution assessment&EHR&RF-XGBoost ensemble& SHAP& Retrospective study\\
\midrule
\cite{dwivedi2021diagnosing} ($2021$)&$0$&FHS detection& CTG/HRV features& LightGBM&feature permutation, SHAP&Kaggle Fetal Health dataset\footnote{https://www.kaggle.com/andrewmvd/fetal-health-classification}\\
\midrule
\cite{zhai2020making} ($2020$)&$32$&Sleep stage classification& Actigraphy, HR/HRV features& RF&SHAP&MESA sleep dataset \cite{zhang2018national}\\
\midrule
\cite{gullapalli2021opitrack} ($2021$)&$1$&Clinical opioid use tracking&HR/HRV, EDA, ST, $3$-D Acc features& CTA-TCN& Attention&Pilot study\\
\bottomrule
\end{tabular*}
\end{minipage}
\end{table}

\begin{table}[!htbp]
\footnotesize
\begin{minipage}{\textwidth}
\caption{Preliminary studies applying XAI to time series data.}\label{res_tab_2}
\begin{tabular*}{\textwidth}[l]{p{.05\textwidth}p{.05\textwidth}p{.2\textwidth}p{.15\textwidth}p{.1\textwidth}p{.1\textwidth}p{.15\textwidth}}
\toprule
\textbf{Ref.}&\textbf{\# Cit.}&\textbf{Application} & \textbf{Input Data} & \textbf{AI model(s)} & 
\textbf{XAI method(s)} &\textbf{Dataset(s)} \\
\midrule
\cite{sun2021interpretable} ($2021$)&$0$&Mortality prediction& Longitudinal EHR& TCoN& Attention&MIMIC-III \cite{johnson2016mimic} \\
\midrule
\cite{thorsen2020dynamic} ($2020$)&$11$&Mortality prediction&Longitudinal EHR& LSTM&SHAP&Retrospective study\\
\midrule
\cite{shamout2019deep} ($2019$)&$39$&Early detection of in-hospital deterioration&HR, HRV, BT, BP, SpO$_2$ & Bi-LSTM ensemble& Attention& Retrospective study\\
\midrule
\cite{lauritsen2020explainable} ($2020$)&$124$&Acute critical illness prediction&Longitudinal EHR&TCN& LRP&Retrospective study\\
\midrule
\cite{jiang2021explainable} ($2021$)&$0$&Hospital readmission prediction&Longitudinal EHR& T-LSTM& Attention& MIMIC-III, New York State MDW\\
\midrule
\cite{luo2020hitanet} ($2020$)&$45$&Disease diagnosis&Longitudinal EHR& HiTANet&Attention&UCI COPD, HF, CKD datasets\\
\midrule
\cite{maweu2021cefes} ($2021$)&$16$&Arrhythmia detection&ECG&CNN&Grad-CAM&MIT-BIH
Arrhythmia database subset\cite{plawiak2017ecg}\\
\midrule
\cite{ma2020interpretable} ($2020$)&$0$&AMI detection& MTS&CNN& PL &MIMIC-III\\
\midrule
\cite{bois2020interpreting} ($2020$)&$2$&BGC forecasting&longitudinal BGC, CHO intake, and insulin data& RETAIN \cite{choi2016retain}&Attention&IDIAB dataset\\
\midrule
\cite{zhang2020adversarial} ($2020$)&$36$&Epileptic seizure detection&EEG&CNN&Attention& TUH-EEG database \cite{obeid2016temple}\\
\midrule
\cite{li2020extraction} ($2021$)&$24$&Stress detection& EDA, ST, $3$-D Acc&LC-LSTM-DAE&Attention&Pilot study\\
\midrule
\cite{he2021multi} ($2021$)&$1$&BP forecasting &Longitudinal EHR&MAC-LSTM&Attention&Retrospective study\\
\bottomrule
\end{tabular*}
\end{minipage}
\end{table}

\end{appendices}

\end{document}